\documentclass{article} % For LaTeX2e
\PassOptionsToPackage{table,xcdraw,dvipsnames}{xcolor}
\usepackage[preprint]{colm2025_conference}

\usepackage{microtype}
\usepackage{hyperref}
\usepackage{url}
\usepackage{booktabs}
\usepackage{amsmath}
\usepackage{lineno}
\usepackage{graphicx}
\usepackage{subfigure}
\usepackage{wrapfig}
\usepackage{xspace}
\usepackage[normalem]{ulem}
\usepackage{enumitem}

\definecolor{darkblue}{rgb}{0,0,0.5}
\hypersetup{colorlinks=true, citecolor=darkblue, linkcolor=darkblue, urlcolor=darkblue}
\definecolor{lightblue}{RGB}{221,235,247}
\newcommand{\ourmethod}{SimpleRL-Zoo\xspace}
\usepackage[most]{tcolorbox}

\title{SimpleRL-Zoo: Investigating and Taming Zero Reinforcement Learning for Open Base Models in the Wild}
\author{Weihao Zeng\thanks{Equal Contribution. Correspondence to Weihao Zeng (wzengak@connect.ust.hk), Yuzhen Huang (yhuanghj@cse.ust.hk), and Junxian He (junxianh@cse.ust.hk).}\hspace{4pt}$^1$ \quad Yuzhen Huang$^{*1}$ \quad Qian Liu$^{*2}$ \quad
Wei Liu$^1$ \quad
Keqing He$^3$ \\ \textbf{Zejun Ma$^2$ \quad Junxian He$^1$}\\
$^1$HKUST \quad $^2$TikTok \quad $^3$Meituan\\
\href{https://github.com/hkust-nlp/simpleRL-reason}{https://github.com/hkust-nlp/simpleRL-reason}
}

\newcommand{\jhc}[2]{\bgroup\textcolor{magenta}{\sout{#1} #2}\egroup}

\usepackage{arydshln}
\begin{document}

\ifcolmsubmission
\linenumbers
\fi

\maketitle

\begin{abstract}
% Advanced large language model has achieved leading performance
% Long chain-of-thought (CoT) reasoning, with high-level reasoning mechanisms such as reflection and verification, has been shown to be a powerful approach for scaling test-time computation and achieving remarkable performance in extremely complex reasoning tasks.

% Advanced models can perform long chain-of-thought (CoT) reasoning on complex tasks to achieve enhanced performance. 
DeepSeek-R1 has shown that long chain-of-thought (CoT) reasoning can naturally emerge through a simple reinforcement learning (RL) framework with rule-based rewards, where the training may directly start from the base models—a paradigm referred to as \emph{zero RL training}. Most recent efforts to reproduce zero RL training have primarily focused on the Qwen2.5 model series, which may not be representative as we find the base models already exhibit strong instruction-following and self-reflection abilities.
In this work, we investigate zero RL training across 10 diverse base models, spanning different families and sizes including LLama3-8B, Mistral-7B/24B, DeepSeek-Math-7B, Qwen2.5-math-7B, and all Qwen2.5 models from 0.5B to 32B. 
% can several critical questions remain unanswered: (1) Can smaller, various base models exhibit similar emergent reasoning with limited, simple data? (2) Does increased CoT length always lead to increased cognitive phenomena such as self-reflection (i.e., the``aha moment'')? (3) What are the essential designs that enable the emergence of long CoT? In this work, we take mathematical reasoning as an example and try to answer these questions. We conduct extensive zero-training experiments across various base models, spanning different families and sizes, including LLama3.1-8B, Mistral-7B/24B, DeepSeekMath-7B, and Qwen2.5-0.5B/1.5B/7B/14B/32B. To keep the recipe simple, we restrict our training data to the original GSM8K and MATH datasets. 
Leveraging several key design strategies—such as adjusting format reward and controlling query difficulty—we achieve substantial improvements in both reasoning accuracy and response length across most settings.
However, by carefully monitoring the training dynamics, we observe that different base models exhibit distinct patterns during training. For instance, the increased response length does not always correlate with the emergence of certain cognitive behaviors such as verification (i.e., the ``aha moment"). Notably, we observe the ``aha moment'' for the first time in small models not from the Qwen family.
% Through comprehensive analyses of the results, we uncover several intriguing findings. For example, increased CoT length does not always correspond to the ``aha moment'', and different base models exhibit distinct behaviors during training. 
% We also compare our approach to traditional RL pipelines that follow a traditional SFT stage, demonstrating that such common practices can slow down or even inhibit the emergence of long CoT reasoning. 
We share the key designs that enable successful zero RL training, along with our findings and practices. 
To facilitate further research, we open-source the code, models, and analysis tools.
\end{abstract}

% 1. The first comprehensive experiment to perform zero RL training on different base models (apart from the Qwen base model) and conduct in-depth analysis, including analysis of length and performance, using comprehensive metrics (including pass@k) to analyze length changes, and whether reflection is produced (in this experimental setting, we need to align most of the settings). **TODO:** We need to quickly align key settings, including whether to remove formatting, the difficulty level of prompts used, and finalize the experiments.
% 2. Analysis of key tricks affecting zero experiment effectiveness, including stop ids, reward design (with or without format), prompt difficulty design, and other key hyperparameters such as batch size/KL coefficient/temperature. **TODO:** We need to select 2-3 representative models for analysis experiments
% 3. Comprehensive analysis of why base models may or may not produce emergence, including patterns of emergence in base models, pass@K factors. Establish connections between base models and the aha momentTODO:
% 4. Explain why traditional SFT experiments don't work? For example, why might using short CoT for SFT be harmful? (Expected conclusion: using short CoT for SFT suppresses emergence) TODO: Also conduct SFT on the representative models from point 2 for analysis experiments

\begin{figure}[htbp]
\vspace{-10pt}
        \centering
\includegraphics[width=0.98\textwidth]{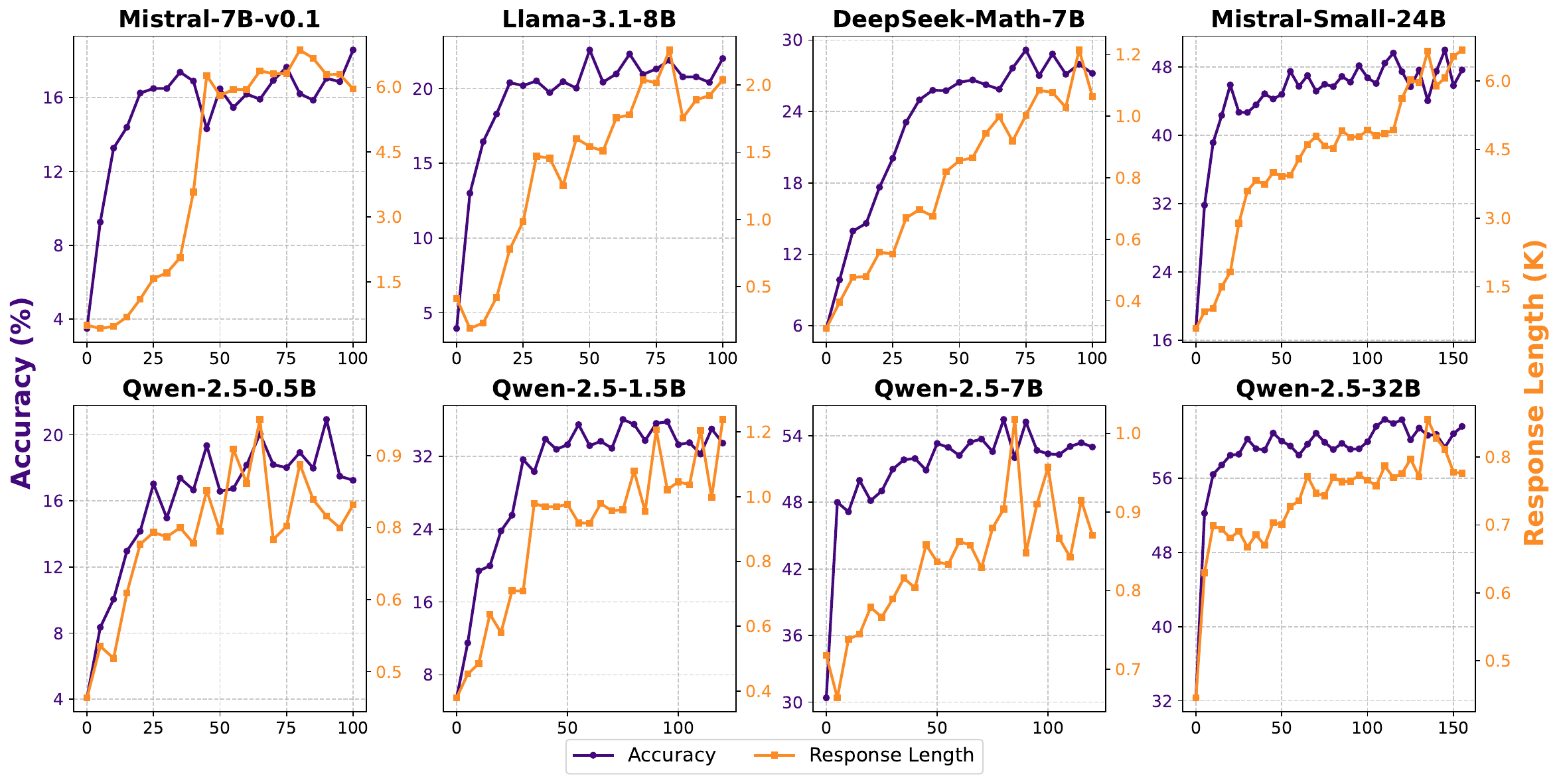}
\vspace{-10pt}
\caption{Accuracy and response length across training iterations for different models, averaged on GSM8K, MATH500, Minerva Math, OlympiadBench, AIME24, and AMC23. Per-benchmark results are in Figure~\ref{fig:appx_acc&len} (Appendix~\ref{appx:DetailedResult}). All training starts from base models.
% All models are trained starting from their base version.
        }
        \label{fig1:acc&len}
    \vspace{-10pt}
\end{figure}

\section{Introduction}
Large reasoning models, including OpenAI-o1~\citep{jaech2024openai}, DeepSeek-R1~\citep{guo2025deepseek}, and Kimi-k1.5~\citep{team2025kimi}, demonstrate remarkable abilities. These models excel at generating long Chains-of-Thought (CoT)~\citep{wei2022chain} responses when solving complex tasks and exhibit advanced, reflection-like reasoning behaviors. 
Recently, DeepSeek-R1~\citep{guo2025deepseek} has revealed that starting from pretrained models (i.e., base models), pure reinforcement learning (RL) with rule-based reward can lead to the spontaneous emergence of long CoT and self-reflection behaviors, called the ``aha moment''. This RL training paradigm starting from base models is often referred to as \emph{zero RL training}.

% Despite DeepSeek's generous open-sourcing of their models, the community still lacks a sufficient understanding of the "zero training" process. 
While the success of zero RL training was initially demonstrated using DeepSeek-V3~\citep{deepseekai2025deepseekv3technicalreport}, a model with 671B parameters, it remained unclear whether such emergent phenomena persist in generally smaller and less capable open base models.
Recent open-source efforts exploring zero-training approaches have predominantly centered on the Qwen2.5-series models~\citep{zeng2025simplerl,yeo2025demystifying,xie2025logic,OpenReasonerZero2025,yu2025dapoopensourcellmreinforcement}, which, even as base models, exhibit strong instruction-following capabilities and display notable cognitive behaviors such as backtracking and verification from the beginning, as we will detail in \S\ref{sec:qwen_behabiur}.
Moreover, the analyses of model behavior in these studies remain largely superficial, focusing primarily on metrics such as response length and accuracy. These observations neither clearly establish whether the models' reasoning behaviors actually change nor clarify the mechanisms underlying the emergence of effective reasoning, leaving a significant gap in understanding.
% Although DeepSeek has generously open-sourced their models, the research community still lacks a deep understanding of the ``zero training" mechanism. Most existing analyses remain superficial—primarily noting increased response lengths or shifts in word frequency—without clearly linking these observations to the underlying mechanisms that enable effective reasoning behaviors to emerge~\citep{OpenReasonerZero2025,yeo2025demystifying,xie2025logic}.

To provide a more transparent understanding of zero RL training across different base models in the wild, this paper addresses the following key questions: (1) How do reasoning capabilities develop across various models during zero RL training? (2) Does an ``aha moment'' still occur for base models that initially lack strong instruction-following and self-verification abilities? (3) What are the critical factors for ensuring successful zero RL training across diverse base models?

% Why did previous attempts at ``zero-training" typically fail for most base models? What are the key secrets that enable successful ``zero-training" of these models?
%(3) How does the difficulty of emergent reflection behavior vary with different base models? 

%(3) Do traditional supervised fine-tuning (SFT) methods—such as enforcing shorter CoT~\citep{yu2023metamath,luo2023wizardmath}—impede the natural development of complex reasoning patterns?

To this end, we perform zero RL training across a diverse range of model series and sizes, including Mistral-7B~\citep{jiang2023mistral7b}, Mistral-24B~\citep{mistral2024small}, Llama3-8B~\citep{dubey2024llama}, DeepSeek-Math-7B~\citep{shao2024deepseekmath}, Qwen2.5-0.5B/1.5B/7B/14B/32B~\citep{yang2024qwen2}, as well as Qwen2.5-Math-7B~\citep{yang2024qwen2math}. 
% To keep the recipe simple, our experiments rely solely on the GSM8K~\citep{cobbe2021training} and MATH~\citep{hendrycks2021measuring} training data, for rule-based reward modeling. 
To maintain simplicity in the training recipe, our experiments rely exclusively on the training sets of GSM8K~\citep{cobbe2021training} and MATH~\citep{hendrycks2021measuring} datasets for rule-based reward modeling.
It is worth noting that we adopt the same training hyperparameters to train all the models.
Using GRPO~\citep{shao2024deepseekmath} as the RL algorithm, combined with several critical factors that we identified, we obtain significant improvements in model accuracy across all base models, along with a notable increase in response length for 9 out of the 10 models, with the exception of Qwen2.5-Math-7B. 
However, through careful monitoring of training dynamics and reasoning behaviors, we find that different base models exhibit distinct patterns during training. Also, certain specific factors require careful attention to ensure successful zero RL training. Below, we summarize our key findings.
\begin{tcolorbox}[colback=lightblue!80,breakable]
\begin{enumerate}[leftmargin=1em]
    \item Increased response length does not always correspond to an ``aha moment'' -- Interestingly, for most Qwen2.5 models, which form the foundation of most recent open-source efforts, we do not observe a rise in the frequency of certain cognitive behaviors, such as self-reflection, despite the increase in response length. (\textsection\ref{sec:qwen_behabiur})
    \item For the first time, we observe a significant increase in the frequency of specific cognitive reasoning behaviors, such as verification, in small models outside the Qwen family, notably in the Llama3-8B and DeepSeek-Math-7B models. (\textsection\ref{sec:deepseek-math_behabiur})
    % \qian{And we may highlight the contribution of using the same hyper-parameter on all models to make their response length increasing and performance increasing?}
    % \qian{For the first time, we observe a significant increase in the frequency of specific cognitive reasoning behaviors, such as verification, in small models not from Qwen family, the Llama3-8B and DeepSeek-Math-7B models.}
    \item Enforcing rigid format reward (e.g., enclosing answers within boxes)~\citep{guo2025deepseek} significantly penalizes exploration~\citep{singh2023beyond,wang2024planning}, particularly for base models that initially struggle with instruction following. This restriction lowers their  performance ceiling and often induces overthinking behaviors~\citep{chen2024not}. (\textsection\ref{sec:remove_format})
    \item The difficulty level of the training data must align closely with the base model's intrinsic exploration capabilities, otherwise zero RL will fail. (\textsection\ref{sec:data_complextiy_behaviur})
    \item In contrast to the observation in~\citet{shao2024deepseekmath}, zero RL training lifts pass@k accuracy by 10-30 absolute points, a strong evidence confirming zero RL training is not just reranking responses. (\textsection\ref{sec:lift_pass_k})
    \item We revisit the traditional training pipeline that performs SFT to learn to follow instructions before RL training. Specifically, we use conventional SFT datasets as a cold start for RL—a de facto approach prior to the release of DeepSeek-R1. While high-quality CoT data~\citep{li2024numinamath} can rapidly enhance a base model's performance through imitation, we find that it significantly limits the model's ability to explore freely during RL. This constraint diminishes post-RL performance and suppresses the emergence of advanced reasoning capabilities. (\textsection\ref{sec:short_cot_influence})
\end{enumerate}
\end{tcolorbox}

\section{On Emerging Reasoning in Zero RL Training}

Existing research on zero RL training primarily focuses on Qwen2.5-series models, tracking only superficial metrics like accuracy and response length~\citep{zeng2025simplerl,OpenReasonerZero2025,yu2025dapoopensourcellmreinforcement}. However, Qwen2.5 models, due to their extensive use of synthetic data during pretraining, already exhibit instruction-following abilities and reflective behaviors, which may not represent base models in diverse scenarios. Additionally, an increase in response length does not necessarily indicate the emergence of cognitive behaviors and can sometimes result from meaningless repetition.
To address these issues, this section explores zero RL training across various base models of different sizes and families. By monitoring a range of metrics beyond accuracy and response length, we aim to provide a more comprehensive and transparent understanding of zero RL training for open base models in the wild.

\subsection{Experimental Setup}
\label{sec:setup}

\label{sec:training_algorithm}

\paragraph{Training Algorithm:}
In our study, we follow the zero RL training recipe in ~\citet{guo2025deepseek} using various open base models, employing the GRPO algorithm~\citep{shao2024deepseekmath}. Here, zero RL training refers to RL directly from the base model without any prior supervised fine-tuning (SFT). A detailed introduction to GRPO is provided in Appendix~\ref{sec:detailed_grpo}.
\label{sec:dataset_setting}

\paragraph{Training Dataset:} We use the GSM8K~\citep{cobbe2021training} and MATH~\citep{hendrycks2021measuring} training datasets. In our experiments, we find that data difficulty is critical for successful zero RL (\textsection\ref{sec:data_complextiy_behaviur}) and it is necessary to use data that aligns with the model's capability. To investigate this phenomenon, we categorize the data into three difficulty levels: Easy (GSM8K and MATH lv.1), Medium (MATH lv.1–4), and Hard (MATH lv.3–5), with each category containing roughly 8,000 problems.
\label{sec:reward_remove}

\textbf{Reward: } We use a rule-based reward function that assigns +1 for correct answers and 0 for incorrect ones. Unlike prior works~\citep{deepscaler2025,chen2025empirical}, we avoid format-based reward, which may hinder exploration, particularly for base models struggling with format adherence, as detailed in \textsection\ref{sec:remove_format}.

\textbf{Models: }We conduct zero RL training experiments on Llama-3.1-8B, DeepSeek-Math-7B, Mistral-v0.1-7B, Mistral-Small-24b-Base-2501, and Qwen-2.5 (0.5B, 1.5B, 7B, 14B, 32B). As we perform experiments for a variety of models, under extremely simple settings with small, simple datasets and only correctness reward, we refer to our obtained models as \emph{SimpleRL-Zoo} to represent a simple training recipe for a zoo of open base models. In our preliminary experiments, we observe that using complex prompts with models that have weak instruction-following capabilities often results in instability during training. Therefore, we apply simpler prompts to some models (Llama-3.1-8B, Mistral-v0.1-7B, and Qwen-2.5-0.5B/1.5B). Examples of these prompts are shown in Figure~\ref{fig:prompt_case} in the Appendix.

\textbf{Benchmark: }We evaluate performance on standard mathematical reasoning benchmarks, including GSM8K~\citep{cobbe2021training}, MATH 500~\citep{hendrycks2021measuring}, Minerva Math~\citep{lewkowycz2022solving}, and OlympiadBench~\citep{he2024olympiadbench}, as well as on competition-level benchmarks such as AIME 2024 and AMC 2023.

For more experimental setup details, please refer to Appendix~\ref{appx:detailed_setup}.
% \paragraph{Other Configurations}: Training is conducted using the verl framework (Sheng et al., 2024), with additional details in Appendix C.2.

% Specifically, during training, we use a prompt batch size of 1024, generate 8 rollouts per prompt, set a maximum rollout length of 8,192 tokens, and train using a mini-batch size of 256. 
% It is worth noting that we use the same training hyperparameters to train all the models.
% During evaluation, we set the sampling temperature to 1.0 and allow a maximum generation length of 16,384 tokens. For most benchmarks, we report pass@1 results. However, for the AIME 2024 benchmark specifically, we report both pass@1 and average accuracy computed over 32 samples (avg@32) due to limited data points. We provide detailed training and evaluation details in the Appendix ~\ref{sec:train_evaluate_details}.
% }

\subsection{Evaluation Metrics}
\label{sec:eval_metrics}
During training, we monitor standard metrics such as accuracy and response length across benchmarks. In our preliminary experiment, we observe that response length as a metric is quite superficial and cannot accurately reflect changes in the model's reasoning behavior. Therefore, we adopt the following metrics additionally:

\textbf{Reasoning Behavior Ratio: }To better understand the model's reasoning patterns throughout the training process, we adopt the cognitive behavior framework proposed by ~\citet{gandhi2025cognitive} and use GPT-4o~\citep{hurst2024gpt} to identify reasoning-related behaviors, including ``Backtracking", ``Verification", ``Subgoal Setting", and ``Enumeration". We compare the consistency between GPT-4o and human annotations of reasoning-related behaviors in the Appendix~\ref{sec:human_consistency}.
We report the ratio of responses that contain such cognitive behaviors.
While some recent studies suggest tracking reflection behavior using related keywords~\citep{yeo2025demystifying,xie2025logic} as monitoring signals, we argue that these keywords only exhibit only a weak correlation with high-level reasoning patterns like reflection and verification. As a result, they fail to adequately capture the development of these reasoning processes. We place the setting details, comparisons of different tracking methods, and reasoning behavior cases of different models in Appendix~\ref{appx:bahaviour}.

% Further details refer to Appendix~\ref{appx:bahaviour}.

\begin{table*}[t]
\centering
\resizebox{\textwidth}{!}{
\begin{tabular}{lcccccccc}
\toprule
\multicolumn{1}{c}{\textbf{Model}} & \textbf{GSM8K} & \textbf{\begin{tabular}[c]{@{}c@{}}MATH\\ 500\end{tabular}} & \textbf{\begin{tabular}[c]{@{}c@{}}Minerva\\ Math\end{tabular}} & \textbf{\begin{tabular}[c]{@{}c@{}}Olympiad\\ Bench\end{tabular}} & \textbf{\begin{tabular}[c]{@{}c@{}}AIME24 \\ (Pass@1)\end{tabular}} & \textbf{\begin{tabular}[c]{@{}c@{}}AIME24 \\ (Avg@32)\end{tabular}} & \textbf{AMC23} & \textbf{Avg.} \\ \midrule
\multicolumn{9}{c}{\textit{Llama, DeepSeek and Mistral Models}}   \\
Mistral-v0.1-7B  & 21.2  & 4.2      & 4.0       & 2.4        & 0.0        & 0.0        & 0.0  & 5.3  \\
\rowcolor[rgb]{ .867, .922, .969}~~~~$\hookrightarrow$~+~\ourmethod  & 75.0  & 15.8      & 6.6       & 4.1        & 0.0        & 0.2        & 10.0  & 18.6  \\
Llama-3.1-8B    & 39.7  & 13.6      & 4.8       & 3.1        & 0.0        & 0.2        & 2.5  & 10.6  \\
\rowcolor[rgb]{ .867, .922, .969}~~~~$\hookrightarrow$~+~\ourmethod  & 79.2  & 23.0      & 9.6       & 5.3        & 0.0        & 0.2        & 15.0  & 22.0  \\
DeepSeek-Math-7B  & 28.4  & 19.4      & 5.5       & 4.7        & 0.0        & 0.0        & 10.0  & 11.3  \\
\rowcolor[rgb]{ .867, .922, .969}~~~~$\hookrightarrow$~+~\ourmethod & 78.5  & 39.6      & 21.0      & 12.6        & 3.3        & 0.6        & 20.0  & 29.2  \\
Mistral-Small-24B  & 78.6  & 43.6      & 10.7      & 11.6        & 3.3        & 0.5        & 17.5  & 27.6  \\
\rowcolor[rgb]{ .867, .922, .969}~~~~$\hookrightarrow$~+~\ourmethod  & 92.0  & 70.6      & 36.8      & 36.6        & 16.7        & 13.1        & 45.0  & 49.6  \\ \midrule
\multicolumn{9}{c}{\textit{Qwen Series Models}}                                        \\
Qwen-2.5-0.5B    & 36.7  & 15.8      & 4.8       & 2.8        & 0.0        & 0.3        & 12.5  & 12.1  \\
\rowcolor[rgb]{ .867, .922, .969}~~~~$\hookrightarrow$~+~\ourmethod  & 49.5  & 34.4      & 10.3      & 8.9        & 0.0        & 0.7        & 22.5  & 20.9  \\
Qwen-2.5-1.5B    & 55.7  & 29.6      & 6.6       & 6.5        & 0.0        & 0.1        & 12.5  & 18.5  \\
\rowcolor[rgb]{ .867, .922, .969}~~~~$\hookrightarrow$~+~\ourmethod  & 74.4  & 59.0      & 20.2      & 21.0        & 6.7        & 4.2        & 35.0  & 36.1  \\
Qwen-2.5-7B    & 88.2  & 64.6      & 25.7      & 30.1        & 3.3        & 0.3        & 30.0  & 40.3  \\
\rowcolor[rgb]{ .867, .922, .969}~~~~$\hookrightarrow$~+~\ourmethod  & 91.7  & 78.2      & 38.6      & 40.4        & 20.0        & 15.6        & 62.5  & 55.2  \\
Qwen-2.5-Math-7B  & 65.5  & 63.6      & 12.5      & 25.8        & 13.3        & 8.6        & 42.5  & 37.2  \\
\rowcolor[rgb]{ .867, .922, .969}~~~~$\hookrightarrow$~+~\ourmethod  & 90.2  & 80.2      & 37.5      & 39.0        & 40.0        & 24.0        & 70.0  & 59.5  \\
Qwen-2.5-14B    & 91.6  & 65.4      & 24.3      & 33.5        & 6.7        & 3.4        & 37.5  & 43.2  \\
\rowcolor[rgb]{ .867, .922, .969}~~~~$\hookrightarrow$~+~\ourmethod  & 94.4  & 80.2      & 40.4      & 44.9        & 23.3        & 14.2        & 57.6  & 56.8  \\
Qwen-2.5-32B    & 92.9  & 68.6      & 27.9      & 31.1        & 10.0        & 4.5        & 45.0  & 45.9  \\
\rowcolor[rgb]{ .867, .922, .969}~~~~$\hookrightarrow$~+~\ourmethod  & 95.9  & 82.4      & 42.6      & 46.4        & 36.7        & 27.2        & 67.5  & 61.9  \\ \bottomrule
\end{tabular}
}
\caption{Detailed performance of various models across multiple benchmarks. The blue lines represent the models trained with our recipe. AIME is evaluated in two ways: Pass@1 (single run) and Avg@32 (average score from 32 runs). For AIME24 (Pass@1) and other benchmarks, baselines use greedy decoding, and models with \ourmethod use temperature=1.0 and top-p=0.95. For AIME24 (Avg@32), we sample 32 responses per model with the same settings. Average scores are based on AIME (Pass@1) and other benchmarks.}
\label{table:performance}
% \vspace{-12pt}
\end{table*}

\textbf{Clip Ratio: }In the early stages of training, the base model exhibits weak instruction-following ability and often fails to stop appropriately, resulting in irrelevant or excessively long outputs. After training collapses, the model may also generate repetitive or overly extended responses. Since the model has a fixed maximum context length, such outputs may be truncated during both training and evaluation. To monitor this issue, we define the proportion of truncated outputs as the ``Clip Ratio".

\textbf{Average Stopped Length: }Generations that are truncated often result from issues such as repetitive patterns or incomplete reasoning, which typically do not contribute to effective trajectories. To account for this factor, we introduce a new metric to track the average length of responses that are stopped under normal conditions. 
% It is a more reliable metric to consider only valid responses, thereby eliminating the interference caused by unstopped responses.

For more evaluation metrics details, please refer to Appendix~\ref{appx:eval_detail}.
% Unstable training can lead to repetitive content generation that exceeds the model's context length. To measure this issue, we track the clip ratio -- the percentage of generated responses truncated due to context length limitations.

%\yh{add more detail about the def of these reasoning behavior. }
% introduce enumeration as a reasoning behavior. Leveraging the prompt in the Figure xxxx, GPT-4o~\citep{hurst2024gpt} identifies reasoning-related behaviors—"Backtracking", "Verification", "Subgoal Setting", and "Enumeration"—and we calculate their proportions.

\subsection{Main Results}

\paragraph{Zero RL Training Improves both Accuracy and Response Length Significantly:}
Figure~\ref{fig1:acc&len} and Figure~\ref{fig:appx_acc&len} in Appendix~\ref{appx:DetailedResult} illustrate a steady improvement in both response length and average accuracy across various benchmarks.  Table~\ref{table:performance} provides a detailed breakdown of the results. Despite using only 8K training samples, we observe significant performance gains for all models. The improvements cover competition-level tests like AIME 2024 and AMC 2023 for most cases. This demonstrates the remarkable generalization capabilities of zero RL training, enabling the model to effectively progress from easier to more challenging problems.
% Remarkably, even with only 8K training data for training, we observe significant performance gains across all benchmarks. Despite the limited training data, consisting solely of GSM8K and MATH 500, we observe substantial performance gains on competition-level benchmarks such as AIME 2024 and AMC 2023. This highlights the generalization abilities of zero RL training allowing the model to bridge the gap from easy to hard.
\label{sec:mistral_fail}
In addition to the Qwen series models, we also significantly improve both performance and response length for other models that initially starts with low baselines. For instance, after just 80 training iterations, the DeepSeek-Math-7B's performance increases more than threefold, while its response length grows from around 300 to over 1200 tokens. 
% We also evaluate the generalization ability across IFEVAL~\citep{zhou2023instruction}, MMLU~\citep{hendrycks2020measuring}, and GPQA-Diamond~\citep{rein2024gpqa}. As shown in Table~\ref{table:generalization_performance}, our method not only improve models' instruction-following ability but also exhibits strong generalization performance, with detailed results provided in the Appendix~\ref{sec:generalization_capability}.
% We also evaluate generalization ability across IFEVAL~\citep{zhou2023instruction}, MMLU~\citep{hendrycks2020measuring}, and GPQA-Diamond~\citep{rein2024gpqa}. As shown in Table~\ref{table:generalization_performance} of Appendix~\ref{sec:generalization_capability}, our method demonstrates strong generalization performance.

\paragraph{Zero RL Training also Demonstrates Strong Generalization Performance.}

We also evaluate the generalization ability of zero RL training using three benchmarks: IFEVAL~\citep{zhou2023instruction}, MMLU~\citep{hendrycks2020measuring}, and GPQA-Diamond~\citep{rein2024gpqa}. IFEVAL measures instruction-following capability, MMLU assesses the model's mastery of general knowledge, and GPQA-Diamond is a challenging benchmark that tests domain-specific expertise in chemistry, physics, and biology. Table~\ref{table:generalization_performance} presents the changes in model performance on IFEval, MMLU, and GPQA-Diamond before and after training. Despite zero RL training being conducted on only ~8K math reasoning-related examples, the model generalizes effectively across a range of tasks. Notably, it shows significant gains in instruction-following and general knowledge on IFEval and MMLU, as well as substantial improvements on the challenging GPQA-Diamond benchmark, which spans chemistry, physics, and biology.

\begin{table*}[t]
\centering
\resizebox{0.75\textwidth}{!}{
\begin{tabular}{lccccc}
\toprule
\multicolumn{1}{c}{\textbf{Model}} & \textbf{\begin{tabular}[c]{@{}c@{}}IFEVAL\\ strict-prompt\end{tabular}} & \textbf{\begin{tabular}[c]{@{}c@{}}MMLU\\Stem\end{tabular}} & \textbf{MMLU} & \textbf{GPQA} & \textbf{Avg.} \\ \midrule
\multicolumn{6}{c}{\textit{Llama, DeepSeek and Mistral Models}} \\
Mistral-v0.1-7B & 13.5 & 26.1 & 28.0 & 23.2 & 22.7 \\
\rowcolor[rgb]{ .867, .922, .969}~~~~$\hookrightarrow$~+~\ourmethod & 21.8 & 28.1 & 34.6 & 30.3 & 28.7 \\
Llama-3.1-8B & 16.1 & 27.1 & 28.7 & 22.7 & 23.6 \\
\rowcolor[rgb]{ .867, .922, .969}~~~~$\hookrightarrow$~+~\ourmethod & 25.1 & 40.7 & 44.5 & 20.2 & 32.6 \\
DeepSeek-Math-7B & 11.5 & 21.6 & 22.7 & 19.2 & 18.7 \\
\rowcolor[rgb]{ .867, .922, .969}~~~~$\hookrightarrow$~+~\ourmethod & 16.3 & 47.4 & 45.5 & 27.3 & 34.1 \\
Mistral-Small-24B & 17.4 & 30.9 & 31.7 & 20.2 & 25.0 \\
\rowcolor[rgb]{ .867, .922, .969}~~~~$\hookrightarrow$~+~\ourmethod & 23.5 & 73.9 & 78.8 & 45.0 & 55.3 \\ \midrule
\multicolumn{6}{c}{\textit{Qwen Series Models}} \\
Qwen-2.5-0.5B & 9.6 & 23.2 & 24.9 & 24.8 & 20.6 \\
\rowcolor[rgb]{ .867, .922, .969}~~~~$\hookrightarrow$~+~\ourmethod & 14.4 & 32.1 & 34.6 & 26.3 & 26.8 \\
Qwen-2.5-1.5B & 15.2 & 33.1 & 35.4 & 24.8 & 27.1 \\
\rowcolor[rgb]{ .867, .922, .969}~~~~$\hookrightarrow$~+~\ourmethod & 20.3 & 42.1 & 45.2 & 28.8 & 34.1 \\
Qwen-2.5-7B & 21.3 & 39.8 & 38.6 & 23.7 & 30.8 \\
\rowcolor[rgb]{ .867, .922, .969}~~~~$\hookrightarrow$~+~\ourmethod & 25.9 & 49.6 & 47.0 & 29.8 & 38.1 \\
Qwen-2.5-Math-7B & 14.1 & 40.6 & 38.0 & 27.8 & 30.1 \\
\rowcolor[rgb]{ .867, .922, .969}~~~~$\hookrightarrow$~+~\ourmethod & 17.0 & 55.6 & 56.6 & 35.4 & 41.1 \\
Qwen-2.5-14B & 22.9 & 59.8 & 63.5 & 24.8 & 42.7 \\
\rowcolor[rgb]{ .867, .922, .969}~~~~$\hookrightarrow$~+~\ourmethod & 29.4 & 76.3 & 79.1 & 50.0 & 58.7 \\
Qwen-2.5-32B & 24.6 & 60.7 & 62.7 & 38.9 & 46.7 \\
\rowcolor[rgb]{ .867, .922, .969}~~~~$\hookrightarrow$~+~\ourmethod & 31.2 & 79.0 & 82.5 & 49.5 & 60.6 \\ \bottomrule
\end{tabular}

}
\caption{Detailed performance of various models across IFEVAL, MMLU and GPQA. The blue lines represent the models trained with our recipe.}
\label{table:generalization_performance}
% \vspace{-12pt}
\end{table*}

\begin{figure}[!t]
        \centering
\includegraphics[width=0.97\columnwidth]{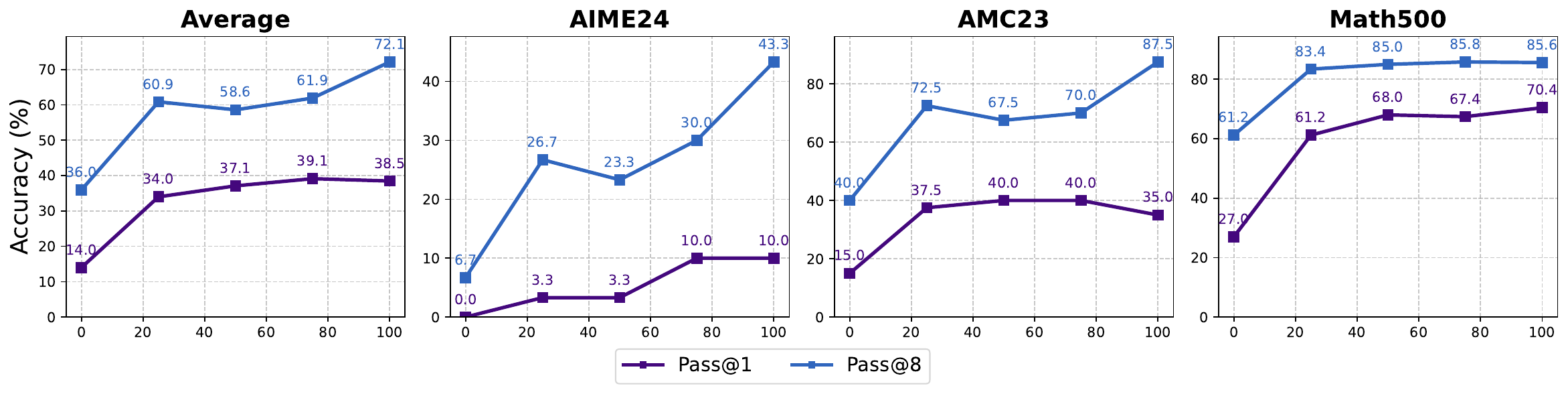}\vspace{-10pt}
\caption{Pass@1 and Pass@8 accuracy over the training iterations of Mistral-Small-24B. The model is trained on the hard data (MATH levels 3–5) as described in \S\ref{sec:setup}. We evaluate its performance on three benchmarks: AIME24, AMC23, and Math500. The reported average score is the mean across these three benchmarks.
        }
        %\vspace{-12pt}
        \label{fig3:passk-mistral}
\end{figure}

\label{sec:lift_pass_k}
\paragraph{Steady Improvement of Pass@k Accuracy:}
% \begin{wrapfigure}{r}{.45\textwidth}
%     \centering
%     \includegraphics[width=\linewidth]{fig/plot_figure7_v1_pass_at_k_avg.pdf}\vspace{-10pt}
%     \caption{Pass@k of Mistral-24B based on the average results from AIME24 and AMC23. }
%     \label{fig:different_passk}
%     % \vspace{-10pt}
% \end{wrapfigure}

\begin{wrapfigure}{r}{.45\textwidth}
    \centering
    \includegraphics[width=\linewidth]{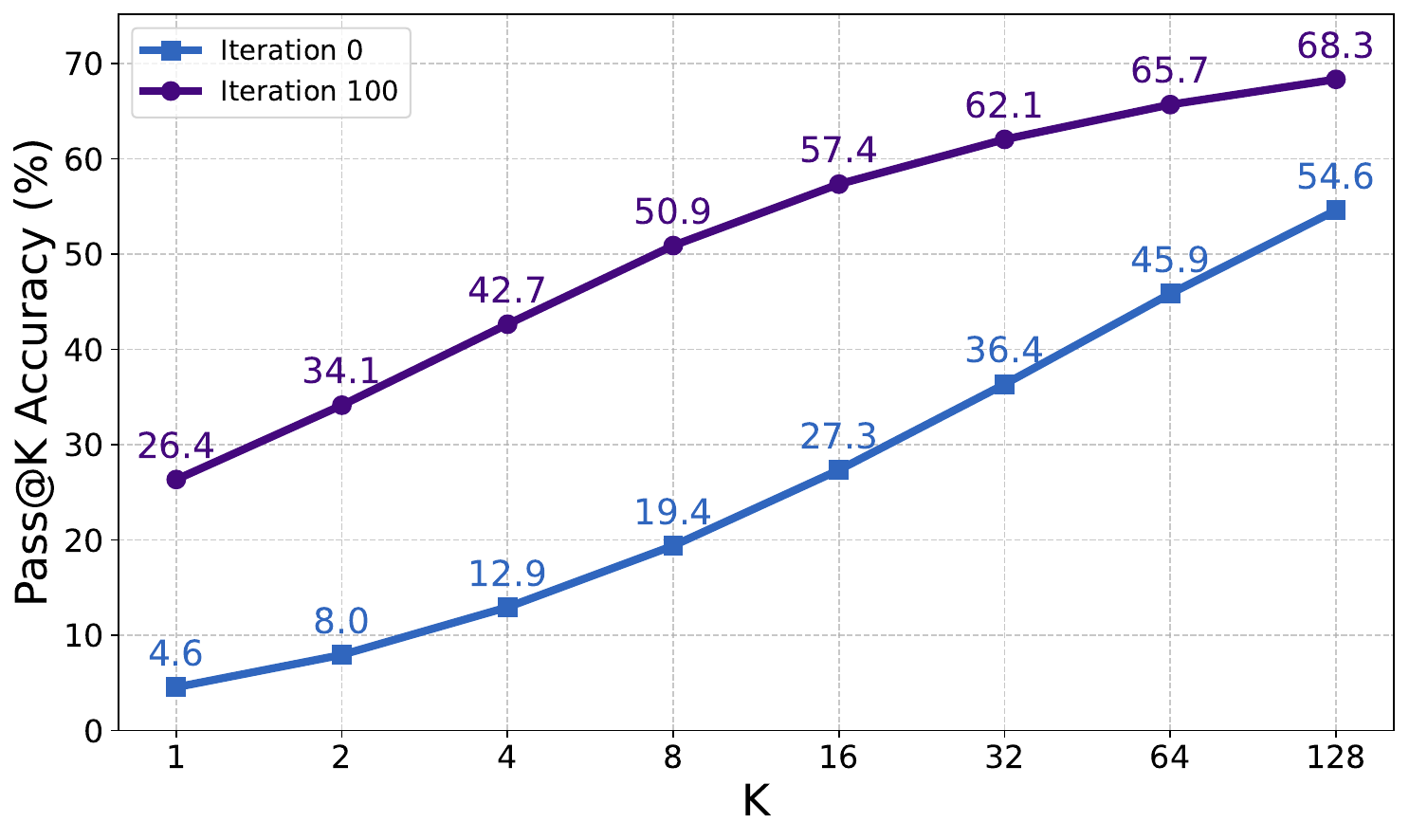}\vspace{-10pt}
    \caption{Pass@k of Mistral-24B based on the average results from AIME24 and AMC23. }
    \label{fig:different_passk}
    % \vspace{-10pt}
\end{wrapfigure}

As shown in Figure~\ref{fig3:passk-mistral}, Mistral-Small-24B exhibits robust growth in pass@8. Furthermore, as training progresses, the model's pass@1 results eventually surpass the initial pass@8 results of the base model. 
By iteration 100, the two metrics differ by more than 30 absolute points on average. This suggests significant potential for further improvements in RL, as our training rolls out 8 responses for each query and pass@8 represents the model's ability to explore correct responses. Surprisingly, the gap between pass@1 and pass@8 does not diminish during training; instead, it widens as training progresses. 
Figure~\ref{fig:different_passk} shows that a significant gap in pass@k persists between the base model and the model after RL training, even at higher values of k -- the gap is from 13 to 30 absolute points when we vary k up to 128. 
% Notably, after just 100 training iterations, the model achieves a pass@1 performance comparable to the base model’s pass@16.
This suggests that zero RL training is not just reranking the model’s output distribution within the top k candidates at a reasonably large range of k~\citep{shao2024deepseekmath}, instead, it enhances the model’s fundamental reasoning abilities.

\begin{figure}[!t]
        \centering
\includegraphics[width=0.97\columnwidth]{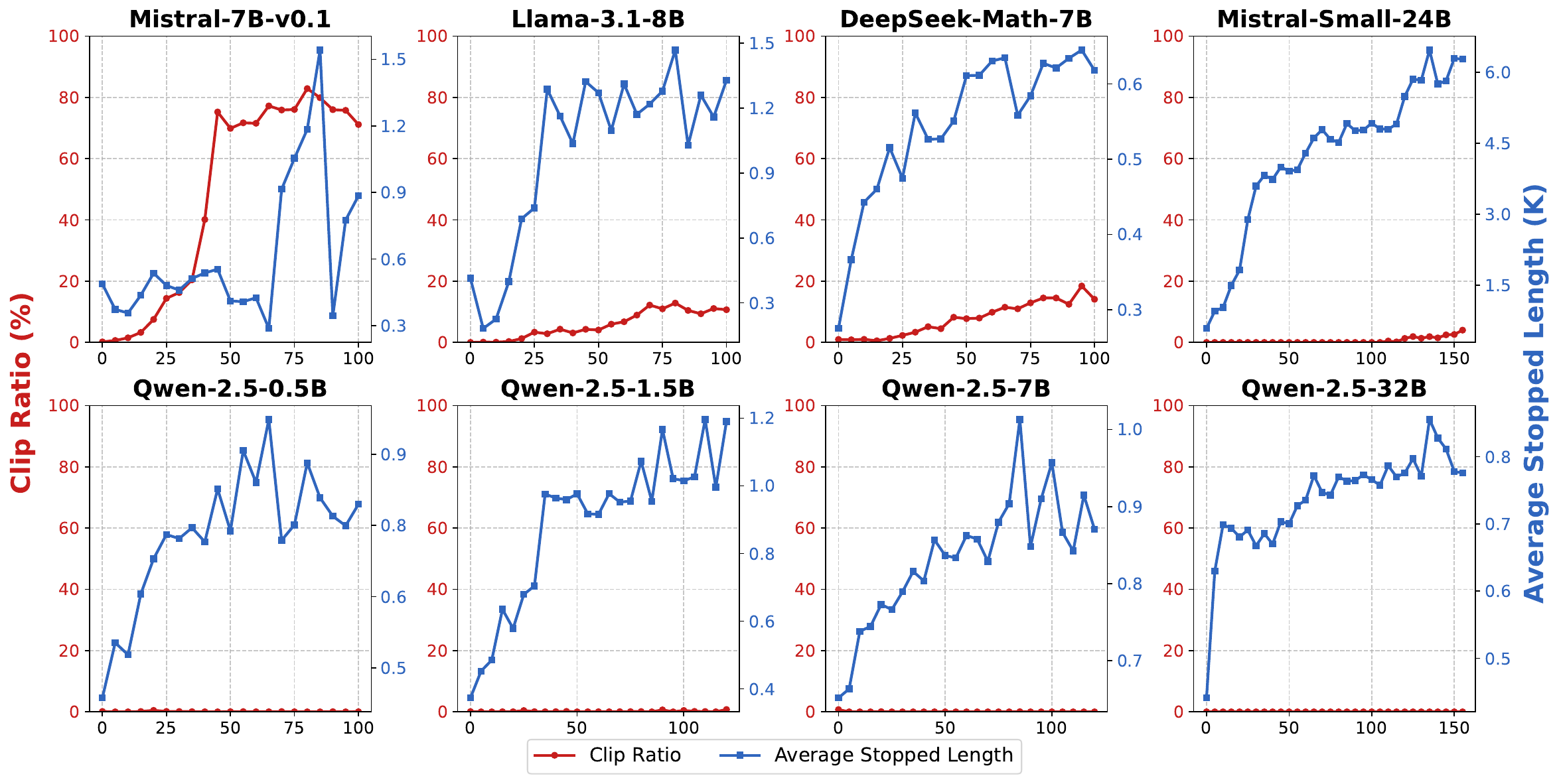}
\vspace{-10pt}
\caption{Average clip ratio and stopped length across training iterations for different models. We assess the models every five steps on a variety of math benchmarks, including GSM8K, MATH500, Minerva Math, and OlympiadBench, as well as competition-level benchmarks like AIME24 and AMC23. The red line indicates the clip ratio, while the blue line represents the stopped length. Per-benchmark results are in Figure~\ref{fig:appx_clip&stop} (Appendix~\ref{appx:DetailedResult}).}
        \label{fig2:clip&stop}
        \vspace{-10pt}
\end{figure}

\paragraph{Growth in Response Length May be Unhealthy:}Response length does not always reflect genuine growth in reasoning. In some cases, unstable training can cause models to generate excessive repetitive content until they hit the context length limit, artificially inflating response length without improving reasoning depth.  For example, Figure~\ref{fig2:clip&stop} shows that while most models maintain a low clip ratio -- below 5\% of the data -- when their average stopping length steadily increases, Mistral-7B-v0.1 exhibits a high clip ratio and significant fluctuations in stopping length. Upon closer inspection of its responses, we find that the responses consist of incoherent, mixed-language gibberish, suggesting that its thinking process is not genuinely expanding.
We note that such patterns would not be captured by response length as in Figure~\ref{fig1:acc&len}.
These findings indicate that most models demonstrate a meaningful increase in response length. This raises an important question: What exactly do models learn as their thinking time increases? We answer this question next.

\subsection{The ``Aha Moment'' -- Quantifying Emergence of Reasoning Behaviors} 
\begin{figure}[!t]
        \centering
\includegraphics[width=\columnwidth]{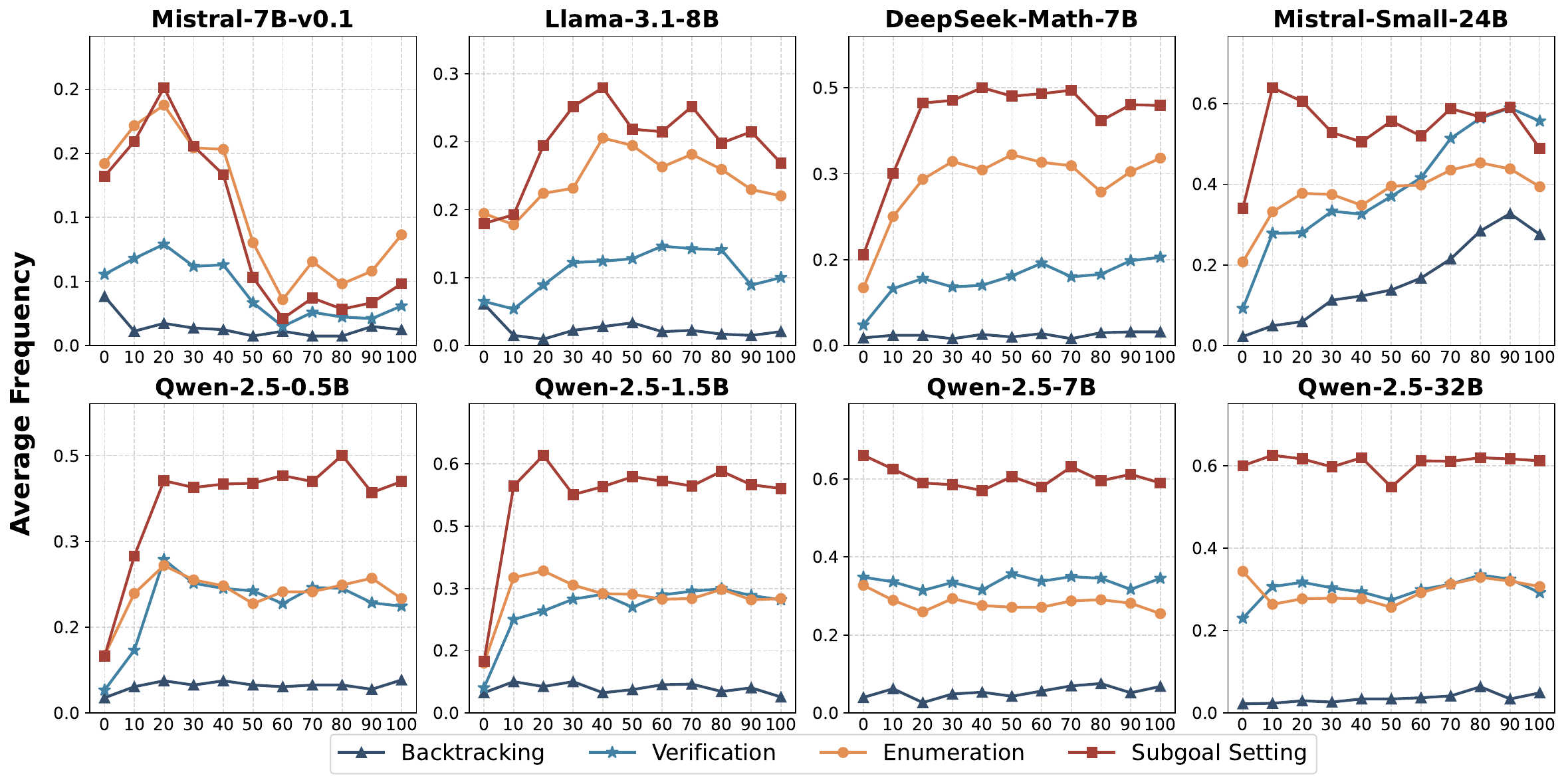}
%\vspace{-10pt}
\caption{The change in reasoning behavior over the training iterations across all models. As described in \S\ref{sec:eval_metrics}, we use GPT-4o to extract and track shifts in reasoning behaviors on OlympiadBench. We focus on four reasoning-related behaviors: ``Backtracking", ``Verification", ``Subgoal Setting", and ``Enumeration".
        }
        \label{fig4:behavior&counts}
    %\vspace{-15pt}
\end{figure}

% \label{sec:behavior}
Figure~\ref{fig4:behavior&counts} illustrates the reasoning behavior ratio on OlympiadBench during model training.
%, while results for other benchmarks are provided in the Appendix \jh{XXXX}. 
By comparing Figure~\ref{fig4:behavior&counts} with Figure~\ref{fig2:clip&stop}, we observe that fluctuations in the reasoning behavior ratio effectively account for variations in the average stopped length. Interestingly, we find that different models exhibit entirely distinct trends in reasoning behavior changes.

% \begin{figure}[!t]
%         \centering
% \includegraphics[width=\columnwidth]{fig/plot_figure4_v4_behavior_lines.pdf}\vspace{-10pt}
% \caption{The change in reasoning behavior over the training iterations across all models. As described in \S\ref{sec:eval_metrics}, we use GPT-4o to extract and track shifts in reasoning behaviors on OlympiadBench. We focus on four reasoning-related behaviors: ``Backtracking", ``Verification", ``Subgoal Setting", and ``Enumeration".
%         }
%         \label{fig4:behavior&counts}
%     \vspace{-15pt}
% \end{figure}

Smaller models, such as Qwen-2.5-0.5B and 1.5B, tend to prioritize learning the "Subgoal Setting" behavior, with its proportion increasing by approximately 4–5 times. Additionally, the proportions of "Verification" and "Enumeration" also show noticeable growth. In contrast, for other base models that inherently possess step-by-step reasoning capabilities, adjustments in "Subgoal Setting" during the RL training process are relatively minor.

% \wz{Based on recent experiments, models with different capabilities focus on learning different behaviors. For example, Qwen 2.5 0.5b prioritizes learning "subgoal setting", while more powerful models tend to learn high level patterns such as "verification" and "backtracking". will add this insteresting finding later.}
% Our analysis indicates that most base models inherently exhibit step-by-step reasoning capabilities. As a result, during the RL training process, adjustments in ``Subgoal Setting" remain relatively subtle.
\label{sec:deepseek-math_behabiur}
DeepSeek-Math-7B, Llama-3.1-8B, and Mistral-Small-24B exhibit substantial increases in the proportions of ``Enumeration" and ``Verification" behaviors, rising from relatively low initial levels by approximately 3-4 times. This growth correlates closely with their changes in average stopped length, suggesting a shift in reasoning patterns over time. For instance, in Mistral-Small-24B, reflection-oriented behaviors such as ``Verification" and ``Backtracking" increase dramatically from nearly 0\% to approximately 50\%, indicating the emergence of reflection behavior from scratch. This shift suggests that the model progressively internalizes verification as part of its reasoning process, offering a promising trajectory for enhancement.

\label{sec:qwen_behabiur}
In contrast, Qwen-2.5-7B and 32B demonstrate strong reasoning behaviors from the outset, with minimal changes throughout training. This phenomenon aligns with their slow length adjustments (Figure~\ref{fig1:acc&len}) and suggests that Qwen models inherently possess robust reasoning capabilities. Rather than undergoing a structural shift in their reasoning processes, they primarily benefit from small increases in thinking time, which yield significant performance improvements. Finally, we observe that Mistral-7B-v0.1 consistently exhibits low reasoning behaviors with no noticeable growth, further supporting our earlier analysis in \textsection\ref{sec:mistral_fail}.

% \begin{figure}[!t]
%         \centering
% \includegraphics[width=\columnwidth]{fig/mistral24b_verification.pdf}\vspace{-10pt}
% \caption{A comparison of Mistral-24B's "verification" and "backtraining" behavior before and after "zero training." Here, "base solution" represents the response of the Mistral-24B base model, while "zero solution" represents the response of the model after training.\yh{move to appx}
%         }
%         \label{fig7:verfication_case}
%     \vspace{-10pt}
% \end{figure}

% \begin{figure}[!t]
%         \centering
% \includegraphics[width=\columnwidth]{fig/mistral24b_enumeration_case.png}
% \caption{A comparison of Mistral-24B's "Enumeration" behavior before and after "zero training." Here, "base solution" represents the response of the Mistral-24B base model, while "zero solution" represents the response of the model after training.\jh{this example is not good and does not look like an enumeration case, maybe just remove it from the paper.}
%         }
%         \label{fig8:enumeration_case}
%     \vspace{-10pt}
% \end{figure}
% \wz{plan to add some cases on model like Mistral-24B}

To intuitively illustrate the changes in reasoning behavior, we present examples of Mistral 24B's reasoning before and after training in Figures~\ref{fig7:verfication_case}. Comprehensive case studies involving other models are available in Appendix~\ref{sec:other_model_behaviour}. In Figure~\ref{fig7:verfication_case}, we observe that unlike the base model, the zero training model actively attempts to verify if its initial solution is valid by substituting it back into the original expression. Upon recognizing that the first solution does not meet the necessary conditions, the model explicitly initiates a backtracking approach, stating "let's try another possibility," eventually arriving at the correct answer.

%\jh{does not look like an enumeration example}In Figure~\ref{fig8:enumeration_case}, we highlight a significant improvement in the current model's ability to enumerate possibilities compared to the base model. The zero training model demonstrates a clearer understanding of the required steps: it first calculates permutations across different groups, then permutations within each group, and finally combines these results through multiplication, correctly obtaining the final answer.

% \wz{add more cases in appendix}

% use more metric to analyse some reflection pattern, maybe some case in appendix

%\section{Dynamics of Emergence in Zero Training}

%  \begin{wrapfigure}{r}{0.48\columnwidth}
%     \centering
%     \includegraphics[width=\linewidth]{fig/plot_figure5_v2_combined_format_reward.pdf}\vspace{-10pt}
%     \caption{Comparison of accuracy and response length with and without format rewards.}
%     \label{fig:w&wo_format_reward}
%     \vspace{-20pt}
% \end{wrapfigure}

\section{Key Factors Shaping Zero RL Training}
In this section, we identify key factors that influence stability and performance during zero RL training, particularly when dealing with early-stage or weaker models.  
First, we explore how an over-reliance on format rewards restricts exploration. Next, we analyze how data difficulty impacts exploratory behavior, illustrating how exposure to varying levels of difficulty shapes the exploration dynamics. We also discuss the impact of exploration-related hyperparameters in Appendix~\ref{sec:impact_explore_hyper}.

% 基本上可以照着做不work的角度走了
% First, they often struggle to accurately interpret complex instructions, such as formatting answers within specific boxes. Second, many base models tend to generate sequences continuously, failing to stop once the question has been answered. These issues lead to a high volume of irrelevant outputs in the early stages of training, contributing to model collapse, where the models get stuck in repetitive, meaningless loops.  To address this, we systematically analyze the impact of such behavior and propose straightforward yet effective solutions to mitigate these problems. \wz{This content is more about tricks, it would be better to move it to the appendix or remove it}

\subsection{Over-Reliance on Format Rewards}

We find that enforcing strict formatting constraints, such as requiring the final answer to be enclosed in a latex command \textit{\textbackslash boxed\{\}}, can hinder model's freely exploration and ultimately degrades performance. This is because many base models cannot follow the format constraint well in the initial stage, and imposing a format reward will penalize many correct explorations. We compare two reward functions: one without format constraints, which rewards responses solely based on answer correctness (our default design in \textsection\ref{sec:reward_remove}), and another that strictly enforces formatting by penalizing responses with a reward of -1 if they fail to adhere to the required format.

%  \begin{wrapfigure}{r}{0.48\columnwidth}
%     \centering
%     \includegraphics[width=\linewidth]{fig/plot_figure5_v2_combined_format_reward.pdf}\vspace{-10pt}
%     \caption{Comparison of accuracy and response length with and without format rewards.}
%     \label{fig:w&wo_format_reward}
%     \vspace{-20pt}
% \end{wrapfigure}

\label{sec:remove_format}
 \begin{wrapfigure}{r}{0.48\columnwidth}
    \centering
    \includegraphics[width=\linewidth]{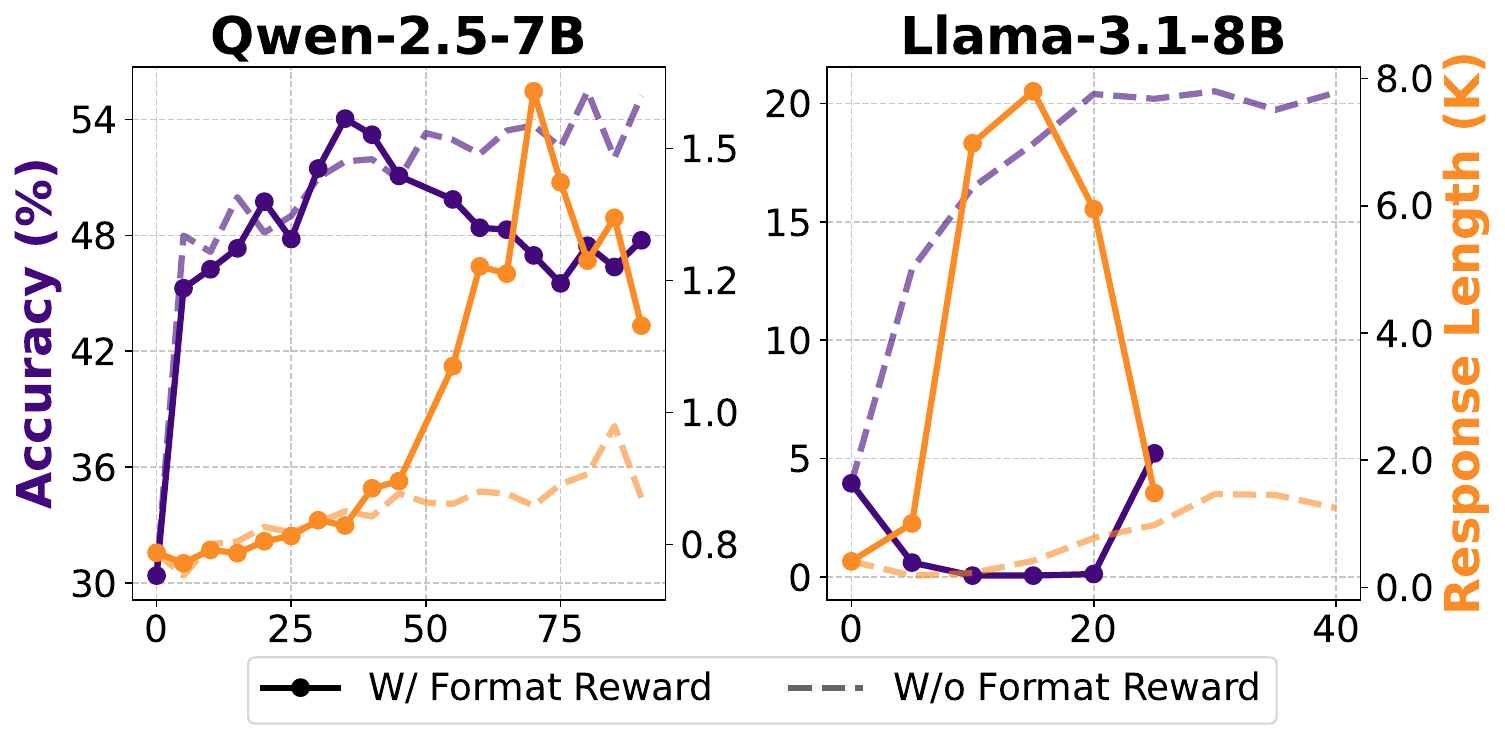}\vspace{-10pt}
    \caption{Accuracy and response length with and without format rewards.}
    \label{fig:w&wo_format_reward}
    \vspace{-5pt}
\end{wrapfigure}

Figure~\ref{fig:w&wo_format_reward} illustrates weaker models like Llama-3.1-8B struggle under strict formatting requirements, leading to a rapid increase in response length early in training without performancec improvement. The model expends excessive effort on adhering to the format but fails to learn how to answer correctly, ultimately resulting in model collapse. 
Figure~\ref{fig:w&wo_format_reward} (Left) further reveals that even stronger models, such as Qwen-2.5-7B, which initially comply with formatting constraints, suffer in later training stages. This includes both performance degradation and a significant reduction in CoT length.  
%  \begin{wrapfigure}{r}{0.48\columnwidth}
%     \centering
%     \includegraphics[width=\linewidth]{fig/plot_figure5_v2_combined_format_reward.pdf}\vspace{-10pt}
%     \caption{Comparison of accuracy and response length with and without format rewards.}
%     \label{fig:w&wo_format_reward}
%     \vspace{-20pt}
% \end{wrapfigure}
% \wz{Here the total length with format reward is much longer, I'm wondering if we need to explain this using other metrics}
These findings highlight that: in a zero RL training setting, rather than imposing rigid formatting rules, we should prioritize maintaining response verifiability while allowing sufficient flexibility for exploration.

% 影响性能上限和reasonng behaviur的涌现
\begin{figure*}[!t]
    \centering
    % \subfigure[Pass@1 on GSM8K]{
    %     \includegraphics[width=0.23\textwidth]{fig/gsm8k_pass@1_math_only_v2.pdf}
    % }
    \subfigure[Mistral-7b-v0.1]{
        \includegraphics[width=0.47\textwidth]{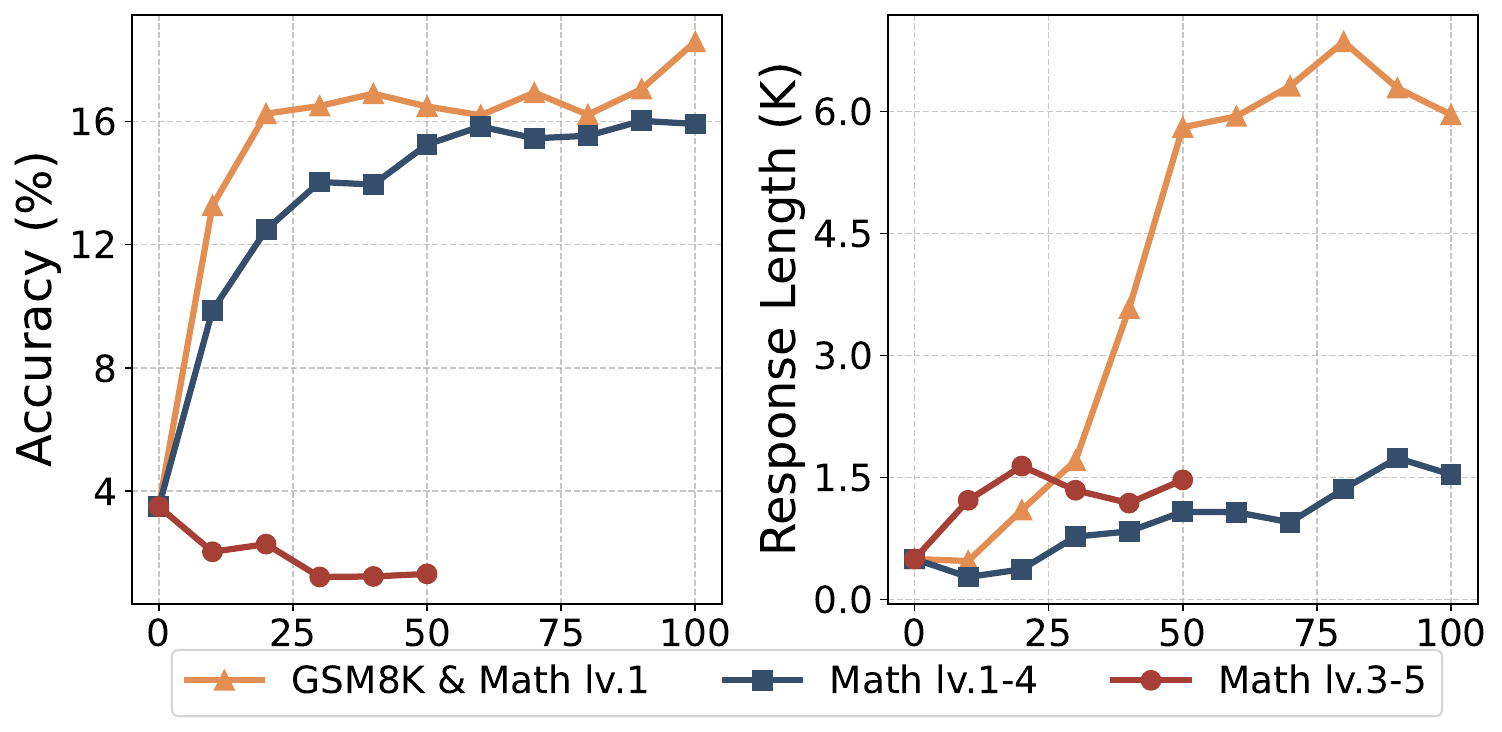}
        \label{fig:mathlv-mistral-7b-v0.1}
    }
    \subfigure[Qwen-2.5-7B]{
        \includegraphics[width=0.47\textwidth]{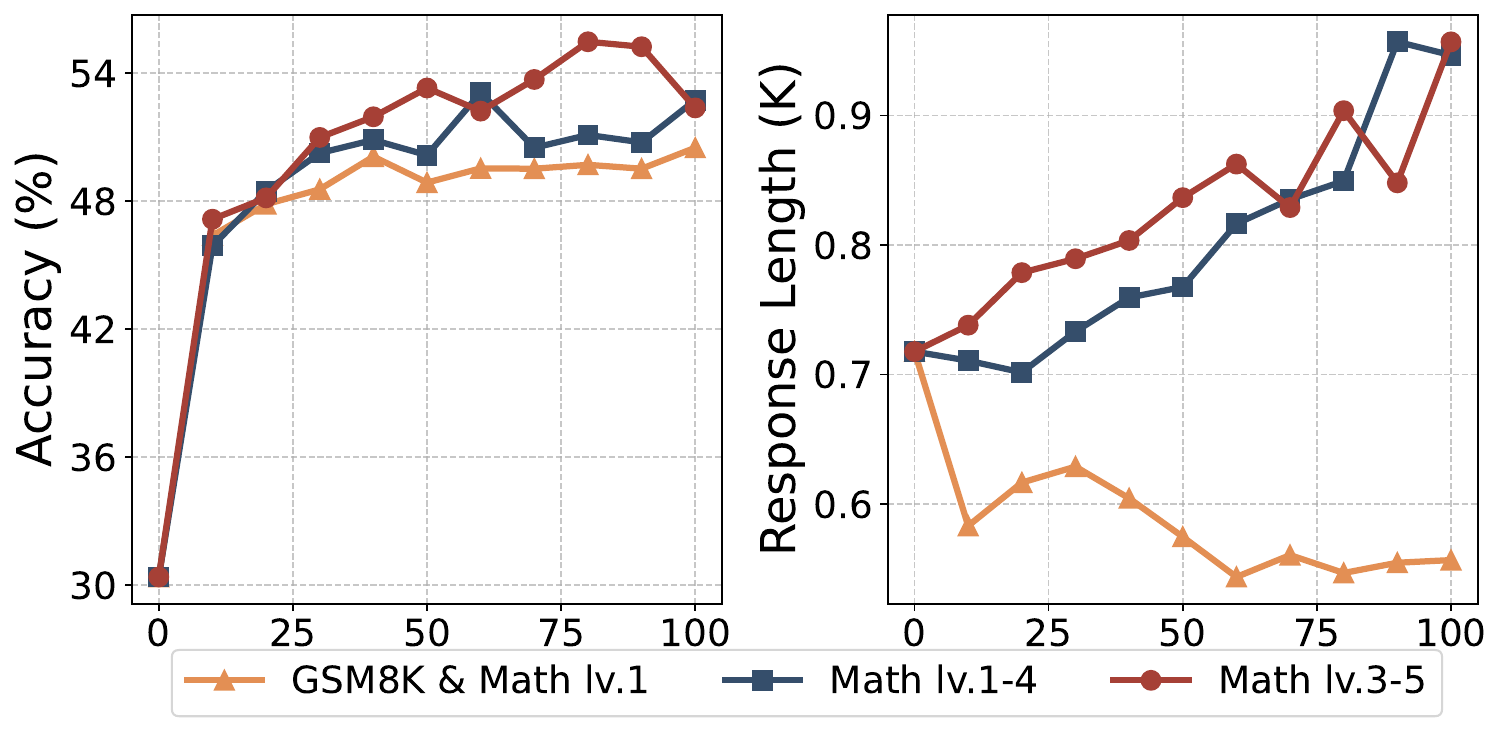}
        \label{fig:mathlv-qwen-2.5-7b}
    }
    %\vspace{-10pt}
    \caption{Comparison of accuracy and response length across different data difficulty levels. We examine three levels of data: Easy (GSM8K and MATH lv.1), Medium (MATH lv.1–4), and Hard (MATH lv.3–5), with each category containing approximately 8,000 problems.}
    \label{fig:mathlv}
    %\vspace{-15pt}
\end{figure*}
\subsection{Data Difficulty on Exploratory Behavior}
\label{sec:data_complextiy_behaviur}
Base models exhibit varying performance and CoT behaviors when trained on different RL data. Figure~\ref{fig:mathlv} compare the performance of Mistral-7B and Qwen-2.5-7B across Easy, Medium, and Hard datasets. We observe a clear trend: as data difficulty increases, Mistral-7B's performance progressively deteriorates. When faced with high-difficulty data (Hard: MATH levels 3-5), the model struggles to generate responses that receive positive feedback from the reward system. This failure results in a significant increase in response length without any corresponding improvement in accuracy, signaling a breakdown in the training process—often referred to as training collapse. Figure~\ref{fig:mathlv} Left demonstrates that Qwen-2.5-7B exhibits a pattern entirely opposite to Mistral-7B-v0.1. Specifically, as dataset difficulty decreases, both the model’s average accuracy and response length decline, with the effect being most pronounced on the simplest dataset, where even response length decreases. This finding aligns with our previous analysis of Qwen-2.5-7B in \textsection\ref{sec:qwen_behabiur}, reinforcing the notion that Qwen inherently possesses strong reasoning capabilities. To further improve its response length, training should incorporate more challenging datasets to encourage deeper reasoning and extended thinking time.
The analysis highlights that zero RL training data must align with the base model's inherent reasoning capabilities.

\section{Revisiting Traditional SFT for RL-Driven Reasoning Emergence}

\label{sec:short_cot_influence}

\begin{figure}[!t]
        \centering
\includegraphics[width=0.97\columnwidth]{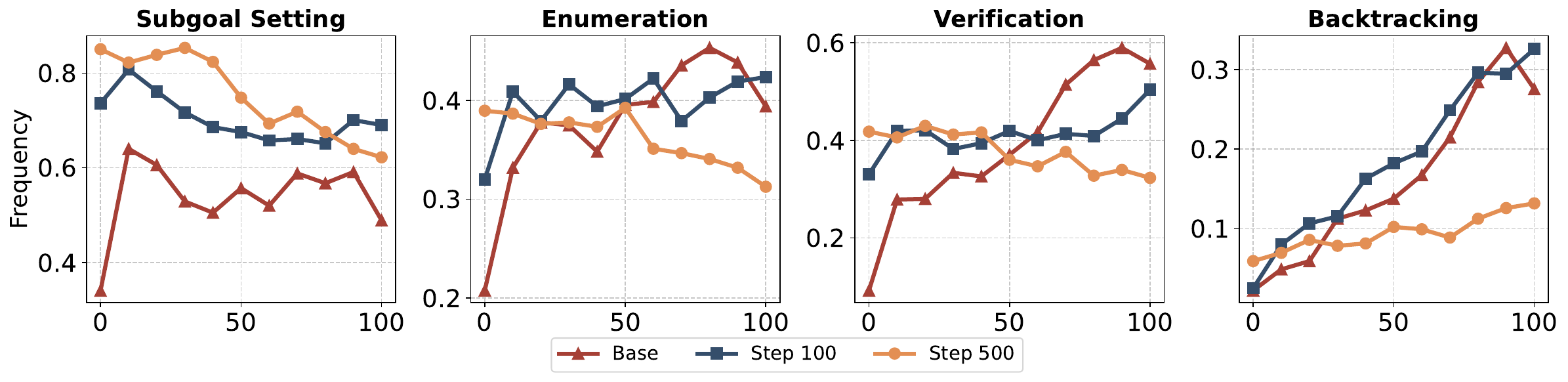}\vspace{-10pt}
\caption{Reasoning behavior ratio over RL training iterations after using different SFT steps as starting points. ``Base" refers to the base Mistral-Small-24B model without any SFT, while ``Step 100" and ``Step 500" represent 100 and 500 steps of SFT, respectively. As described in \S\ref{sec:setup}, we use GPT-4o to track shifts in reasoning behaviors on OlympiadBench.}
        %\vspace{-15pt}
         \label{fig:sft_behaviur}
\end{figure}

As base models may not follow instruction well and pose a major challenge for zero RL training, one may wonder a simple SFT stage as a cold start may be helpful to learn to follow instructions well. In this section, we revisit the impact of traditional SFT methods (where the responses are not from long CoT models) as a cold start on RL training performance and reasoning behavior--notably, this was the most commonly used post-training pipeline with RL following an SFT stage, before DeepSeek-R1. Specifically, we use a subset of the NuminaMath~\citep{li2024numinamath} dataset derived from GSM8K and MATH,~\footnote{We also conduct experiments using general SFT dataset beyond math-related ones, which can be found in Appendix~\ref{sec:general_sft} and implies similar conclusion.} containing approximately 15K high-quality short CoT responses. We conduct SFT using Mistral 24B and select models at 100 and 500 training steps as starting points for RL training. 
% \begin{wrapfigure}{r}{0.48\columnwidth}
%     \centering
%     \includegraphics[width=\linewidth]{fig/plot_figure9_v2_diff_n-mistral-24b.pdf}
%     %\vspace{-10pt}
%     \caption{Accuracy and response length averaged on the six benchmarks over RL training iterations after running different SFT steps as starting points. }
%     \label{fig:sft_accuracy_length} 
%     %\vspace{-10pt}
% \end{wrapfigure}

\begin{wrapfigure}{r}{0.48\columnwidth}
    \centering
    \includegraphics[width=\linewidth]{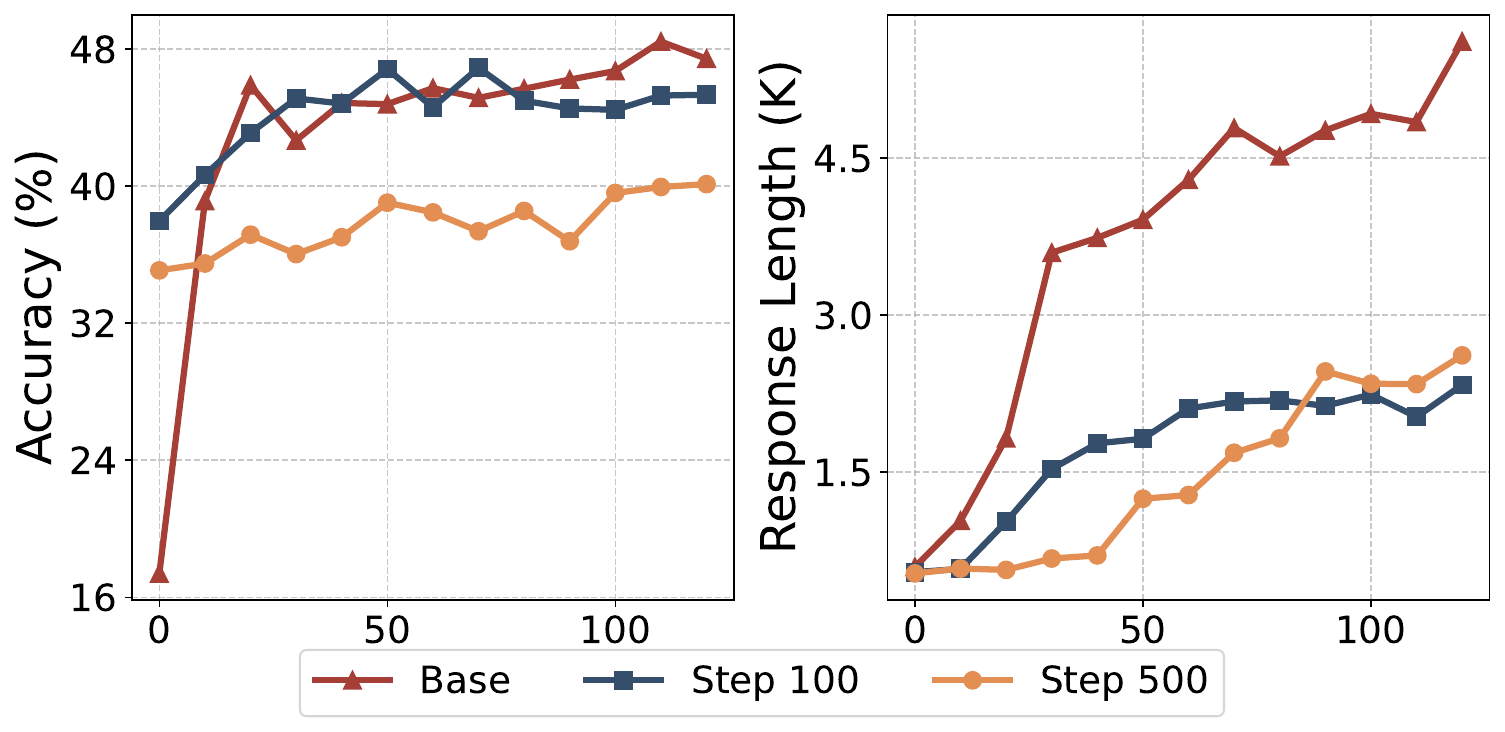}
    %\vspace{-10pt}
    \caption{Accuracy and response length averaged on the six benchmarks over RL training iterations after running different SFT steps as starting points. }
    \label{fig:sft_accuracy_length} 
    %\vspace{-10pt}
\end{wrapfigure}

Figure~\ref{fig:sft_accuracy_length} illustrates how model accuracy and output length evolve during RL training when different initial models are used. Our results indicate that starting from SFT models initially boosts performance significantly; however, these models encounter notable limitations in their maximum achievable accuracy and response length compared to starting from the base model during RL training. Crucially, we observe that these limitations become increasingly pronounced as the number of initial SFT steps grows.

%For instance, while the base model can attain a pass@1 accuracy of approximately 49.6\% during RL training, models initialized with 100 and 500 SFT steps achieve maximum accuracies of only about 47.3\% and 40.3\%, respectively.

%For instance, \jh{need to cite the table here}while the base model can attain a pass@1 accuracy of approximately \jhc{72}{72\%} on MATH-500 during RL training, models initialized with 100 and 500 SFT steps achieve maximum accuracies of only about \jhc{68}{68\%} and \jhc{59}{59\%}, respectively.

To further investigate how initial SFT affects the emergence of reasoning behaviors, we analyze how often specific reasoning behaviors appeared during training at different starting points, as shown in Figure~\ref{fig:sft_behaviur}. Our analysis reveals that initial SFT negatively impacts the development of critical reasoning behaviors. Specifically, models with 100 SFT steps exhibit reduced upper limits in essential reasoning behaviors such as "enumeration," "verification," and "backtracking," compared to the base model. Even more notably, models with 500 SFT steps experience significant declines in "enumeration" and "verification" behaviors in later training stages, highlighting a detrimental long-term effect of extensive sft on reasoning capabilities.
% \begin{wrapfigure}{r}{0.48\columnwidth}
%     \centering
%     \includegraphics[width=\linewidth]{fig/plot_figure9_v2_diff_n-mistral-24b.pdf}\vspace{-10pt}
%     \caption{Accuracy and response length averaged on the six benchmarks over RL training iterations after running different SFT steps as starting points. ``Base" refers to the base Mistral-Small-24B model without any SFT, while ``Step 100" and ``Step 500" represent 100 and 500 steps of SFT on the base model, respectively.}
%     \label{fig:sft_accuracy_length}
% \end{wrapfigure}
This prompts a reconsideration of whether traditional SFT inherently restricts model exploration, perhaps highlighting the need for future cold-start strategies to prioritize exploration capacity—whether by incorporating long CoT data~\citep{guo2025deepseek,yeo2025demystifying} or designing SFT techniques~\citep{li2025preserving} that strike a balance between imitation and exploration—to enable sustained improvements in model reasoning performance.

\section{Conclusion}
Our paper demonstrates the effectiveness of zero RL training across a diverse range of base models, yielding significant improvements in accuracy and response length. We provide strong evidence that zero RL training is not merely reranking, but rather a genuine enhancement. Furthermore, we identify key factors such as reward design, data difficulty, and models' inherent abilities that shape the emergence of advanced reasoning behaviors. Our findings also indicate that starting RL training from models with traditional SFT may limit the development of advanced reasoning behaviors. Overall, our work highlights key factors for effective zero RL training and offers insights for future model improvements.

% \section*{Author Contributions}

% \section*{Acknowledgments}
% Use unnumbered first level headings for the acknowledgments. All

% \section*{Ethics Statement}
% Authors can add an optional ethics statement to the paper. 
% For papers that touch on ethical issues, this section will be evaluated as part of the review process. The ethics statement should come at the end of the paper. It does not count toward the page limit, but should not be more than 1 page. 

\bibliography{colm2025_conference}
\bibliographystyle{colm2025_conference}
\newpage

\appendix

\section{Detailed Background: ``Zero RL Training"}
\label{sec:detailed_grpo}
% DeepSeek-R1-Zero~\citep{guo2025deepseek} demonstrates that reasoning capabilities can be effectively developed through large-scale reinforcement learning (RL), even without supervised fine-tuning (SFT) as an initial step. We follow this ``zero RL training" approach, conduct our experiments on various open base models 
In our study, we follow the zero RL training recipe in ~\citet{guo2025deepseek} using various open base models, employing the GRPO algorithm~\citep{shao2024deepseekmath}. Here, zero RL training refers to reinforcement learning directly from the base model without any prior supervised fine-tuning (SFT).
%Additional experiments exploring alternative algorithms are included in the Appendix xxxx. 
GRPO optimizes computational efficiency by eliminating the need for a separate value model; instead, it directly utilizes group-normalized rewards to estimate advantages. For a query \( q \) and a set of responses \( O = \{o_1, o_2, \dots, o_G\} \) sampled from the old policy model \( \pi_{\text{old}} \), we adopt a token-level, length-rectified GRPO objective to optimize the policy model \( \pi \):\footnote{The original GRPO objective has a length normalization term that introduces length biases. We remove the length normalization term similar to concurrent works~\citep{yu2025dapoopensourcellmreinforcement,liu2025understanding} -- this length-rectified objective was the default implementation of GRPO in our adapted codebase, verl~\citep{sheng2024hybridflow}.}
% \begin{align}
% \mathcal{J}_{GRPO}(\theta) &= \mathbb{E}[q \sim P(Q), \{ o_i \}_{i=1}^G \sim \pi_{\theta_{\text{old}}}(O|q)] \\
% &= \frac{1}{G} \sum_{i=1}^{G} \frac{1}{|o_i|} \sum_{t=1}^{|o_i|} \left\{ \min \left[ \frac{\pi_{\theta}(o_{i,t} | q, o_{i,<t})}{\pi_{\theta_{\text{old}}}(o_{i,t} | q, o_{i,<t})} \hat{A}_{i}, 1 - \epsilon, 1 + \epsilon \right] \right\} - \beta \mathbb{D}_{KL} [\pi_{\theta} || \pi_{\text{ref}}]
% \end{align}

% \begin{equation}
%     \begin{split}
%         \mathcal{J}_{\text{GRPO}}(\theta) 
%         = & \underset{\text{Clipped policy update}}{\underbrace{\frac{1}{\sum_{i=1}^{G}|o_{i}|} \sum_{i=1}^G\sum_{t=1}^{|o_i|} \min\left[ r_{i,t}(\theta) \hat{A}_i,\, \operatorname{clip}\left(r_{i,t}(\theta); 1-\epsilon, 1+\epsilon\right) \hat{A}_i \right]}} \\
%         & - \underset{\text{KL penalty}}{\underbrace{\beta \, \mathbb{D}_{\text{KL}}[\pi_\theta \parallel \pi_{\text{ref}}]}}
%     \end{split}
% \end{equation}

\begin{equation}
    \begin{aligned}
        \mathcal{J}_{\text{GRPO}}(\theta) &= \underbrace{\frac{1}{\sum_{i=1}^{G}|o_{i}|} \sum_{i=1}^G\sum_{t=1}^{|o_i|} \min\left[ r_{i,t}(\theta) \hat{A}_i,\, \operatorname{clip}\left(r_{i,t}(\theta); 1-\epsilon, 1+\epsilon\right) \hat{A}_i \right]}_{\text{Clipped policy update}} 
        - \underbrace{\beta \, \mathbb{D}_{\text{KL}}[\pi_\theta \parallel \pi_{\text{ref}}]}_{\text{KL penalty}} \\
        &\text{where } r_{i,t}(\theta)=\frac{\pi_\theta(o_{i,t} \mid q, o_{i,<t})}{\pi_{\theta_{\text{old}}}(o_{i,t} \mid q, o_{i,<t})}
    \end{aligned}
\end{equation}

% \begin{equation}
%         \mathcal{J}_{\text{GRPO}}(\theta) = \underset{\text{Clipped policy update}}{\underbrace{\frac{1}{\sum_{i=1}^{G}|o_{i}|} \sum_{i=1}^G\sum_{t=1}^{|o_i|} \min\left[ \frac{\pi_\theta(o_{i,t} \mid q, o_{i,<t})}{\pi_{\theta_{\text{old}}}(o_{i,t} \mid q, o_{i,<t})} \hat{A}_i,\, \operatorname{clip}(\frac{\pi_\theta(o_{i,t} \mid q, o_{i,<t})}{\pi_{\theta_{\text{old}}}(o_{i,t} \mid q, o_{i,<t})}; 1-\epsilon, 1+\epsilon) \hat{A}_i \right]}}  - \underset{\text{KL penalty}}{\underbrace{\beta \, \mathbb{D}_{\text{KL}}[\pi_\theta \parallel \pi_{\text{ref}}]}}
% \end{equation}

where  \( \pi_{\text{ref}} \) represents the reference model, and the term \( \mathbb{D}_{KL} \) introduces a KL divergence constraint to limit how much the model can deviate from this reference. The advantage estimate \( \hat{A}_i \) measures how much better the response \( o_i \) is compared to the average response, which is computed using a group of rewards \( \{r_1, r_2, \dots, r_G\} \) for the responses in set \( O \):
\begin{equation}
   \hat{A}_i = \frac{r_i - \text{mean}(\{r_1, r_2, \dots, r_G\})}{\text{std}(\{r_1, r_2, \dots, r_G\})}
\end{equation}

\section{Detailed Experimental Setup}
\label{appx:detailed_setup}
\subsection{Dataset}
% \label{sec:dataset_setting}
To keep the training recipe simple, we select training data exclusively from the GSM8K~\citep{cobbe2021training} and MATH~\citep{hendrycks2021measuring} datasets. 
For the MATH dataset, following prior studies~\citep{lightman2023let, wang2023math, sun2024easy}, we reserve the MATH500 subset as the test set, uniformly sample an additional 500 problems for validation, and combine the remaining 4,000 test problems with the original 7,500 training problems to form our training set. Each example in the MATH dataset is originally labeled with a difficulty level ranging from 1 to 5. In our experiments, we find that data difficulty is critical for successful zero RL (\textsection\ref{sec:data_complextiy_behaviur}) and it is necessary to use data that aligns with the model's capability. To investigate this phenomenon, we categorize the data into three difficulty levels: Easy (GSM8K and MATH lv.1), Medium (MATH lv.1–4), and Hard (MATH lv.3–5), with each category containing roughly 8,000 problems. 
For our main training runs, we use Easy for LLama-3.1-8B, Mistral- v0.1-7B, and DeepSeek-Math-7B; Medium for Qwen-2.5-0.5B; Hard for Mistral-Small-24B, Qwen-2.5-Math-7B, and Qwen-2.5-1.5B/7B/14B/32B, and we will report ablation study on data difficulty in \textsection\ref{sec:data_complextiy_behaviur}.

% \label{sec:reward_remove}

\subsection{Reward} 
% \label{sec:reward_remove}
We use a rule-based reward function that assigns rewards solely based on the correctness of the generated response: a correct final answer receives a reward of +1, while an incorrect one receives a reward of 0. Recent studies~\citep{deepscaler2025,chen2025empirical} often incorporate format-based rules into reward calculation, encouraging the model to follow specific output formats. However, we find that this approach may hinder the model's exploration and ultimately harm its performance particularly for the base models which struggle with following the format in the initial stage, as detailed in \textsection\ref{sec:remove_format}.

% We also discuss the impact of incorporating format-based rewards on training in \textsection\ref{}.

\subsection{Models} 
We conduct zero RL training experiments on Llama-3.1-8B~\citep{dubey2024llama}, DeepSeek-Math-7B~\citep{shao2024deepseekmath}, Mistral-v0.1-7B~\citep{jiang2023mistral7b}, Mistral-Small-24b-Base-2501~\citep{mistral2024small}, and Qwen-2.5 (0.5B, 1.5B, 7B, 14B, 32B)~\citep{yang2024qwen2}.
As we perform experiments for a variety of models, under extremely simple settings with small, simple datasets and only correctness reward, we refer to our obtained models as \emph{SimpleRL-Zoo} to represent a simple training recipe for a zoo of open base models.
For models with weaker instruction-following capabilities (Llama-3.1-8B, Mistral-v0.1-7B, and Qwen-2.5-0.5B/1.5B), we employ simpler prompts~\citep{abel} requiring only step-by-step reasoning. For models with stronger instruction-following abilities, we use more complex prompts~\citep{yang2024qwen2} that require the final answers to be placed in boxes. In our preliminary experiments, we observe that using complex prompts with models that have weak instruction-following capabilities often results in large amounts of irrelevant or nonsensical content being generated early in training, leading to instability.
The content of simpler prompts and more complex prompts is shown in Figure~\ref{fig:prompt_case} in Appendix.
%% todo: add figure for this prompt

% \begin{figure}[!t]
%         \centering
% \includegraphics[width=0.95\columnwidth]{fig/Prompt_Case.pdf}
% \caption{Comparison between simple prompts and more complex prompts.  \yh{This figure can be smaller, or move to appendix. }      }
%         \label{fig:prompt_case}
% \end{figure}

\subsection{Benchmark} 
We evaluate performance on standard mathematical reasoning benchmarks, including GSM8K~\citep{cobbe2021training}, MATH 500~\citep{hendrycks2021measuring}, Minerva Math~\citep{lewkowycz2022solving}, and OlympiadBench~\citep{he2024olympiadbench}, as well as on competition-level benchmarks such as AIME 2024 and AMC 2023. 
% We also evaluate the generalization ability of zero RL training using three benchmarks: IFEVAL~\citep{zhou2023instruction}, MMLU~\citep{hendrycks2020measuring}, and GPQA-Diamond~\citep{rein2024gpqa}. IFEVAL measures instruction-following capability, MMLU assesses the model's mastery of general knowledge, and GPQA-Diamond is a challenging benchmark that tests domain-specific expertise in chemistry, physics, and biology.
% \jh{We also need to mention that we also test generalization, on xxx benchmarks}
% \yh{
% \subsection{Other Configurations:}
% % \jh{I think we need to talk some important training/eval configs. for example, the rollout size (which is relevant to pass@k), the training max length and eval max length that we will mention again later, and mention we use verl [cite]. For some detailed configs that we want include in appendix (batch size, some GRPO hyperparams), we also need to cite that appendix here}
% We train our models using the verl~\citep{sheng2024hybridflow} framework. We provide detailed training and evaluation details in the Appendix ~\ref{sec:train_evaluate_details}.
% \label{sec:train_evaluate_details}
% We typically use the same set of hyperparameters to train and evaluate all models in the SimpleRL-Zoo series in default main experiment setting.

\subsection{Training and Evaluation Details}

\label{sec:train_evaluate_details}
We train our models using the verl~\citep{sheng2024hybridflow} framework. And we typically use the same set of hyperparameters to train and evaluate all models in the SimpleRL-Zoo series in default main experiment setting. We use a prompt batch size of 1,024 and generate 8 rollouts per prompt, with a maximum rollout length of 8,192 tokens. Training is performed using a mini-batch size of 256. The default sampling temperature is set to 1.0, and the clip ratio is 0.2. For models ranging from 0.5B to 14B parameters, we use a KL loss coefficient of 1e-4. For models larger than 14B, the KL loss coefficient is set to 1e-3. We build our evaluation script based on ~\citet{yang2024qwen2math}, using a temperature of 1.0 and a maximum generation length of 16K tokens. To ensure consistency, we adopt the same prompt template used during training. For most benchmarks, we report pass@1 results. However, for AIME 2024, which contains fewer problems, we report both pass@1 and average accuracy (avg@32), computed over 32 generated samples per problem.

\section{Detailed Evaluation Metrics}
\label{appx:eval_detail}
\paragraph{Reasoning Behavior Ratio:} To better understand the model's reasoning patterns throughout the training process, we adopt the cognitive behavior framework proposed by ~\citet{gandhi2025cognitive} and use GPT-4o~\citep{hurst2024gpt} to identify reasoning-related behaviors, including ``Backtracking", ``Verification", ``Subgoal Setting", and ``Enumeration". 
We report the ratio of responses that contain such cognitive behaviors.
While some recent studies suggest tracking reflection behavior using related keywords~\citep{yeo2025demystifying,xie2025logic} as monitoring signals, we argue that these keywords only exhibit only a weak correlation with high-level reasoning patterns like reflection and verification. As a result, they fail to adequately capture the development of these reasoning processes.  Further details can be found in Appendix~\ref{appx:bahaviour}.

\paragraph{Clip Ratio:} In the early stages of training, the base model exhibits weak instruction-following ability and often fails to stop appropriately, resulting in irrelevant or excessively long outputs. After training collapses, the model may also generate repetitive or overly extended responses. Since the model has a fixed maximum context length, such outputs may be truncated during both training and evaluation. To monitor this issue, we define the proportion of truncated outputs as the ``Clip Ratio".

\paragraph{Average Stopped Length:} Generations that are truncated often result from issues such as repetitive patterns or incomplete reasoning, which typically do not contribute to effective trajectories. To account for this factor, we introduce a new metric to track the average length of responses that are stopped under normal conditions. It is a more reliable metric to consider only valid responses, thereby eliminating the interference caused by unstopped responses.
\paragraph{Pass@k Accuracy:}
We track the pass@k accuracy, which represents the percentage of questions for which at least one correct response is obtained when sampling k responses per question. Pass@k serves as an indicator of the model's exploration capabilities and is particularly relevant for RL, as it reflects the model's ability to generate responses that can achieve a positive reward. Previously, some researchers believed that RL training might merely reorder responses within the original model distribution, as evidenced by the lack of improvement in pass@k accuracy following RL training~\citep{shao2024deepseekmath}.

\section{Detailed Result of SimpleRL}
\label{appx:DetailedResult}
Following the setup described in Section~\ref{sec:setup}, we perform ``zero training" on various base models. The trained models are then evaluated on multiple benchmarks, including GSM8K, MATH 500, Minerva Math, OlympiadBench, AIME2024, and AMC2023. The average results across all these benchmarks are presented in Figures~\ref{fig1:acc&len} and~\ref{fig2:clip&stop}. In this section, we provide more detailed results. Figure~\ref{fig:appx_acc&len} illustrates the trends in accuracy and response length, while Figure~\ref{fig:appx_clip&stop} shows the trends in clip ratio and stopped length.

\begin{figure}[!t]
        \centering
\includegraphics[width=0.8\columnwidth]{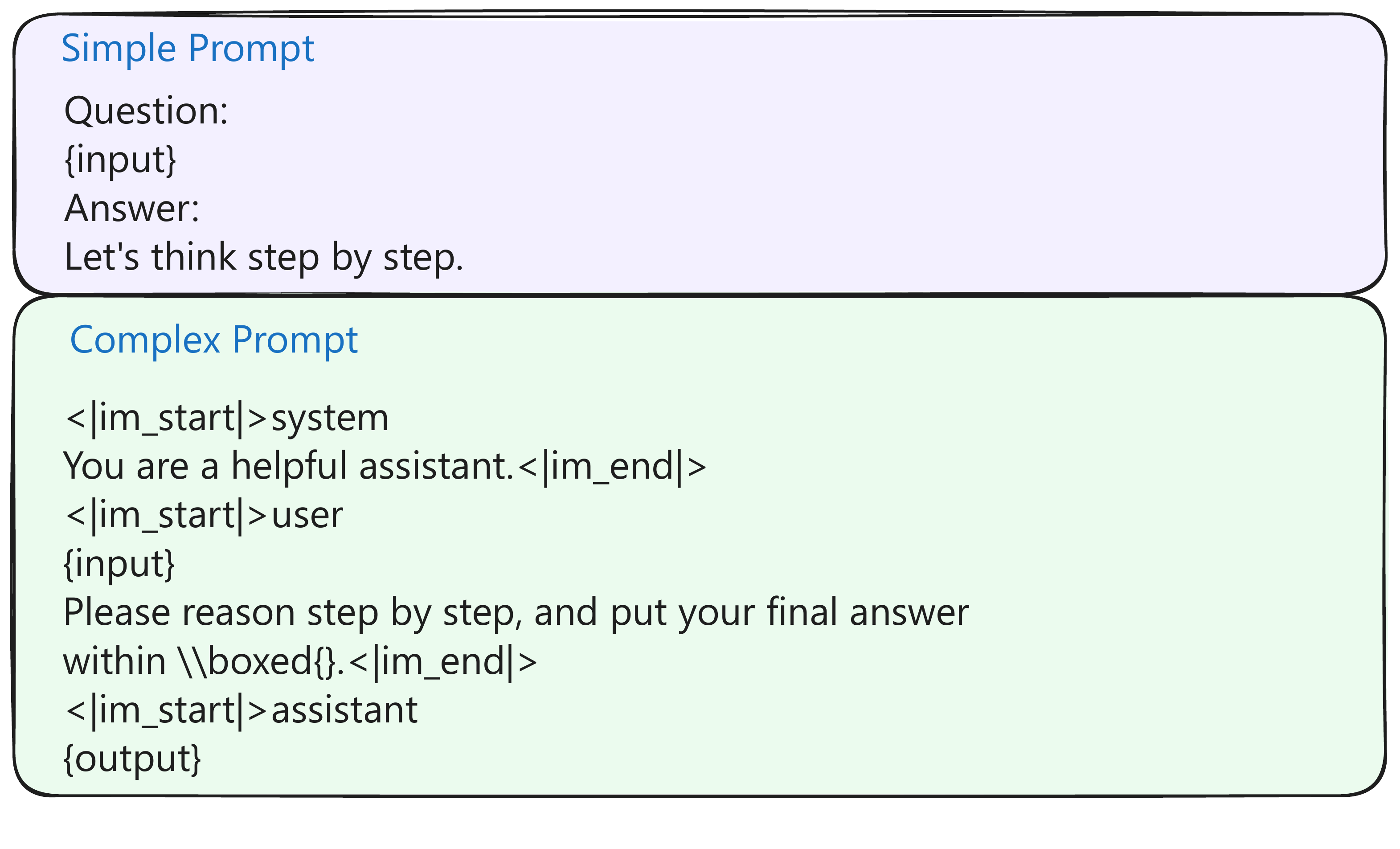}
\caption{Comparison between simple prompts and more complex prompts.}
        \label{fig:prompt_case}
\end{figure}

\begin{figure*}[!h]
    \centering
    \includegraphics[width=\textwidth]{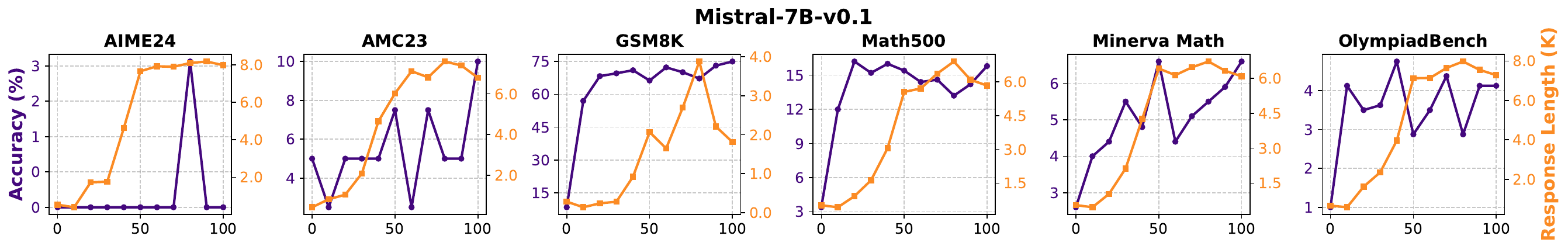} \\ \vspace{-3pt}
    \includegraphics[width=\textwidth] {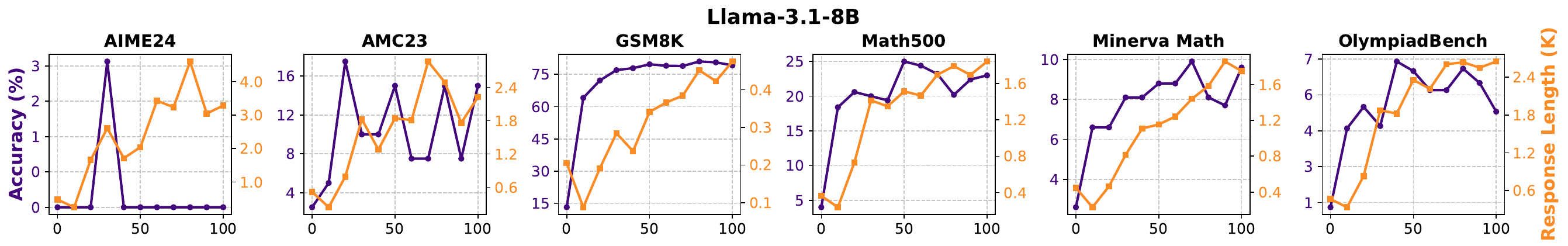}
    \\ \vspace{-3pt}
    \includegraphics[width=\textwidth] {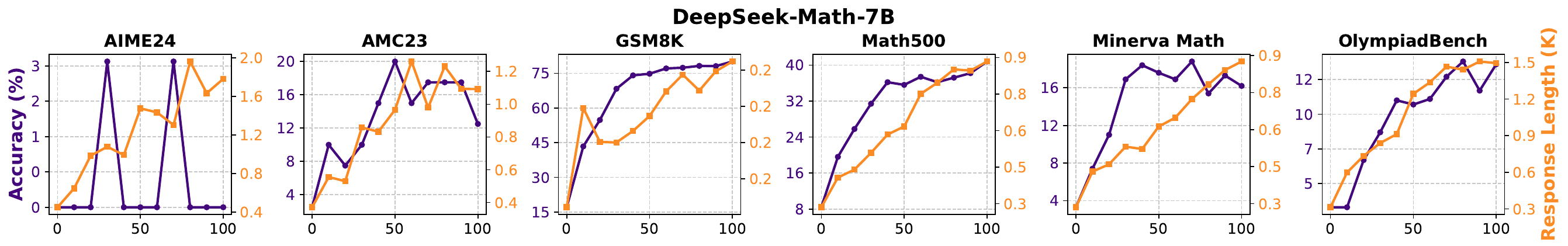}
    \\ \vspace{-3pt}
    \includegraphics[width=\textwidth] {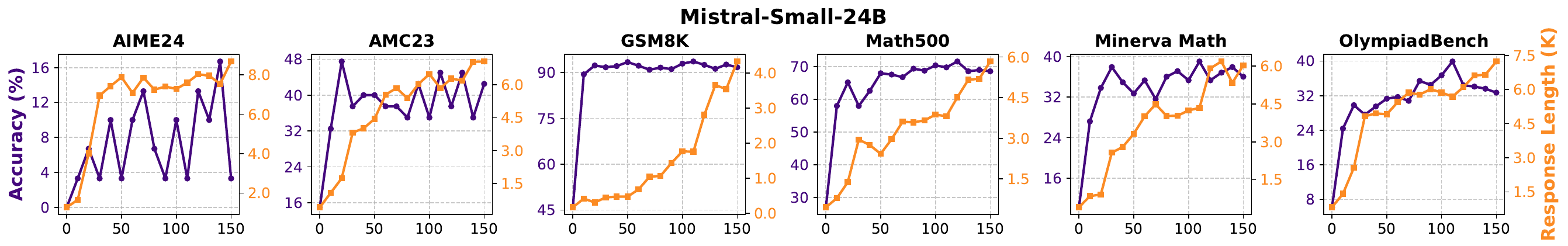}
    \\ \vspace{-3pt}
    \includegraphics[width=\textwidth] {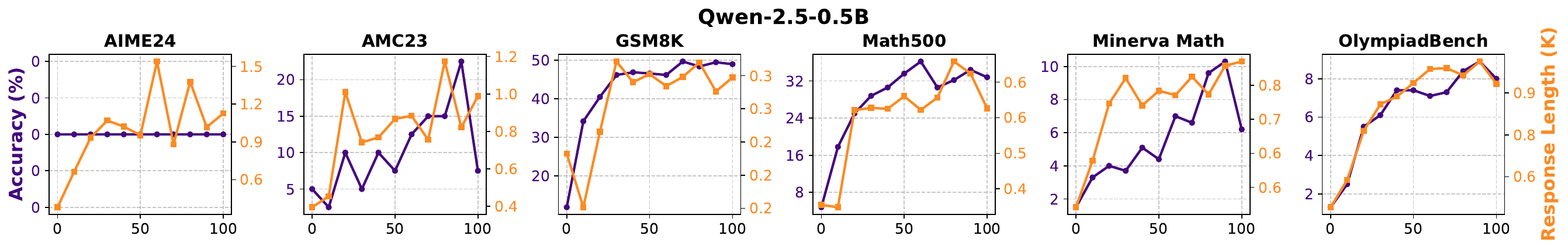}
    \\ \vspace{-3pt}
    \includegraphics[width=\textwidth] {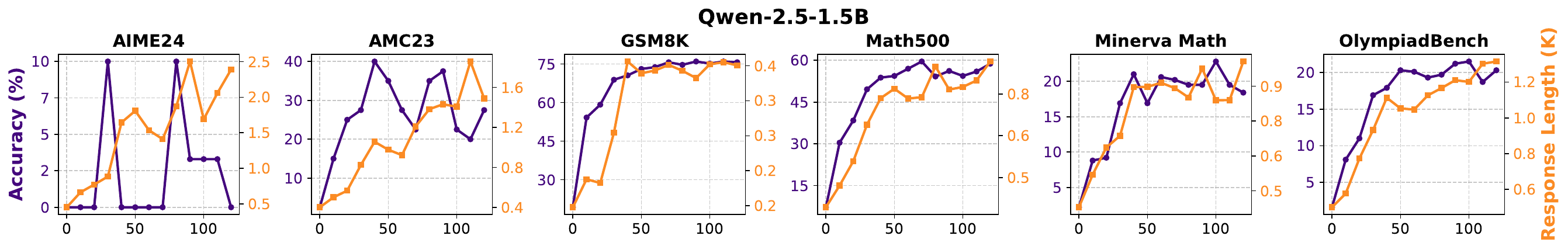}
    \\ \vspace{-3pt}
    \includegraphics[width=\textwidth] {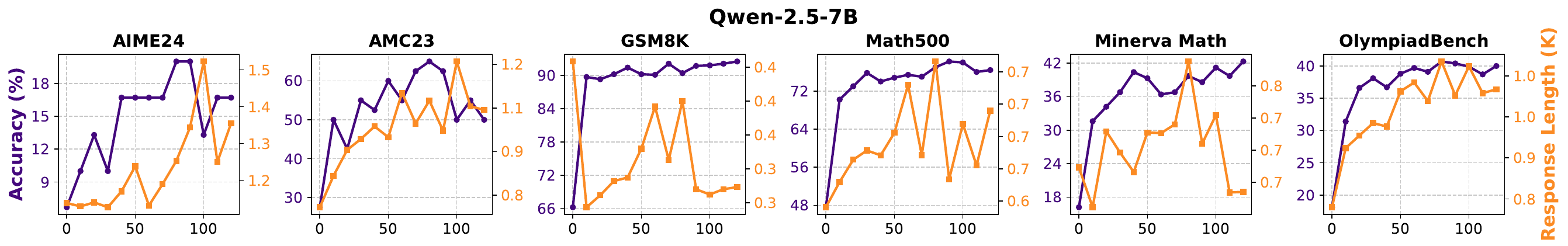}
    \\ \vspace{-3pt}
    \includegraphics[width=\textwidth] {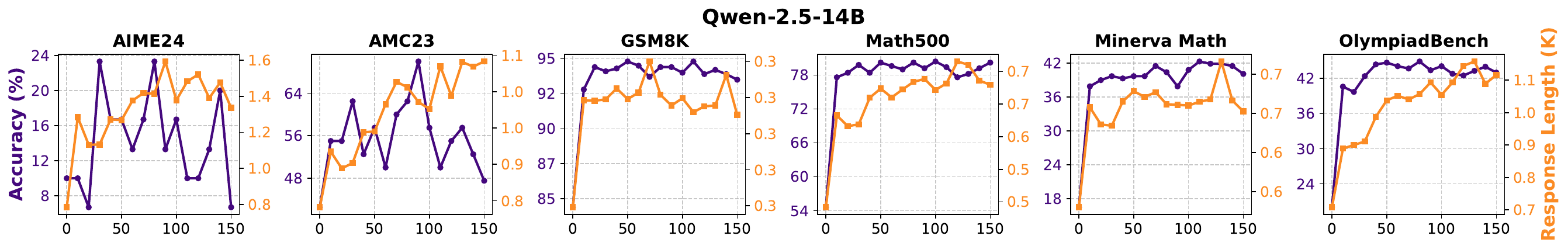}
    \\ \vspace{-3pt}
    \includegraphics[width=\textwidth] {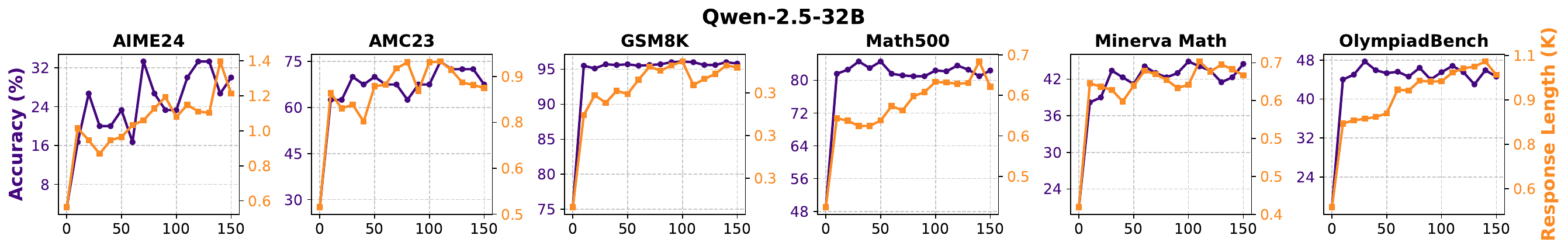}
    \\ \vspace{-3pt}
     \includegraphics[width=\textwidth] {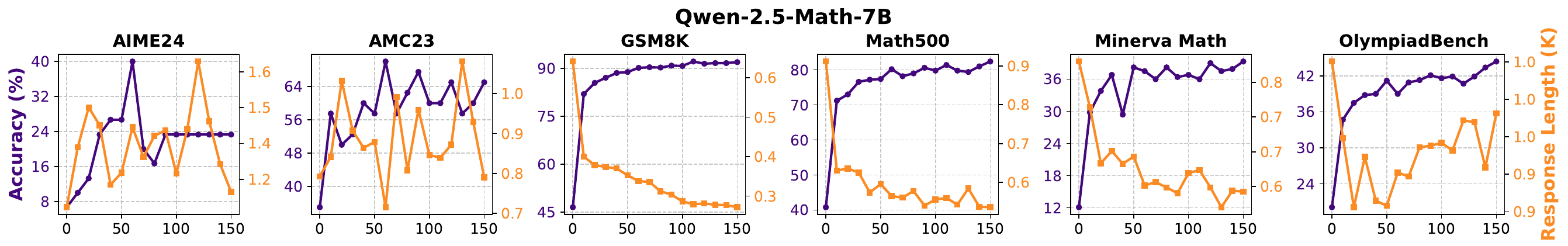}
    \\ 
    \vspace{-12pt}
    \caption{A detailed evaluation of accuracy and response length throughout the training steps for various models. The x-axis represents the training steps, with the purple line showing the accuracy trend and the yellow line depicting the response length.}
    \label{fig:appx_acc&len}
    \vspace{-10pt}
\end{figure*}

\begin{figure*}[!h]
    \centering
    \includegraphics[width=\textwidth]{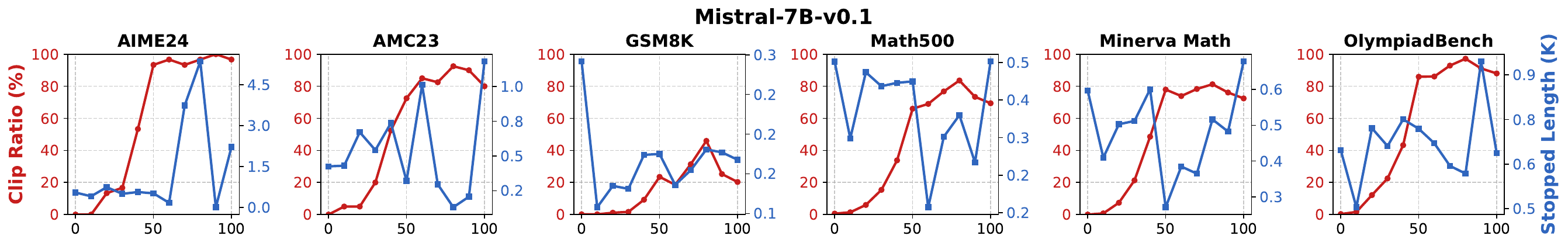} \\ \vspace{-3pt}
    \includegraphics[width=\textwidth] {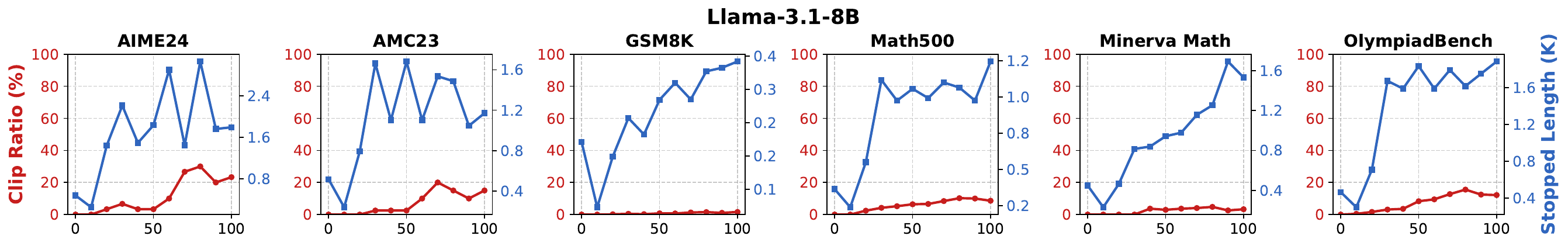}
    \\ \vspace{-3pt}
    \includegraphics[width=\textwidth] {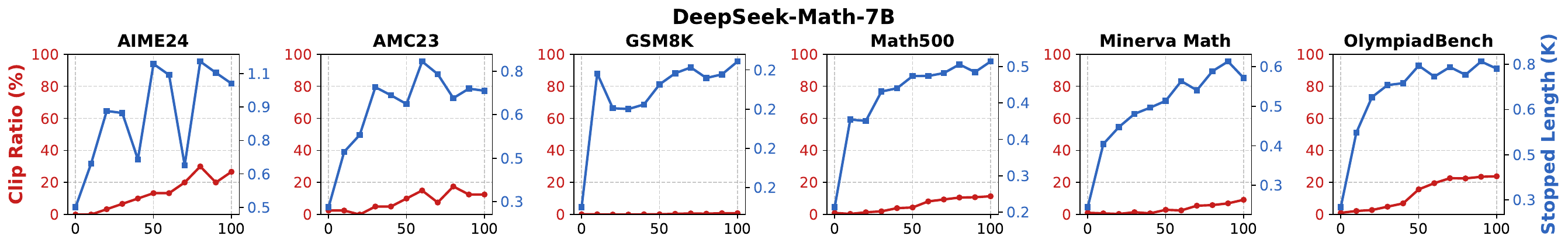}
    \\ \vspace{-3pt}
    \includegraphics[width=\textwidth] {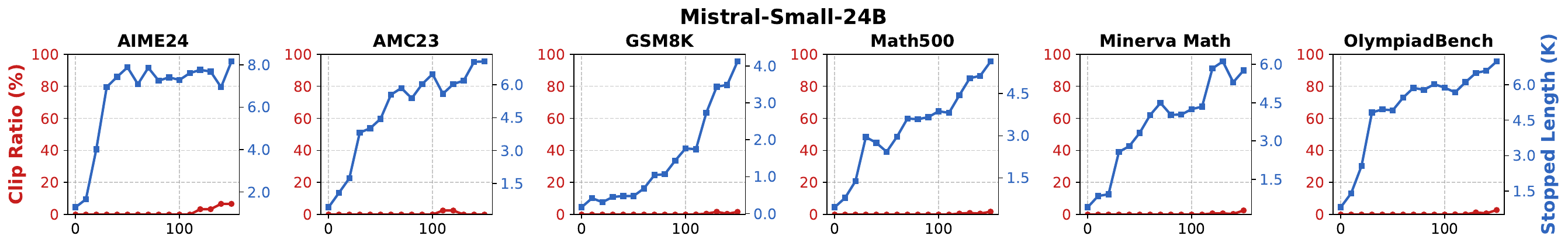}
    \\ \vspace{-3pt}
    \includegraphics[width=\textwidth] {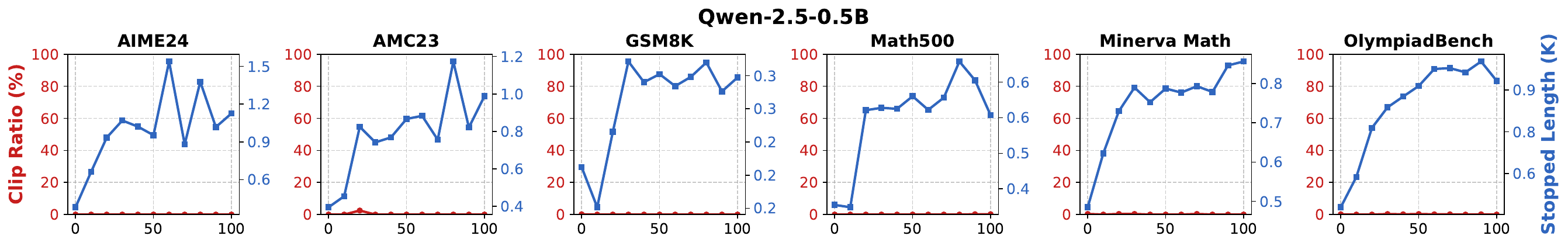}
    \\ \vspace{-3pt}
    \includegraphics[width=\textwidth] {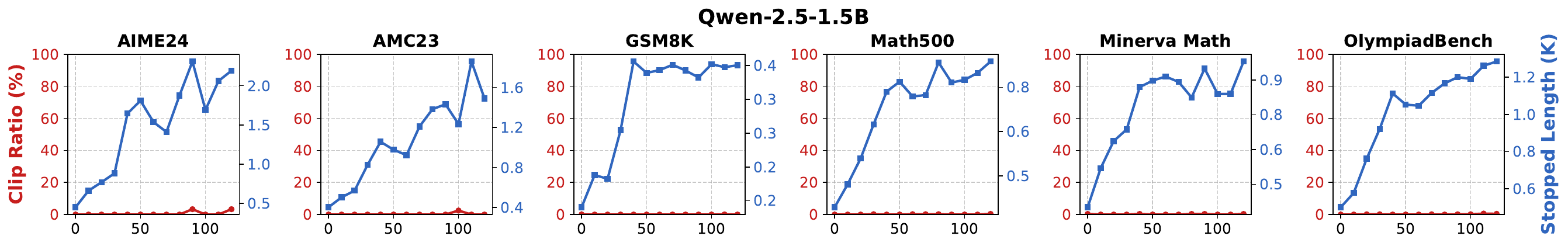}
    \\ \vspace{-3pt}
    \includegraphics[width=\textwidth] {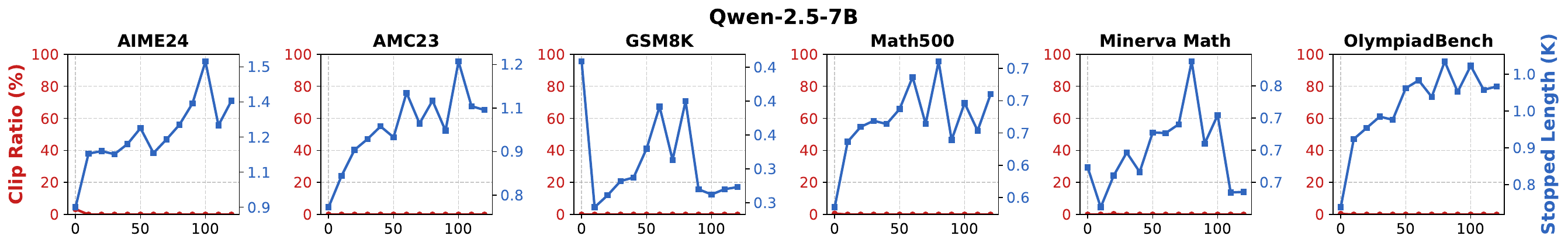}
    \\ \vspace{-3pt}
    \includegraphics[width=\textwidth] {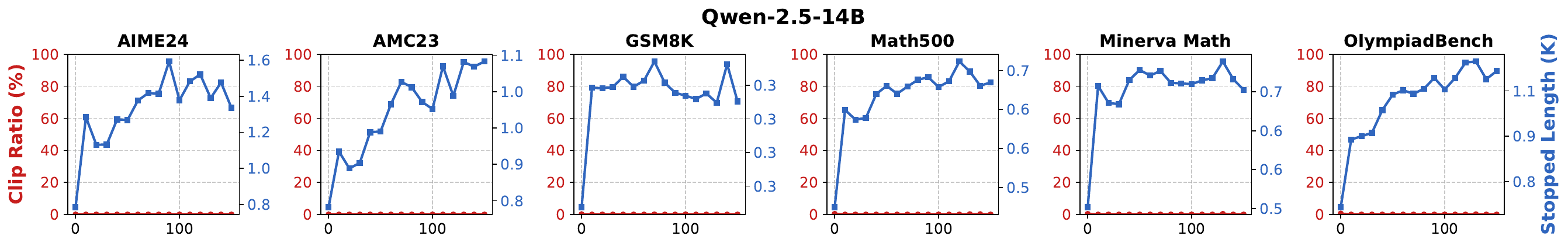}
    \\ \vspace{-3pt}
    \includegraphics[width=\textwidth] {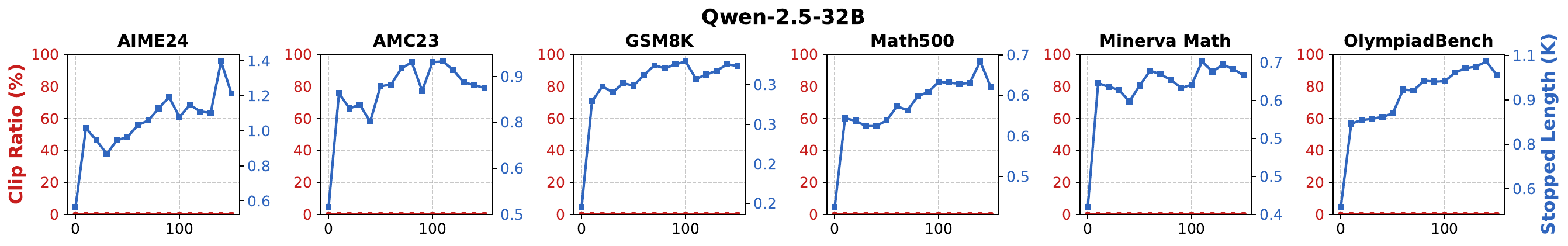}
    \\ \vspace{-3pt}
     \includegraphics[width=\textwidth] {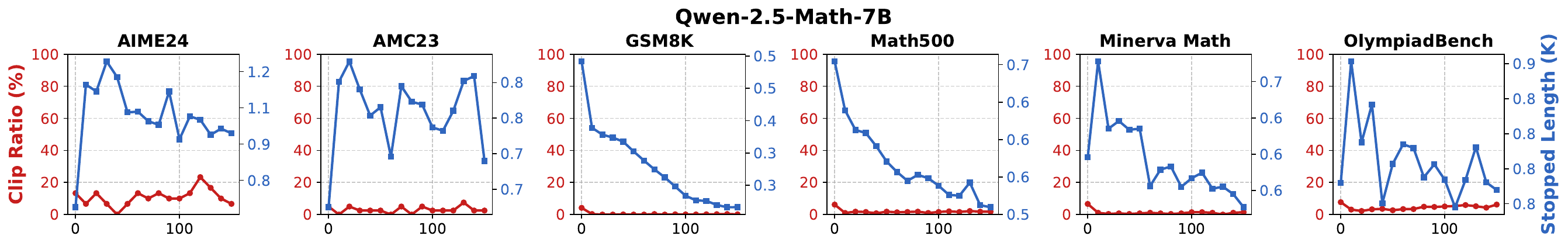}
    \\ 
    \vspace{-12pt}
    \caption{
    A detailed evaluation of clip ratio and stopped length throughout the training steps for various models. The x-axis represents the training steps, with the red line showing the clip ratio trend and the blue line depicting the average stopped length. }
    \label{fig:appx_clip&stop}
    \vspace{-10pt}
\end{figure*}

\section{Quantitative Behavior Validation}
\label{sec:human_consistency}
We assess the consistency between GPT-4o labeled reasoning behaviors and human annotations by having human experts annotate 105 model outputs. Table~\ref{tab:human_consistency} below presents the prediction rates and agreement rate. The prediction rate reflects how frequently each reasoning behavior is identified, while the agreement rate is the proportion of data on which the labelers (Human and GPT-4o) make the same prediction.

Our results indicate a generally good level of agreement between GPT-4o and human annotations. However, GPT-4o tends to be more conservative when labeling certain behaviors such as Verification and Subgoal Setting. Upon closer examination, we observe that in cases with long CoT containing multiple reasoning behaviors, the model often favors labeling more obvious behaviors like Enumeration, while overlooking subtler ones.

\begin{table}[t]
    \centering
\resizebox{0.85\textwidth}{!}{
\begin{tabular}{llll}
\hline
\textbf{Behavior} & \textbf{Score by GPT-4o (\%)} & \textbf{Score by Human (\%)} & \textbf{Raw Agreement (\%)} \\ \hline
Verification      & 78.10\% (82/105)              & 85.71\% (90/105)             & 90.48\% (95/105)            \\
Backtracking      & 33.33\% (35/105)              & 35.24\% (37/105)             & 98.10\% (103/105)           \\
Subgoal Setting   & 66.67\% (70/105)              & 74.29\% (78/105)             & 90.48\% (95/105)            \\
Enumeration       & 61.90\% (65/105)              & 63.81\% (67/105)             & 94.29\% (99/105)            \\ \hline
\end{tabular}}
\caption{The consistency between GPT-4o labeled reasoning behaviors and human annotations}
\label{tab:human_consistency}
\end{table}

\section{Impact of General SFT on the Performance of Reinforcement Learning}\label{sec:general_sft}

\begin{table}[t]
    \centering
    \small
    \begin{tabular}{cccccccc}
    \toprule
\multicolumn{1}{c}{\textbf{Init Model}} & \textbf{GSM8K} & \textbf{\begin{tabular}[c]{@{}c@{}}MATH \\ 500\end{tabular}} & \textbf{\begin{tabular}[c]{@{}c@{}}Minerva \\ Math\end{tabular}} & \textbf{\begin{tabular}[c]{@{}c@{}}Olympiad\\ Bench\end{tabular}} & \textbf{\begin{tabular}[c]{@{}c@{}}AIME24 \\ (pass@1)\end{tabular}} & \textbf{AMC23} & \textbf{Avg.} \\ \midrule
0 Step & 92.0 & 70.6 & 36.8 & 36.6 & 16.7 & 45.0 & 49.6 \\
10 Step & 93.0 & 69.4 & 39.7 & 32.3 & 10.4 & 44.1 & 48.2 \\
20 Step & 92.6 & 65.2 & 34.2 & 30.7 & 6.7 & 38.4 & 44.6 \\
200 Step & 90.3 & 59.0 & 31.6 & 23.3 & 2.1 & 26.9 & 38.9 \\
1000 Step & 88.9 & 48.8 & 27.6 & 20.7 & 2.5 & 18.1 & 34.4 \\
2000 Step & 89.8 & 49.0 & 23.2 & 18.1 & 0.8 & 20.3 & 33.5 \\
4000 Step & 87.7 & 52.0 & 23.5 & 17.2 & 2.1 & 21.6 & 34.0 \\
\bottomrule
\end{tabular}
\caption{Experimental results from multiple Mistral-Small-24B models, each fine-tuned with a different number of SFT steps on a general SFT dataset for RL. The "number of steps" refers to the number of SFT steps applied. The reported benchmarks reflect the performance metrics on various evaluation benchmarks, measured using the model that achieved the best average performance after 100 iterations of reinforcement learning training.}
\label{tab:general_sft_results}
\end{table}

We also investigated the general SFT setting beyond math-related datasets. In this setup, we first conducted SFT on Mistral-Small-24B using the widely adopted OpenHermes-2.5 dataset.\footnote{\url{https://huggingface.co/datasets/teknium/OpenHermes-2.5}} We implement with LLaMA-Factory~\citep{zheng-etal-2024-llamafactory} and adopt common hyperparameters of SFT, including 512 examples per batch with a constant learning rate of 1e-5. For consistency with our other experiments, we fine-tuned the model using the Qwen chat template.
After SFT, we preserved multiple checkpoints at different training steps, and nearly 800 steps correspond to 1 epochs on the SFT dataset. We then performed reinforcement learning on these models using identical hyperparameters as in our zero-RL training experiments.

Table~\ref{tab:general_sft_results} presents our findings, with performance reported as the best results achieved during RL training up to 100 iterations. The results demonstrate an inverse relationship between SFT steps and subsequent RL performance: models with more SFT steps showed diminished performance after RL training. While the average performance after 10 SFT steps remained comparable to the base model, it still exhibited some negative effects. More significantly, models with more than 20 steps showed substantially reduced RL potential.
Therefore, we conclude that RL training produces the best performance gain when applied directly to the base model without any supervised fine-tuning, i.e., the zero RL training.

\section{Impact of Exploration-Related Hyperparameters}
\label{sec:impact_explore_hyper}
In this section, we examine the effects of exploration-related hyperparameters on "zero-training." Drawing inspiration from ~\citet{zeng2024b,liu2024diving}, we focus on two key factors: sampling size (the number of responses per query) and sampling temperature.

\begin{figure}[!t]
        \centering
\includegraphics[width=0.6\columnwidth]{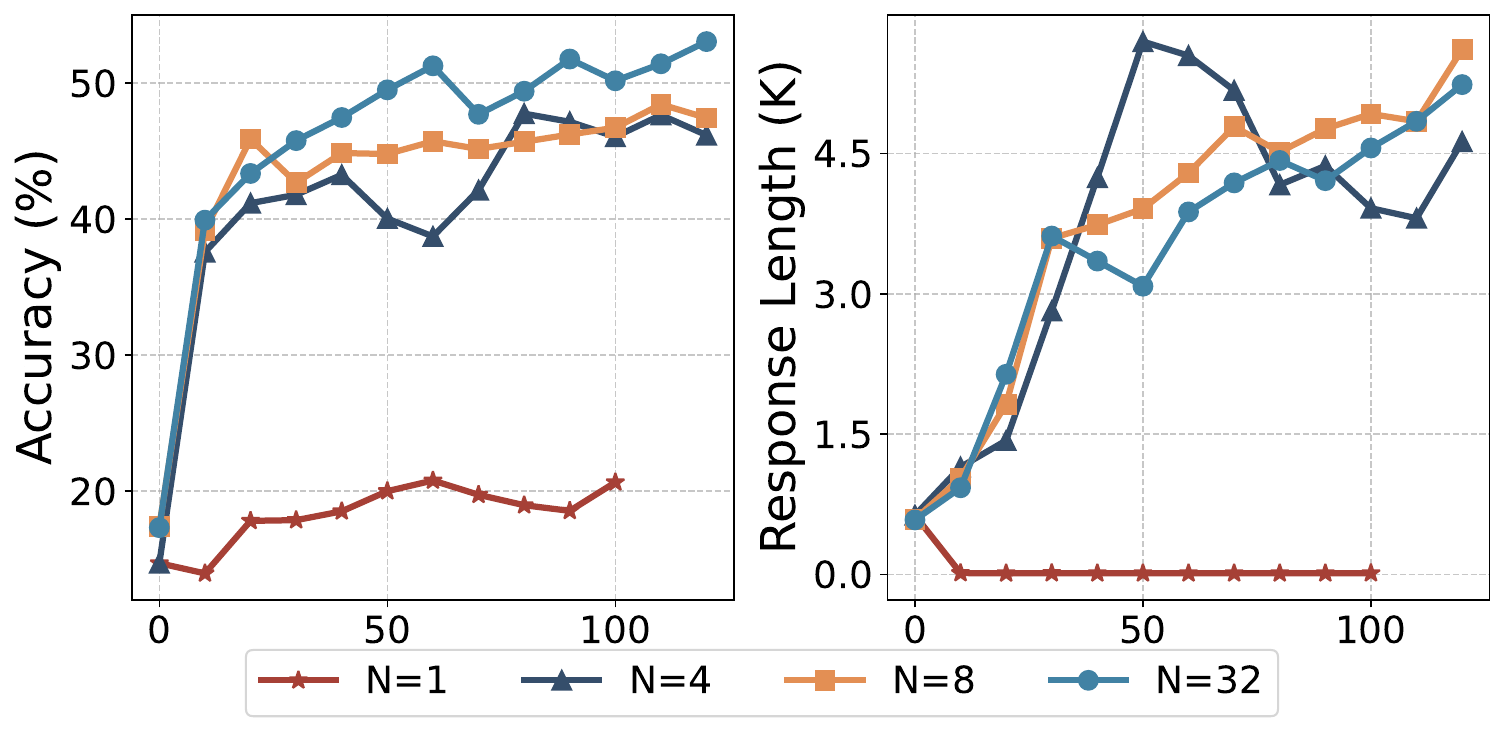}
\caption{Comparison of accuracy and response length using different sampling numbers N = 1, 4, 8, 32. The training data is the Hard part (MATH lv.3–5) with the same setting in main results, as described in \S~\ref{sec:setup}.}
        \label{fig:sampling_number}
\end{figure}

\paragraph{Sampling Size:} We examine how varying sampling sizes $N \in \{1,4,8,16,32\}$ influence the training process using the Mistral 24B model; these results are presented in Figure~\ref{fig:sampling_number}. Our analysis reveals a clear trend: as $N$ increases, the model's average performance notably improves, and variability in response lengths becomes significantly more stable. For example, after 100 training steps, the scenario with $N=32$ achieves an average accuracy approximately 6 points higher than that with $N=8$. Conversely, smaller sampling sizes ($N=1$ and $N=4$) cause training instability and potential collapse, indicated by rapid growth in generated length without corresponding accuracy improvements. We hypothesize that larger sample sizes enable the model to explore a broader and more diverse training space, which stabilizes advantage estimation and sustains continuous performance improvement.

% \begin{figure}[!t]
%         \centering
% \includegraphics[width=0.7\columnwidth]{fig/plot_temperature.pdf}
% \caption{Impact of training and evaluation temperatures on Qwen-2.5-0.5b's average final performance (x-axis: evaluation temp, y-axis: training temp).}
%         \label{fig:train_sample_temperature}
% \end{figure}

% \begin{figure}[!t]
%     % 左侧图片
%     \begin{minipage}[t]{0.48\columnwidth} % 留2%间隙
%         \centering
%         \includegraphics[width=\linewidth]{fig/plot_figure8_v3_diff_n-mistral-24b}\vspace{-10pt}
%         \caption{Comparison of accuracy and response length using different sampling numbers N = 1, 4, 8, 32. The training data is the Hard part (MATH lv.3–5) with the same setting in main results, as described in \S~\ref{sec:setup}.}
%         \label{fig:sampling_number}
%     \end{minipage}
%     \hfill % 填充间隙
%     % 右侧图片
%     \begin{minipage}[t]{0.48\columnwidth}
%         \centering
%         \includegraphics[width=\linewidth]{fig/plot_temperature.pdf}
%         \caption{Impact of training and evaluation temperatures on Qwen-2.5-0.5b's average final performance (x-axis: evaluation temp, y-axis: training temp).}
%         \label{fig:train_sample_temperature}
%     \end{minipage}
% \end{figure}

\paragraph{Sampling Temperature:} We conduct research on Qwen-2.5-0.5B to analyze the impact of sampling temperature during both training and evaluation on model performance. The results, presented in Figure~\ref{fig:train_sample_temperature} , indicate that training with higher temperatures generally leads to better average performance. For instance, models trained with temperatures of 1.0 and 1.2 outperform those trained with 0.8 and 0.6. Additionally, we find that the optimal evaluation temperature depends on the training temperature. Specifically, models trained at higher temperatures require higher sampling temperatures during evaluation, as using greedy sampling often results in repetitive outputs. Conversely, models trained at lower temperatures perform best when evaluated with lower sampling temperatures.

\section{SimpleRL-Zoo For Qwen2.5-Math-7B}
\begin{figure*}[]
    \centering
    \subfigure{
\includegraphics[width=0.45\textwidth]{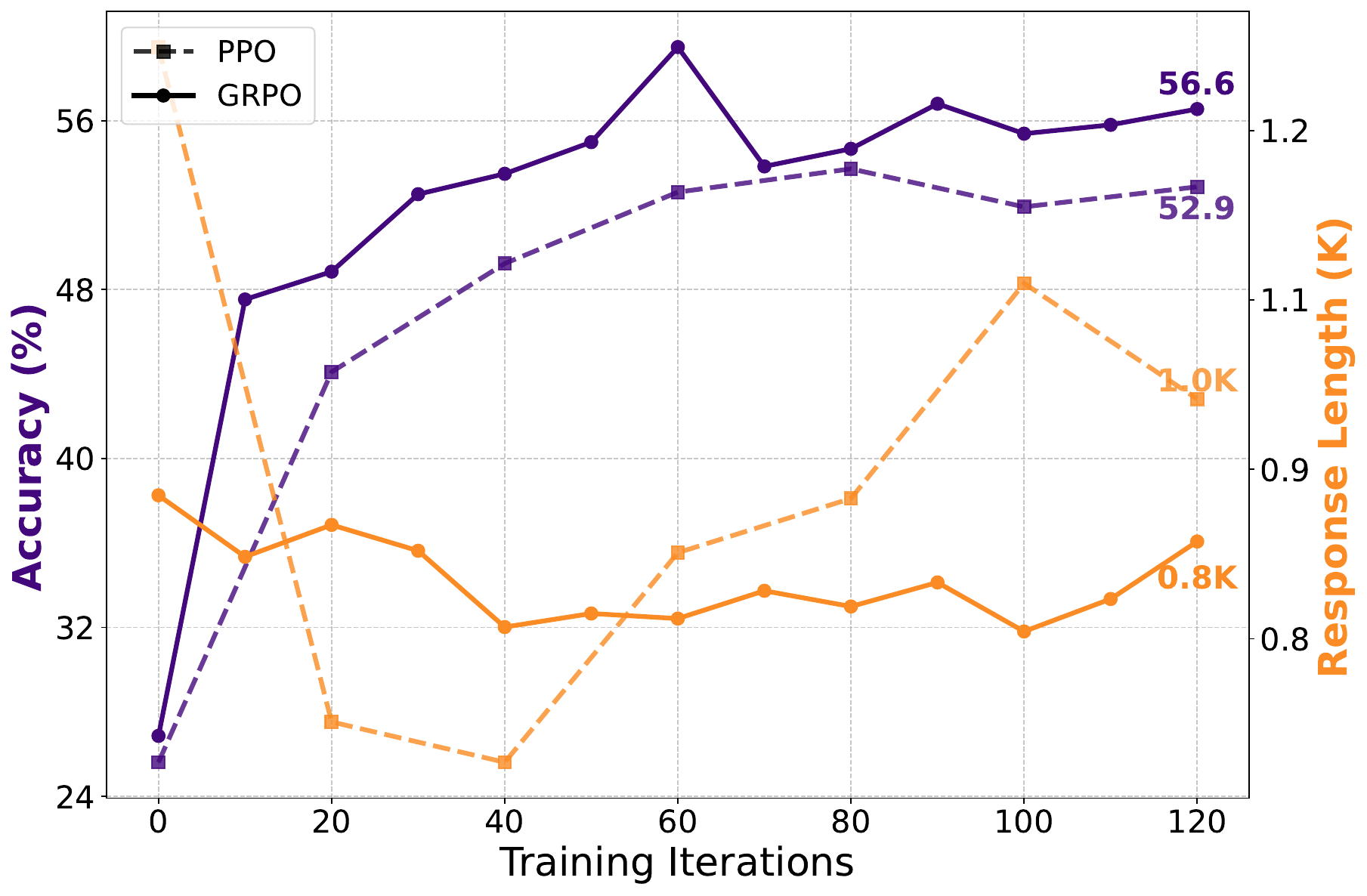}
    }
    \subfigure{
        \includegraphics[width=0.45\textwidth]{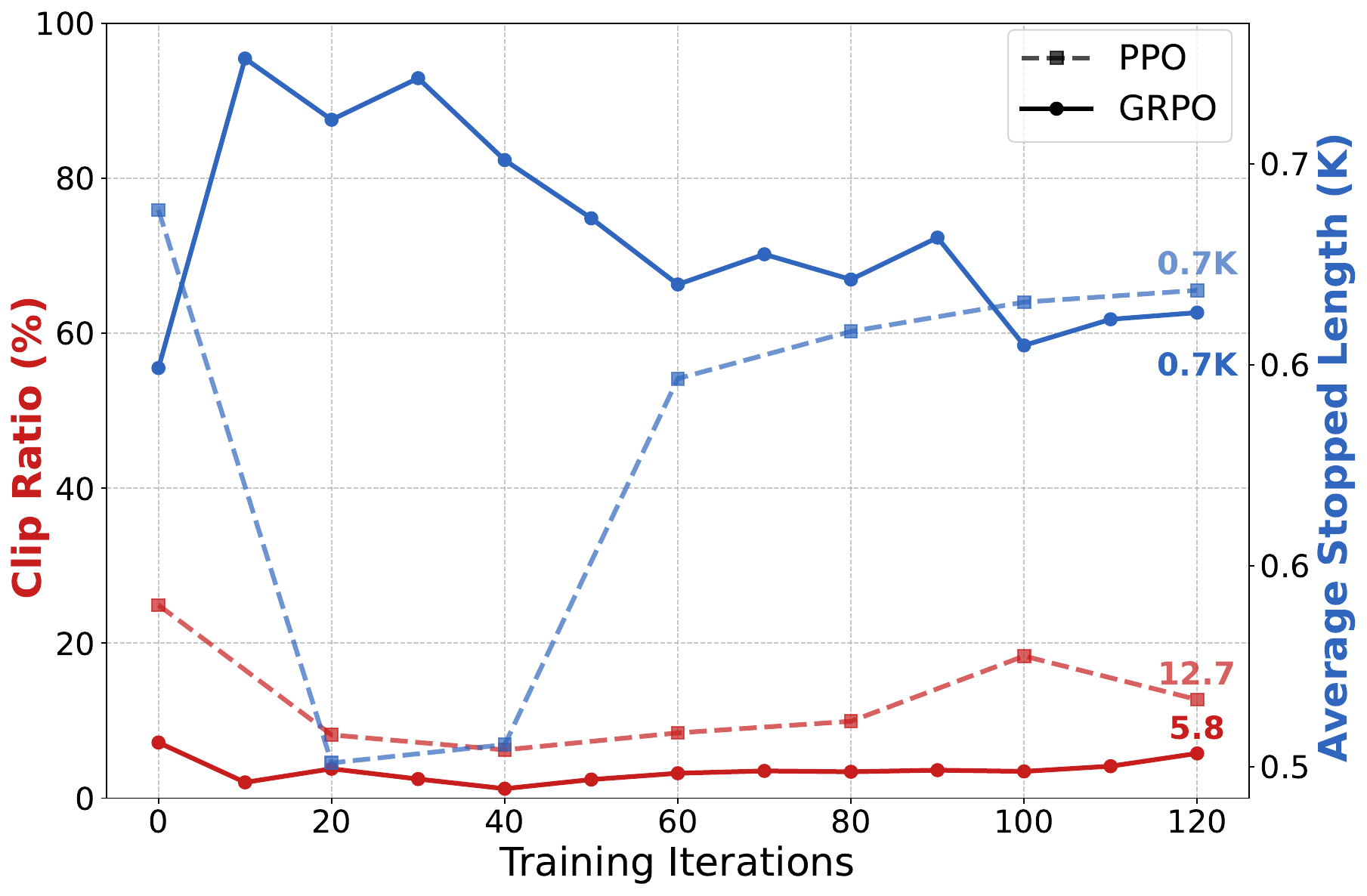}
    } \vspace{-10pt}
    \caption{Comparison of accuracy and response length between PPO and GRPO on Qwen2.5-Math-7B. The base model is trained using 8K examples from MATH lv3-5, with the same settings described in \S~\ref{sec:setup}.}
    \label{fig:qwen-math}
\end{figure*}

In this section, we conduct experiments on Qwen2.5-Math-7B~\citep{yang2024qwen2} using the ``hard part" data, as described in \S~\ref{sec:setup}, which consists of only 8K examples from MATH lv3-5. We apply both the PPO and GRPO algorithms to train our base model, and the overall evaluation results across training steps are shown in Figure~\ref{fig:qwen-math}. The final performance and response length for both algorithms converge to similar values, with GRPO slightly outperforming PPO. While the performance continues to improve, the response length does not exhibit a similar trend. Specifically, the stopping length for both algorithms remains relatively unchanged, and fluctuations in the average response length are primarily attributed to changes in the clip ratio. There are two main reasons for this behavior: First, the maximum context length for Qwen2.5-Math-7B is 4K, which is limited compared to other models with context lengths exceeding 8K, leading to a high clip ratio. Second, as a math-specific model, Qwen2.5-Math-7B already performs very well on MATH, the dataset we used for training, so it may not face enough challenge to further extend its response length. Therefore, we hypothesize that more challenging data might be needed to push this capable model further.

\section{Reasoning Behavior Analysis}
\label{appx:bahaviour}
We apply ~\citet{gandhi2025cognitive}'s cognitive behavior framework to perform a detailed analysis of how model reasoning behaviors change during "zero training." We first describe our analysis setup, then compare reflection keyword tracking against this framework to monitor reflective behaviors. Finally, we use case studies to illustrate how the reasoning behaviors of various models evolve during training.

\subsection{Setup}
\label{sec:bahaviour_setup}
We use GPT4-o to  identify and analyze the following key reasoning behaviors exhibited in the model's responses, with the prompt shown in Figure~\ref{fig:prompt_reasoning_behaviors}:

(1) \textbf{Backtracking}: The model actively identifies errors during response generation and explicitly revises previously used methods.

(2) \textbf{Verification}: The model systematically checks intermediate results to ensure correctness.

(3) \textbf{Subgoal Setting}: The model decomposes complex problems into smaller, manageable steps.

(4) \textbf{Enumeration}: The model exhaustively considers multiple cases or possibilities to solve problems.

Note that we replaced "Backward Chaining" with "Enumeration," as the former was not relevant to our task.

\begin{figure*}[!t]
    \centering
    \begin{minipage}{0.50\textwidth}
        \centering
        \includegraphics[width=\columnwidth]{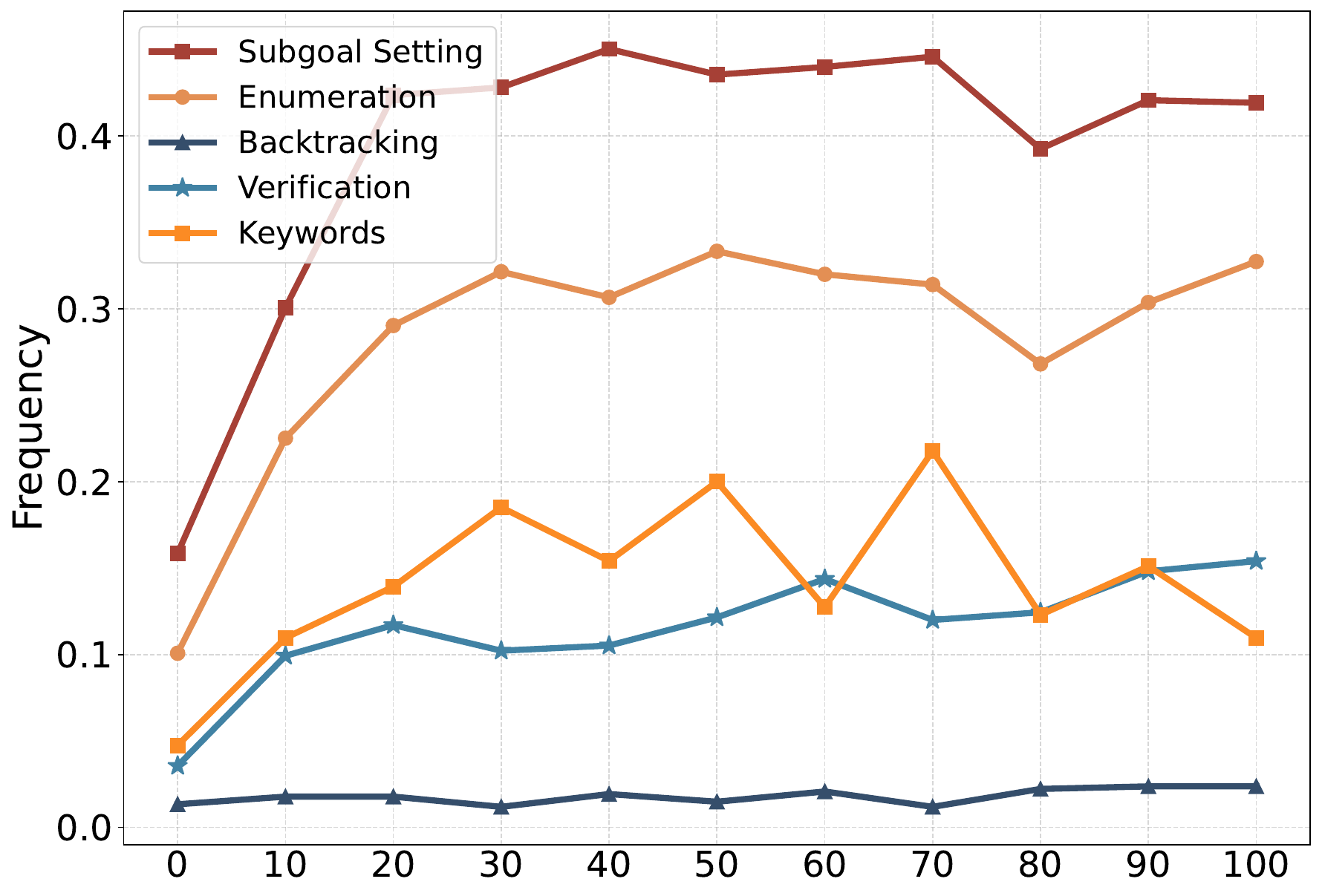}
        \caption{Changes in reflection behavior identified by different methods.}
        \label{fig:method_reasoning_behaviors}
    \end{minipage}
    \hfill
    \begin{minipage}{0.40\textwidth}
        \centering
        \includegraphics[width=\columnwidth]{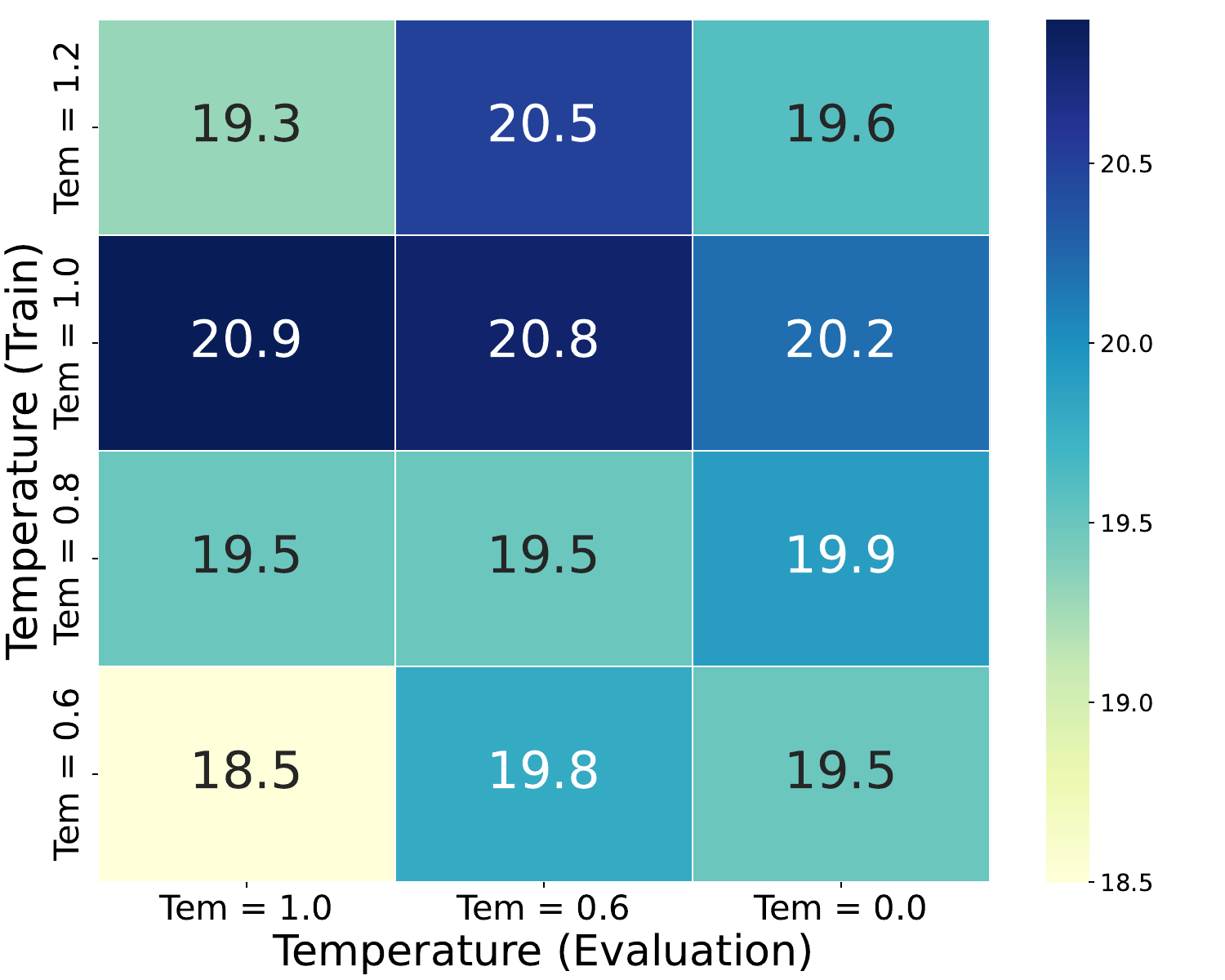}
        \caption{Impact of training and evaluation temperatures on Qwen-2.5-0.5b's average final performance (x-axis: evaluation temp, y-axis: training temp).}
        \label{fig:train_sample_temperature}
    \end{minipage}
\end{figure*}

\begin{figure}[!t]
        \centering
\includegraphics[width=\columnwidth]{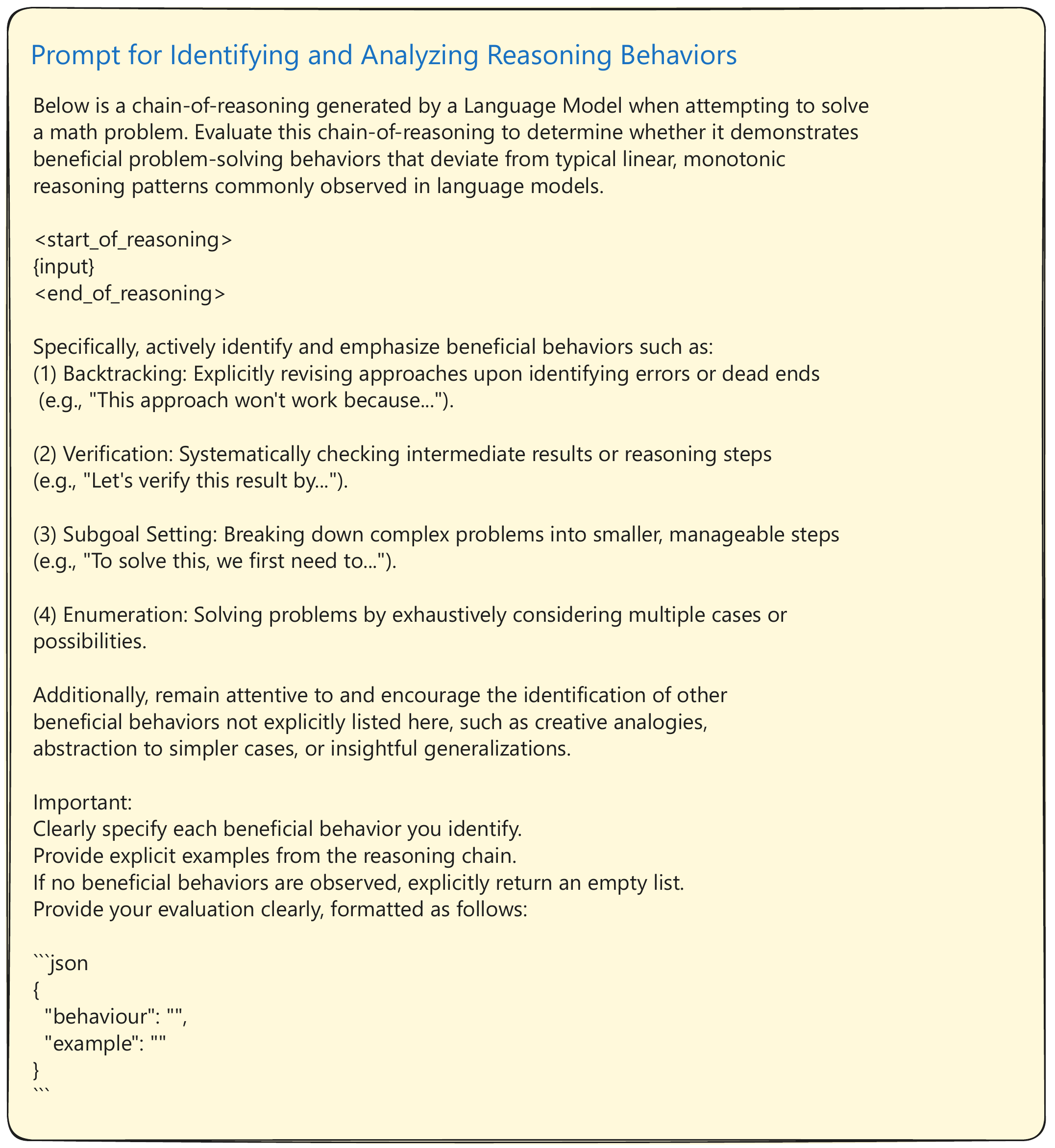}
\caption{Prompt for identifying and analyzing reasoning behaviors.
        }
        \label{fig:prompt_reasoning_behaviors}
\end{figure}

\subsection{Comparison of Different Reasoning Behavior Tracking Methods}
% \label{appx:reason_behavior_analysis}
% \begin{figure}[!t]
%     \centering
%     \includegraphics[width=0.7\columnwidth]{fig/plot_appx_figure5_v1_behaviors_plus_keywords_deepseek-math-7b.pdf}
%     \caption{Changes in reflection behavior identified by different methods.}
%     \label{fig:method_reasoning_behaviors}
% \end{figure}

% \begin{figure*}[!t]
%     \centering
%     \begin{minipage}{0.50\textwidth}
%         \centering
%         \includegraphics[width=\columnwidth]{fig/plot_appx_figure5_v1_behaviors_plus_keywords_deepseek-math-7b.pdf}
%         \caption{Changes in reflection behavior identified by different methods.}
%         \label{fig:method_reasoning_behaviors}
%     \end{minipage}
%     \hfill
%     \begin{minipage}{0.40\textwidth}
%         \centering
%         \includegraphics[width=\columnwidth]{fig/plot_temperature.pdf}
%         \caption{Impact of training and evaluation temperatures on Qwen-2.5-0.5b's average final performance (x-axis: evaluation temp, y-axis: training temp).}
%         \label{fig:train_sample_temperature}
%     \end{minipage}
% \end{figure*}

Using DeepSeek Math's "zero-training" process as an example, we compare two different methods for monitoring reasoning behavior. The first method tracks the occurrence of specific keywords in the model's responses, such as "recheck," "rethink," "try again," "wait," "alternatively," "retry," and "however." The second method employs ~\citep{gandhi2025cognitive}'s cognitive framework for evaluation. Figure~\ref{fig:method_reasoning_behaviors} illustrates the observed changes in reasoning behavior according to these two approaches. During the training process, we observe that the proportion of specified keywords in the DeepSeek math model's responses remains consistently low, exhibiting minimal variation. Conversely, reasoning behaviors identified by the cognitive framework demonstrate a significant upward trend.

\begin{figure}[!t]
        \centering
\includegraphics[width=\columnwidth]{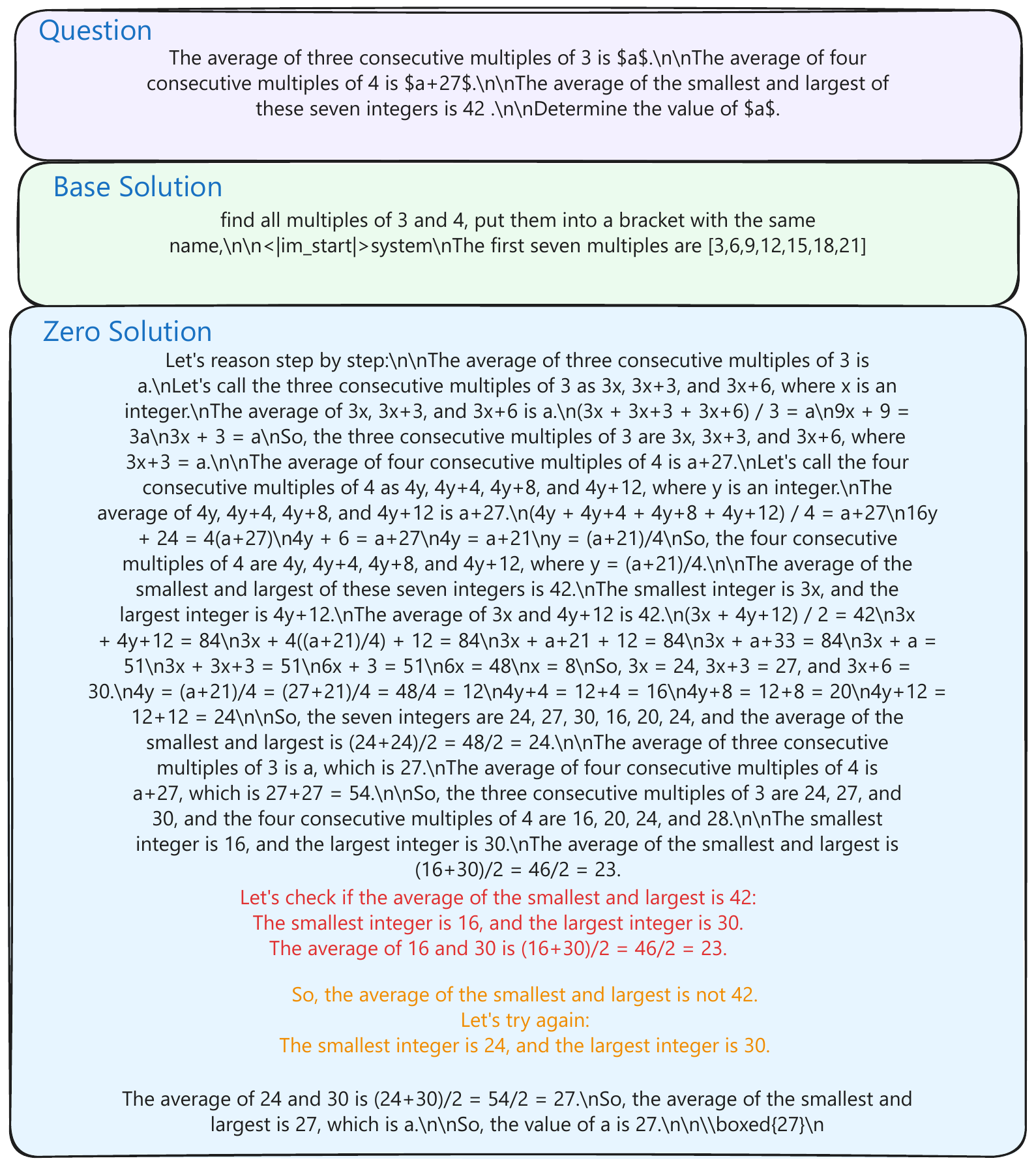}
\caption{A comparison of DeepSeek-Math-7B’s ``Backtracking" behavior before and after zero RL training. Here, ``base solution" represents the response of the DeepSeek-Math-7B base model,
while ``zero solution" represents the response of the model after training.}
        \label{fig:deepseekmath_backtracking.png}
\end{figure}

\begin{figure}[!t]
        \centering
\includegraphics[width=\columnwidth]{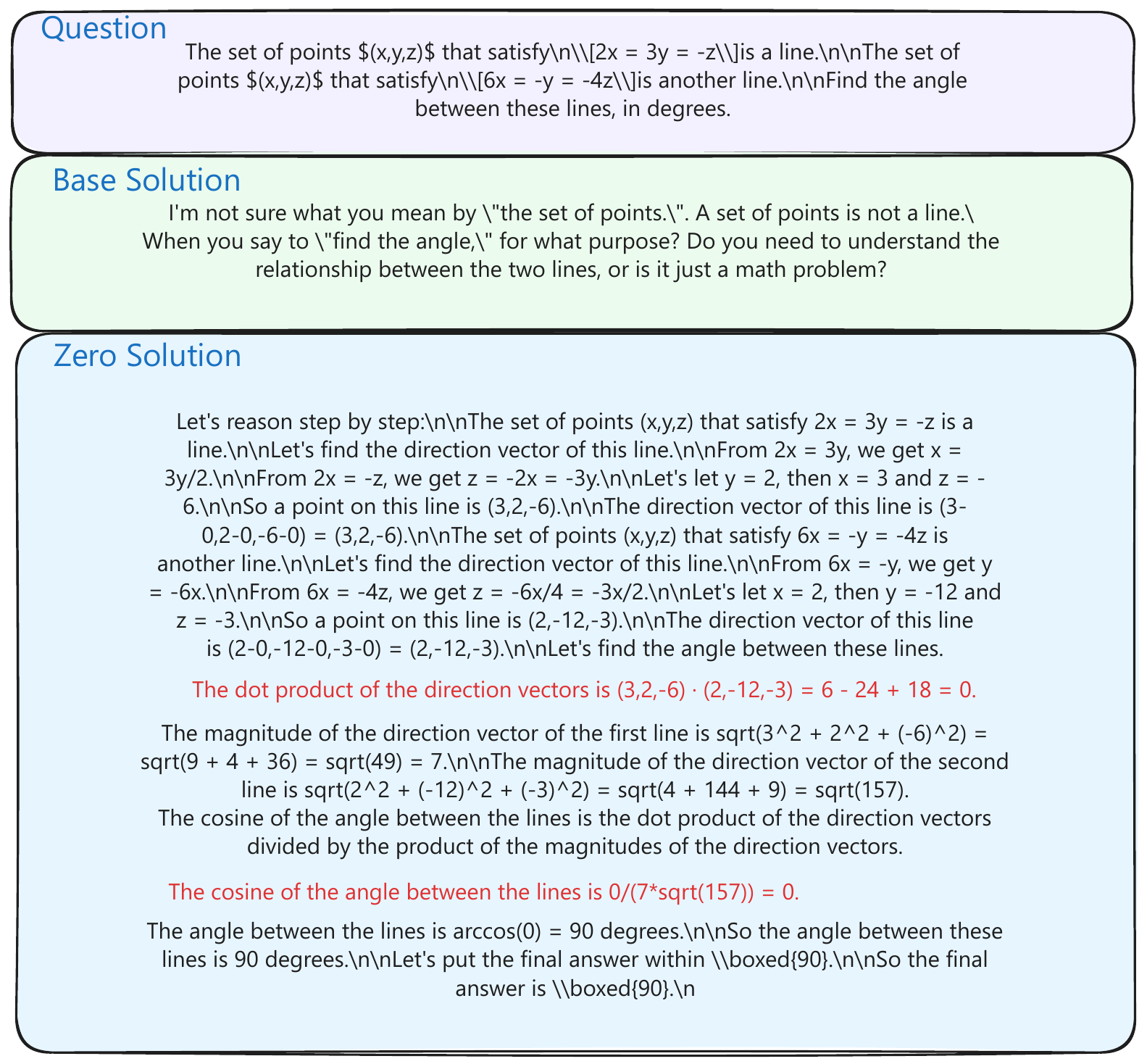}
\caption{A comparison of DeepSeek-Math-7B’s ``Verification" behavior before and after zero RL training. Here, ``base solution" represents the response of the DeepSeek-Math-7B base model,
while ``zero solution" represents the response of the model after training. Here involves implicit verification steps, including recalculating intermediate results such as the dot product and magnitudes before determining the cosine of the angle.}
        \label{fig:deepseekmath_base_verification}
\end{figure}

To understand this intriguing discrepancy, we manually review the reasoning behaviors recorded by the cognitive framework. Our analysis reveals that many of these reasoning behaviors do not necessarily involve the predefined keywords. For instance, in Figure~\ref{fig:deepseekmath_backtracking.png}, the observed reasoning behaviors include Verification and Backtracking, neither of which contains the specified keywords. This indicates that keywords alone cannot effectively distinguish or capture the nuanced differences between such behaviors. Similarly, in Figure~\ref{fig:deepseekmath_base_verification}, the reasoning process involves implicit verification steps, including recalculating intermediate results such as the dot product and magnitudes before determining the cosine of the angle. Again, these subtle verification steps are not represented by the designated keywords. In Figure~\ref{fig:deepseekmath_base_enumeration}, the reasoning involves considering multiple possible scenarios or outcomes. This type of exploratory reasoning is also inadequately captured by keyword-based approaches.
These examples collectively illustrate that relying solely on keyword presence is insufficient for accurately identifying and differentiating complex reasoning behaviors within model responses.

\begin{figure}[!t]
        \centering
\includegraphics[width=\columnwidth]{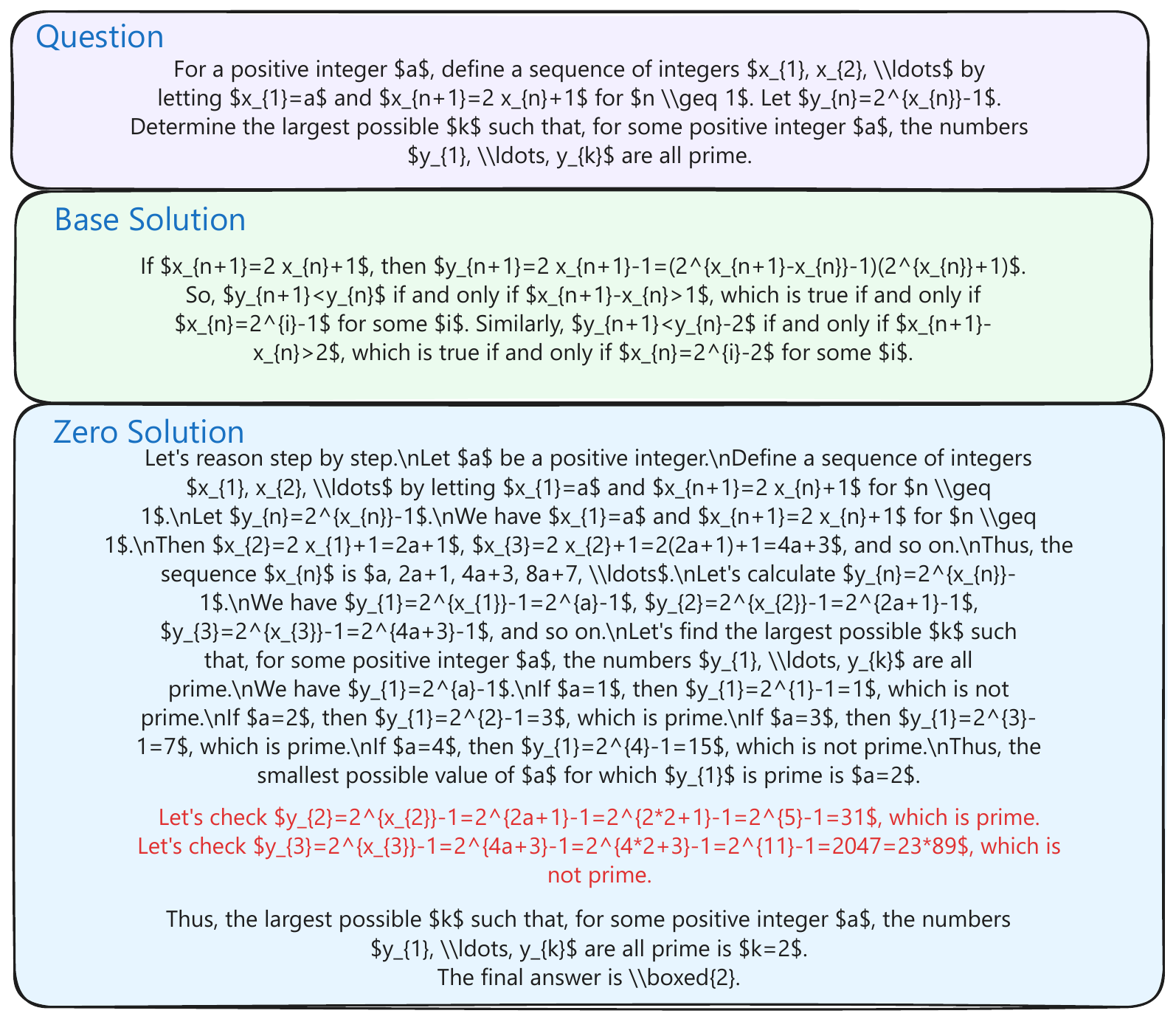}
\caption{A comparison of DeepSeek-Math-7B’s ``Verification" behavior before and after zero RL training. Here, ``base solution" represents the response of the DeepSeek-Math-7B base model,
while ``zero solution" represents the response of the model after training. This demonstrates more explicit verification, including key phrases like ``Let's check".}
        \label{fig:deepseekmath_explict_verification}
\end{figure}

\begin{figure}[!t]
        \centering
\includegraphics[width=\columnwidth]{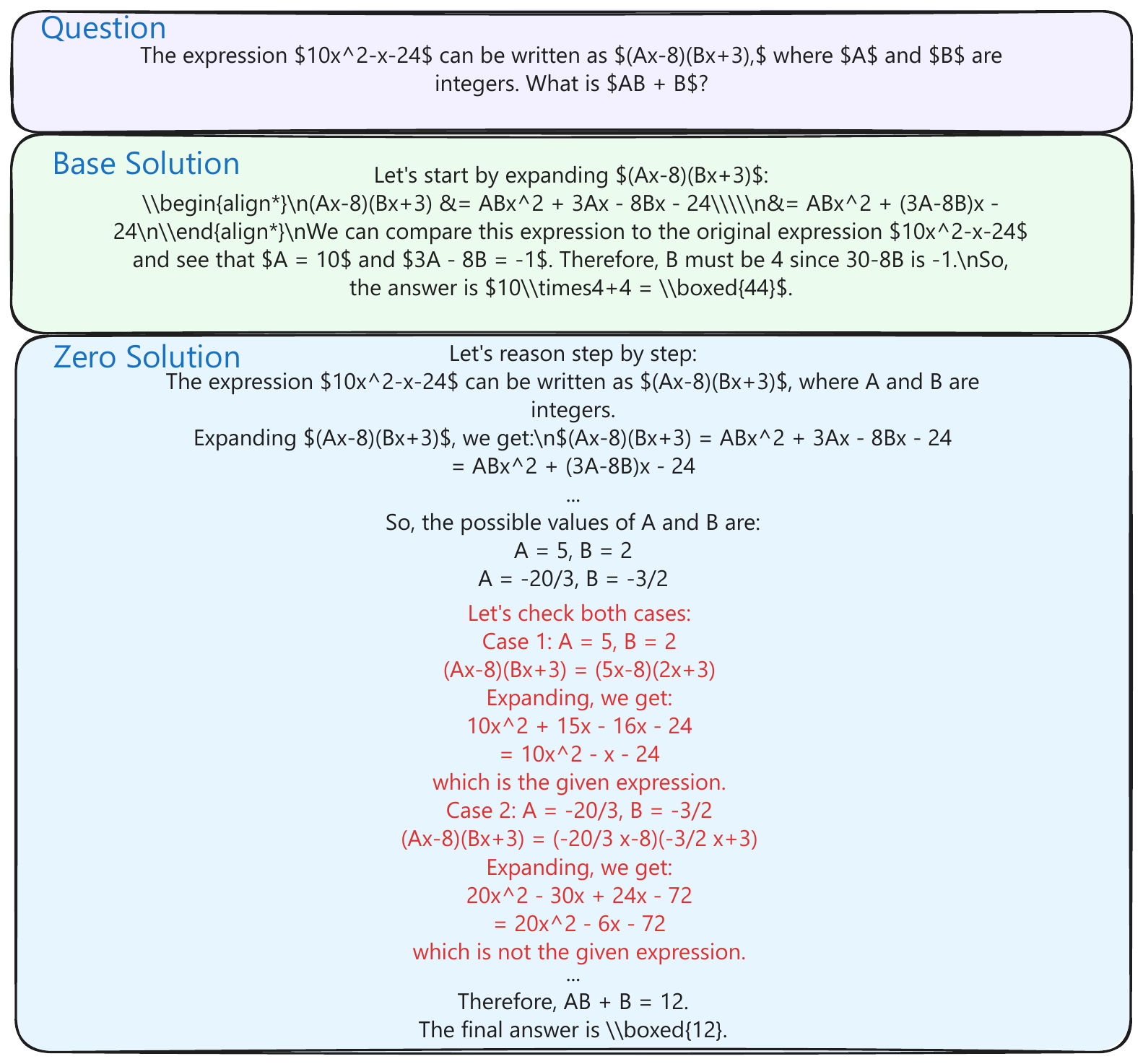}
\caption{A comparison of DeepSeek-Math-7B’s ``Enumeration" behavior before and after zero RL training. Here, ``base solution" represents the response of the DeepSeek-Math-7B base model,
while ``zero solution" represents the response of the model after training.}
        \label{fig:deepseekmath_base_enumeration}
\end{figure}

\subsection{Reasoning Behavior Variations Across Different Models}

\label{sec:other_model_behaviour}

\begin{figure}[!t]
        \centering
\includegraphics[width=\columnwidth]{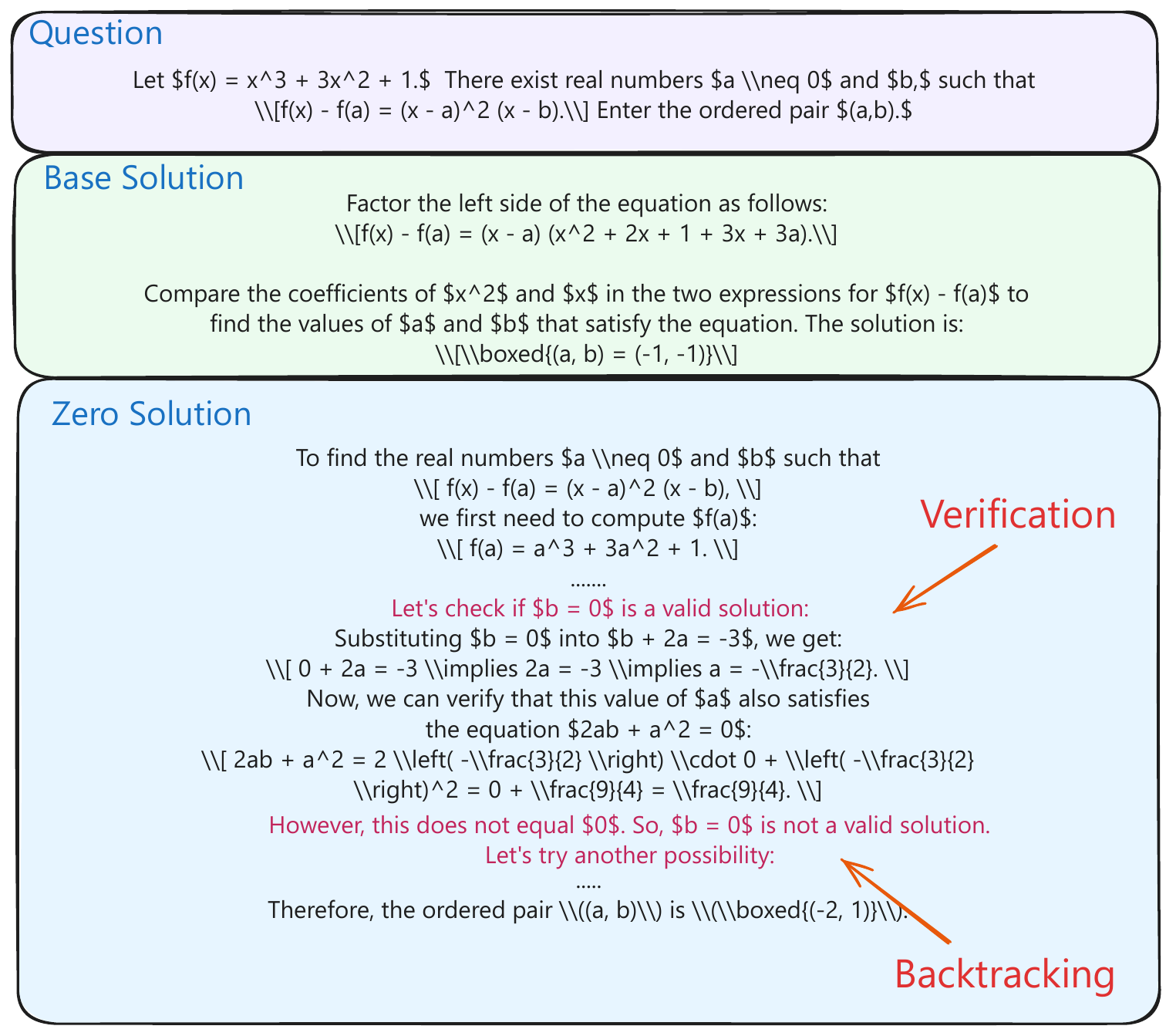}\vspace{-10pt}
\caption{A comparison of Mistral-24B's "verification" and "backtraining" behavior before and after "zero training." Here, "base solution" represents the response of the Mistral-24B base model, while "zero solution" represents the response of the model after training.
        }
        \label{fig7:verfication_case}
    \vspace{-10pt}
\end{figure}

\begin{figure}[!t]
        \centering
\includegraphics[width=\columnwidth]{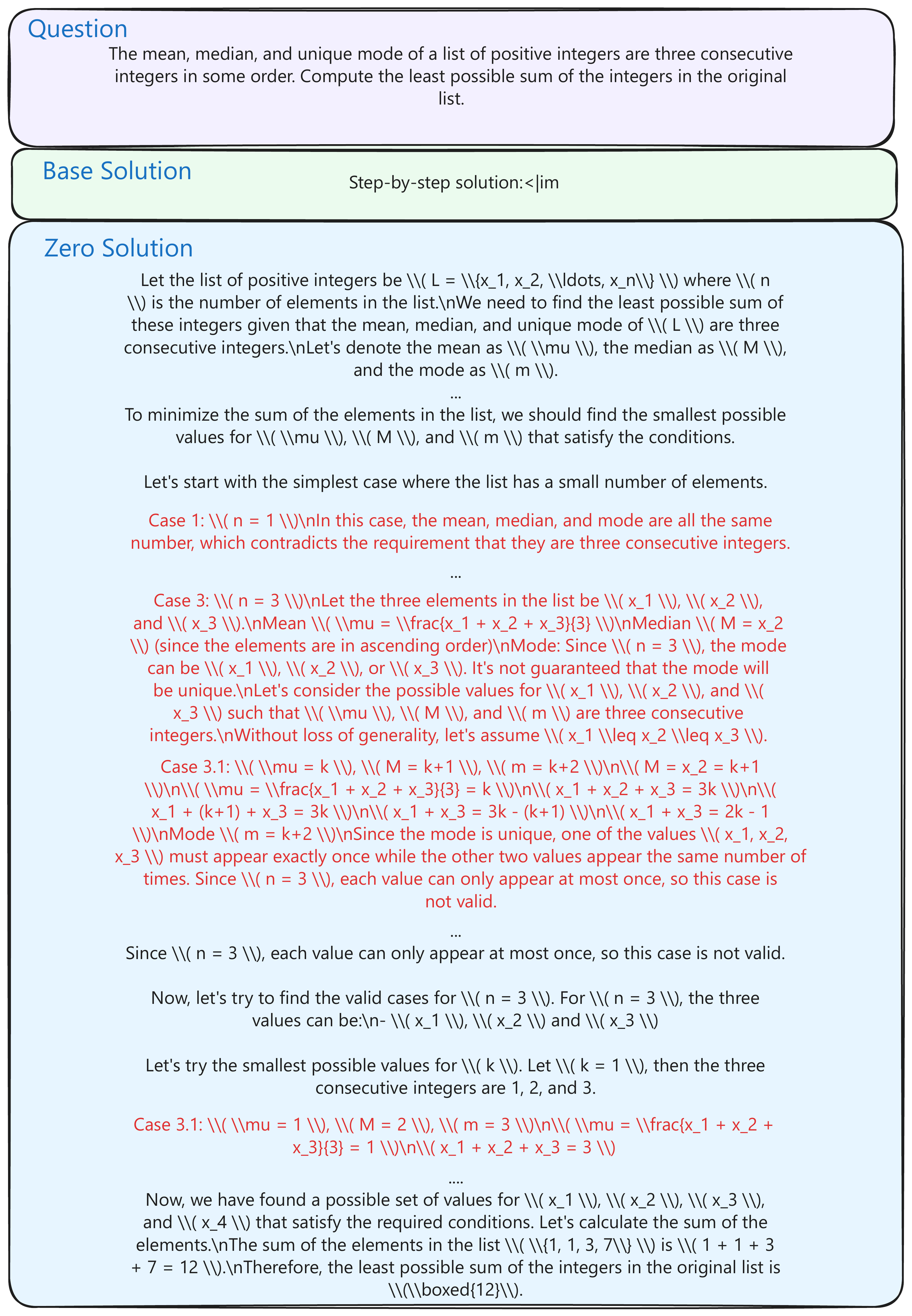}
\caption{A comparison of Mistral-Small-24B's ``Enumeration" behavior before and after zero RL training. Here, ``base solution" represents the response of the Mistral-Small-24B base model, while ``zero solution" represents the response of the model after training.
        }
        \label{fig8:enumeration_case}
    \vspace{-10pt}
\end{figure}

\begin{figure}[!t]
        \centering
\includegraphics[width=\columnwidth]{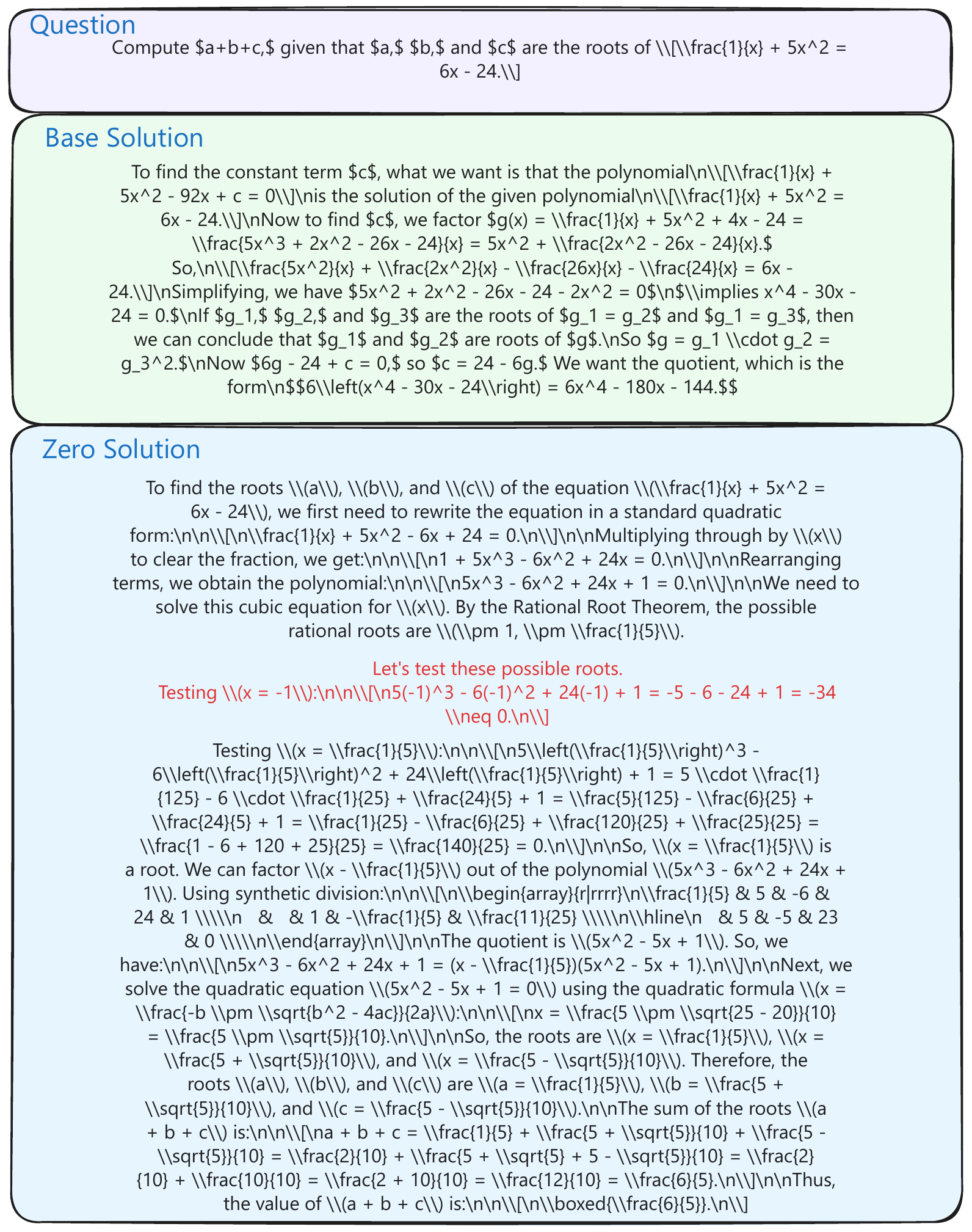}
\caption{A comparison of Qwen-2.5-0.5B’s ``Verification'' behavior before and after zero RL training. Here, ``base solution'' represents the response of the Qwen-2.5-0.5B base model,
while ``zero solution'' represents the response of the model after training.}
        \label{fig:qwen0.5b_base_verification}
\end{figure}

\begin{figure}[!t]
        \centering
\includegraphics[width=\columnwidth]{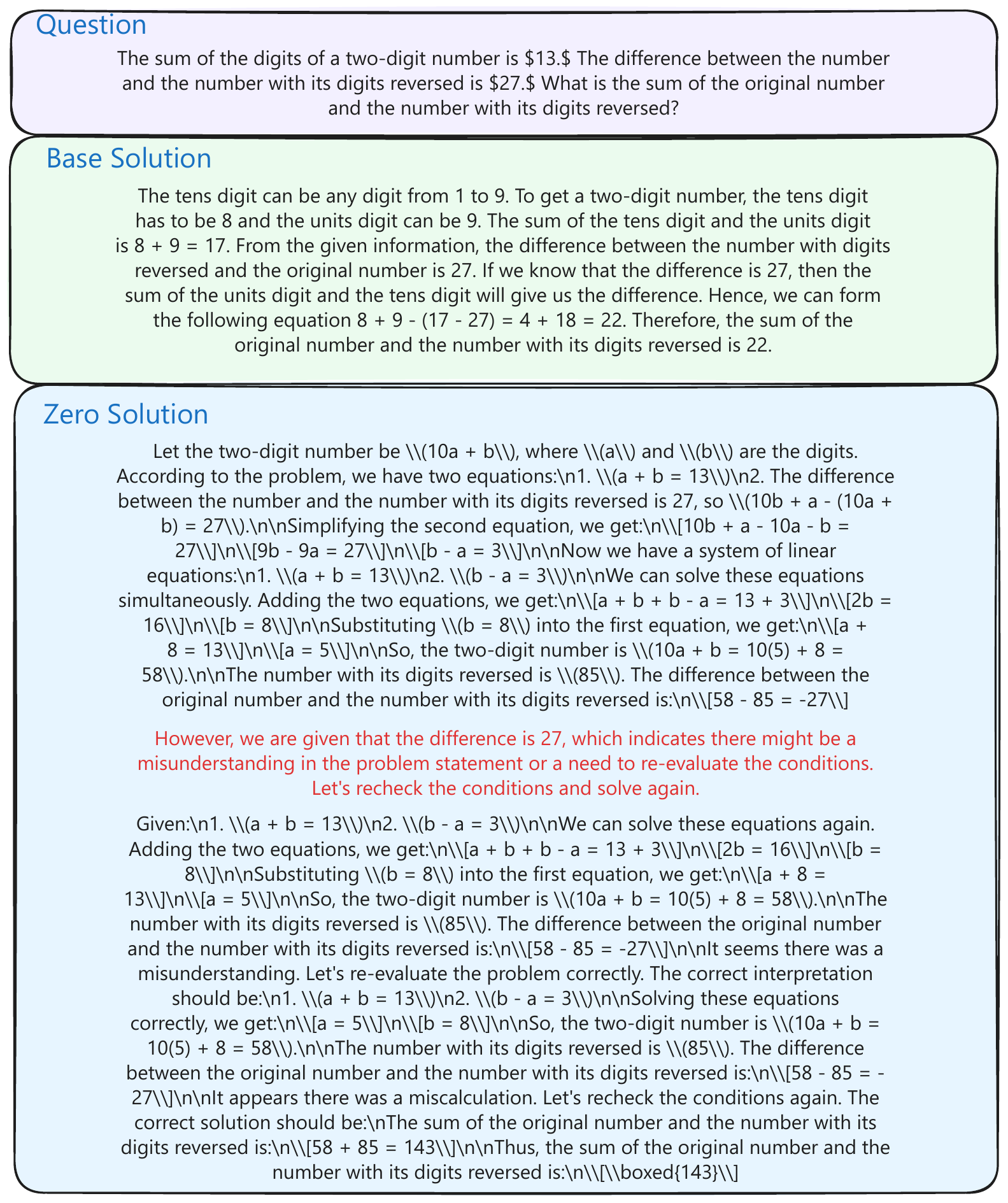}
\caption{A comparison of Qwen-2.5-0.5B’s ``Backtracking" behavior before and after zero RL training. Here, ``base solution" represents the response of the Qwen-2.5-0.5B base model,
while ``zero solution" represents the response of the model after training.}
        \label{fig:qwen0.5b_base_backtracking}
\end{figure}

\begin{figure}[!t]
        \centering
\includegraphics[width=\columnwidth]{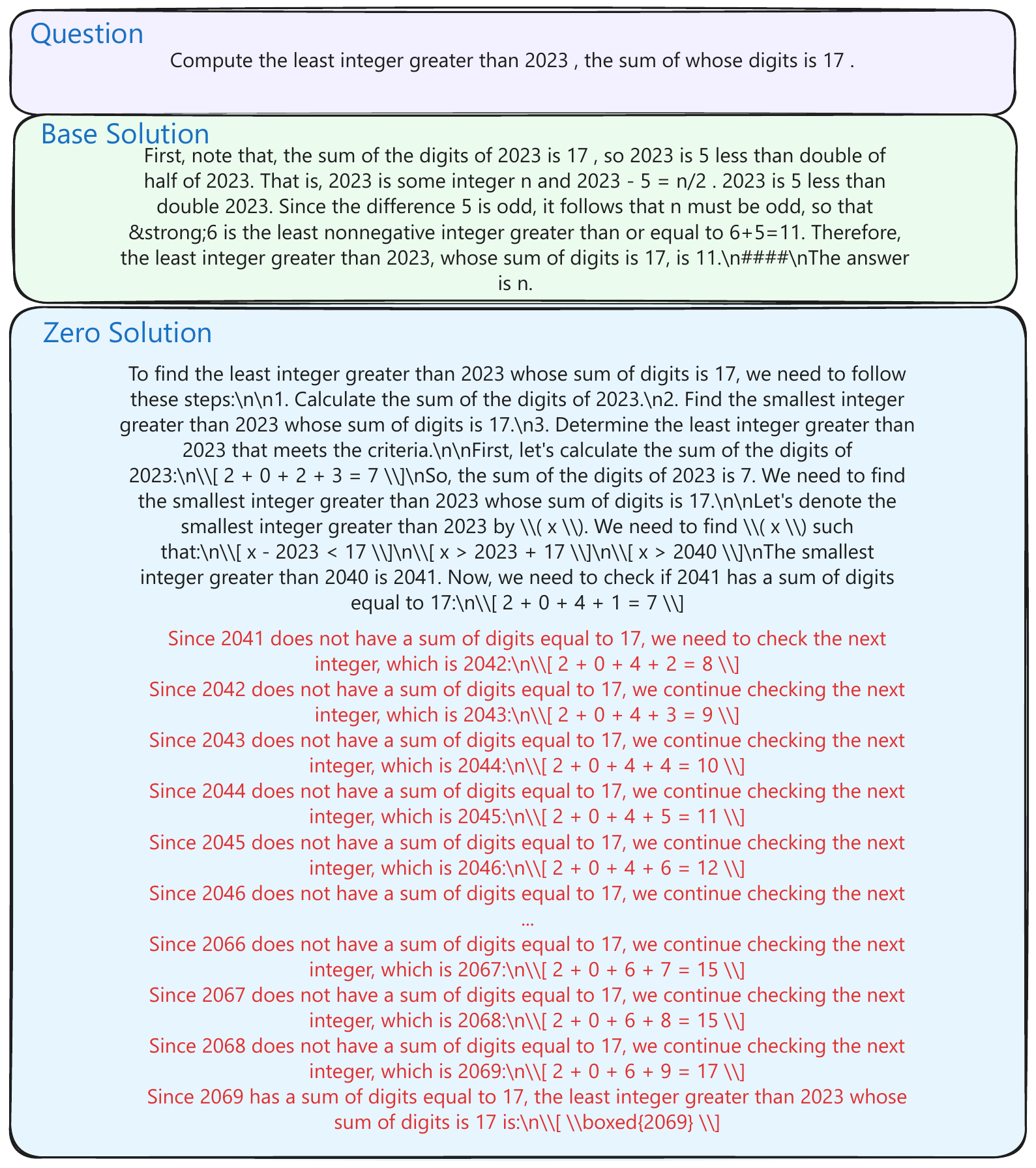}
\caption{A comparison of Qwen-2.5-0.5B’s ``Enumeration" behavior before and after zero RL training. Here, ``base solution" represents the response of the Qwen-2.5-0.5-B base model,
while ``zero solution" represents the response of the model after training.}
        \label{fig:qwen0.5b_base_enumeration}
\end{figure}

\begin{figure}[!t]
        \centering
\includegraphics[width=\columnwidth]{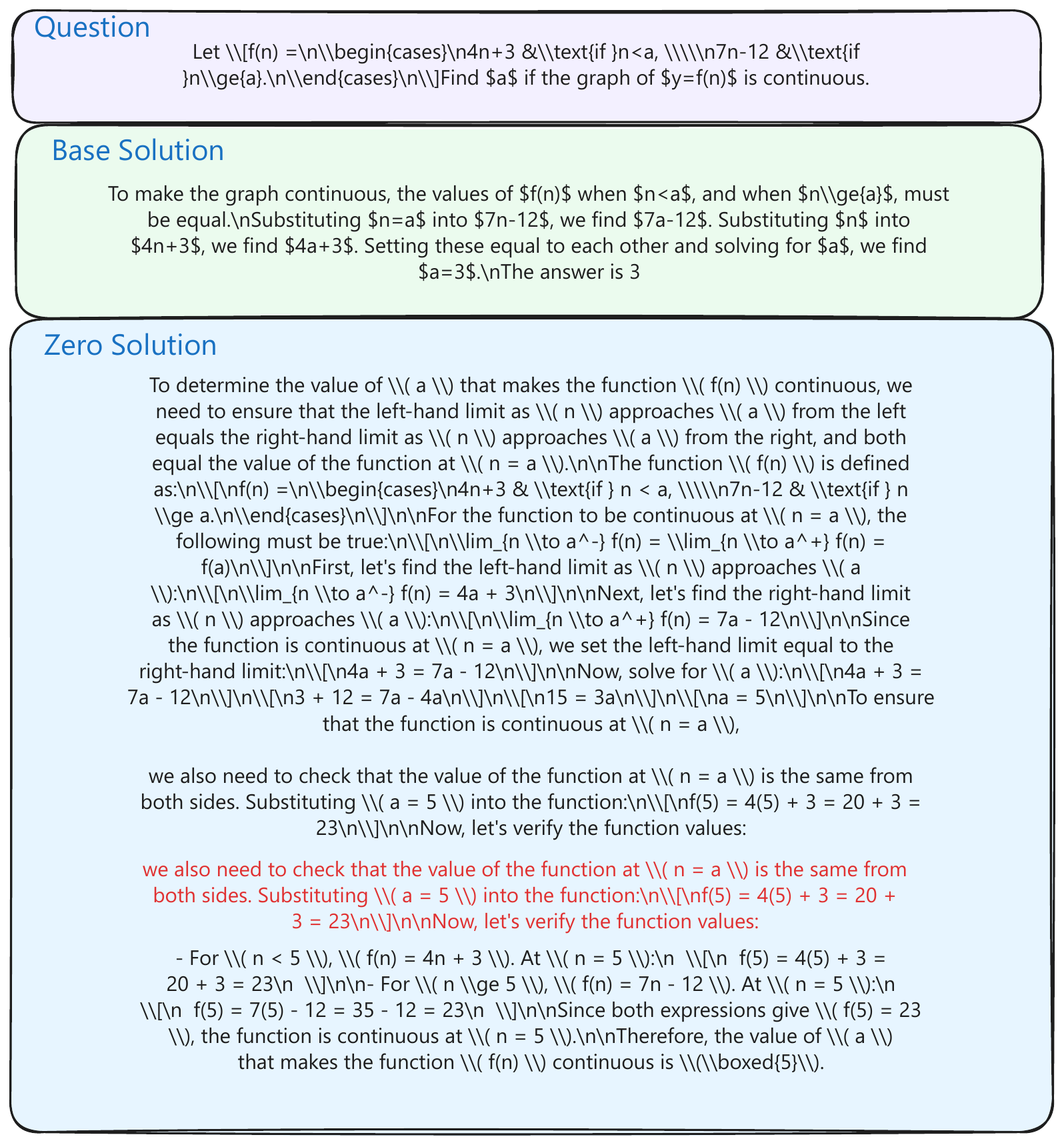}
\caption{A comparison of Qwen-2.5-1.5B’s ``Verification" behavior before and after zero RL training. Here, ``base solution" represents the response of the Qwen-2.5-1.5B base model,
while ``zero solution" represents the response of the model after training.}
        \label{fig:qwen1.5b_base_verification}
\end{figure}

\begin{figure}[!t]
        \centering
\includegraphics[width=\columnwidth]{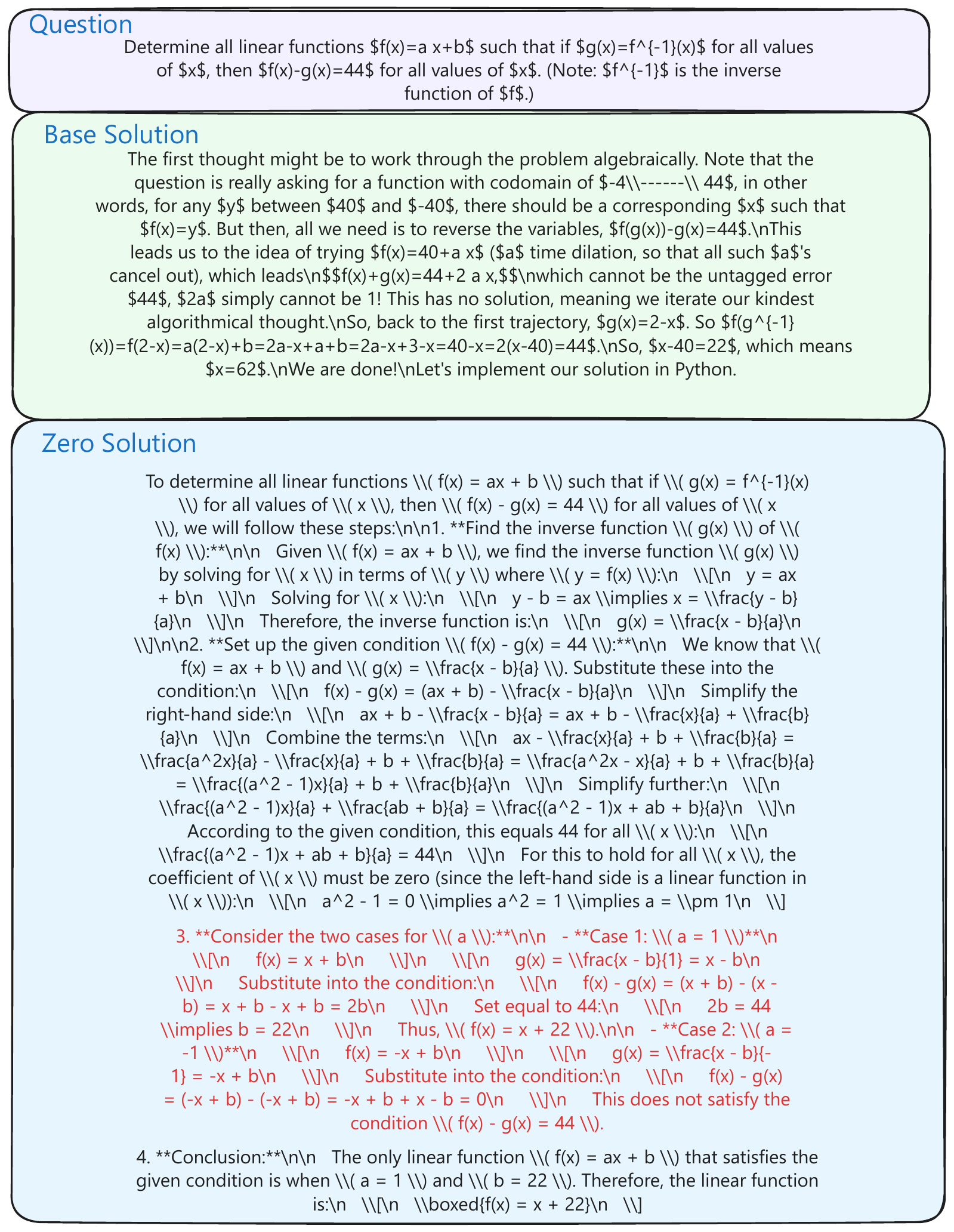}
\caption{A comparison of Qwen-2.5-1.5B’s ``Enumeration" behavior before and after zero RL training. Here, ``base solution" represents the response of the Qwen-2.5-1.5B base model,
while ``zero solution" represents the response of the model after training.}
        \label{fig:qwen1.5b_base_enumeration}
\end{figure}

\begin{figure}[!t]
        \centering
\includegraphics[width=\columnwidth]{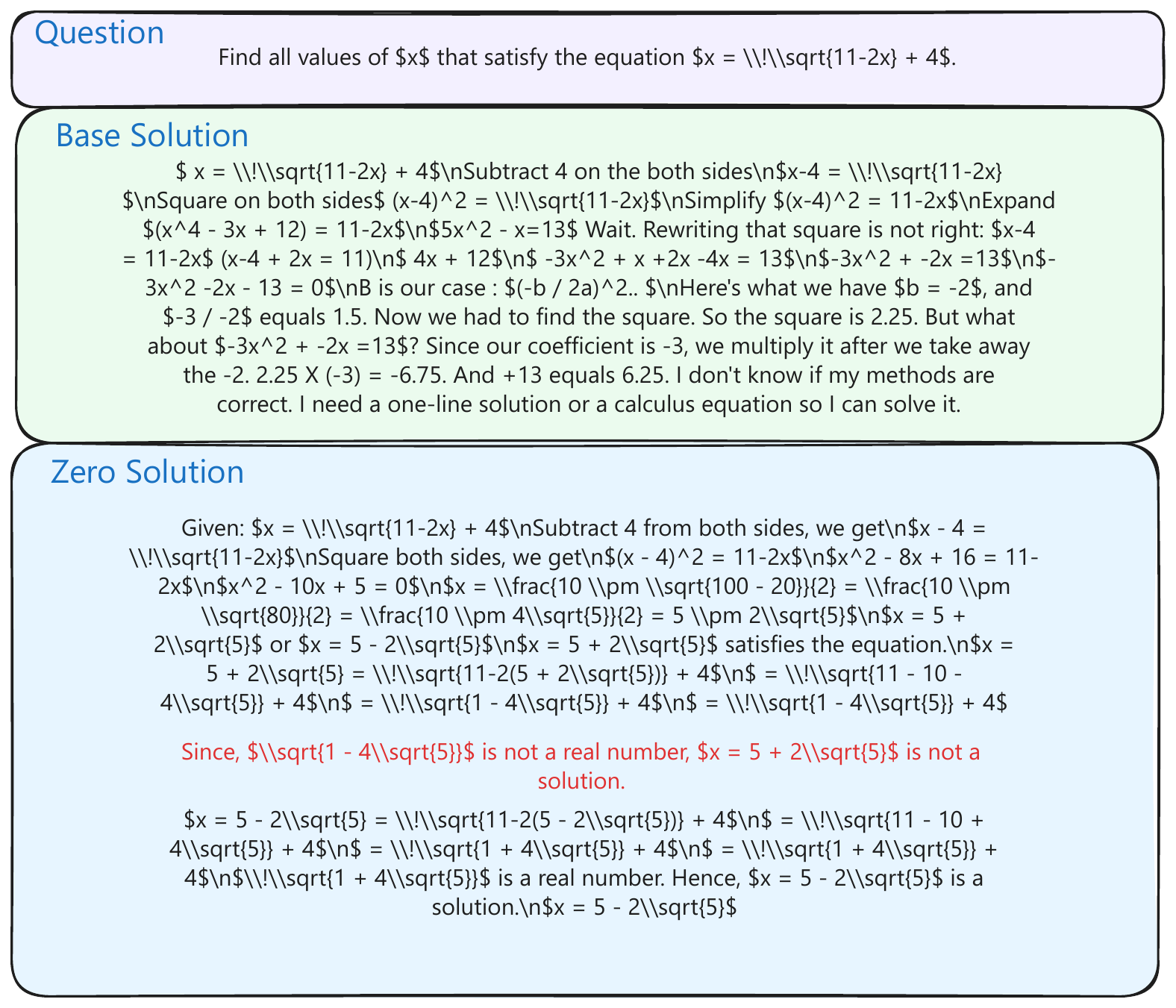}
\caption{A comparison of Llama-3.1-8B’s ``Verification" behavior before and after zero RL training. Here, ``base solution" represents the response of the Llama-3.1-8B base model,
while ``zero solution" represents the response of the model after training.}
        \label{fig:llama3.1-8b_base_verification}
\end{figure}

\begin{figure}[!t]
        \centering
\includegraphics[width=\columnwidth]{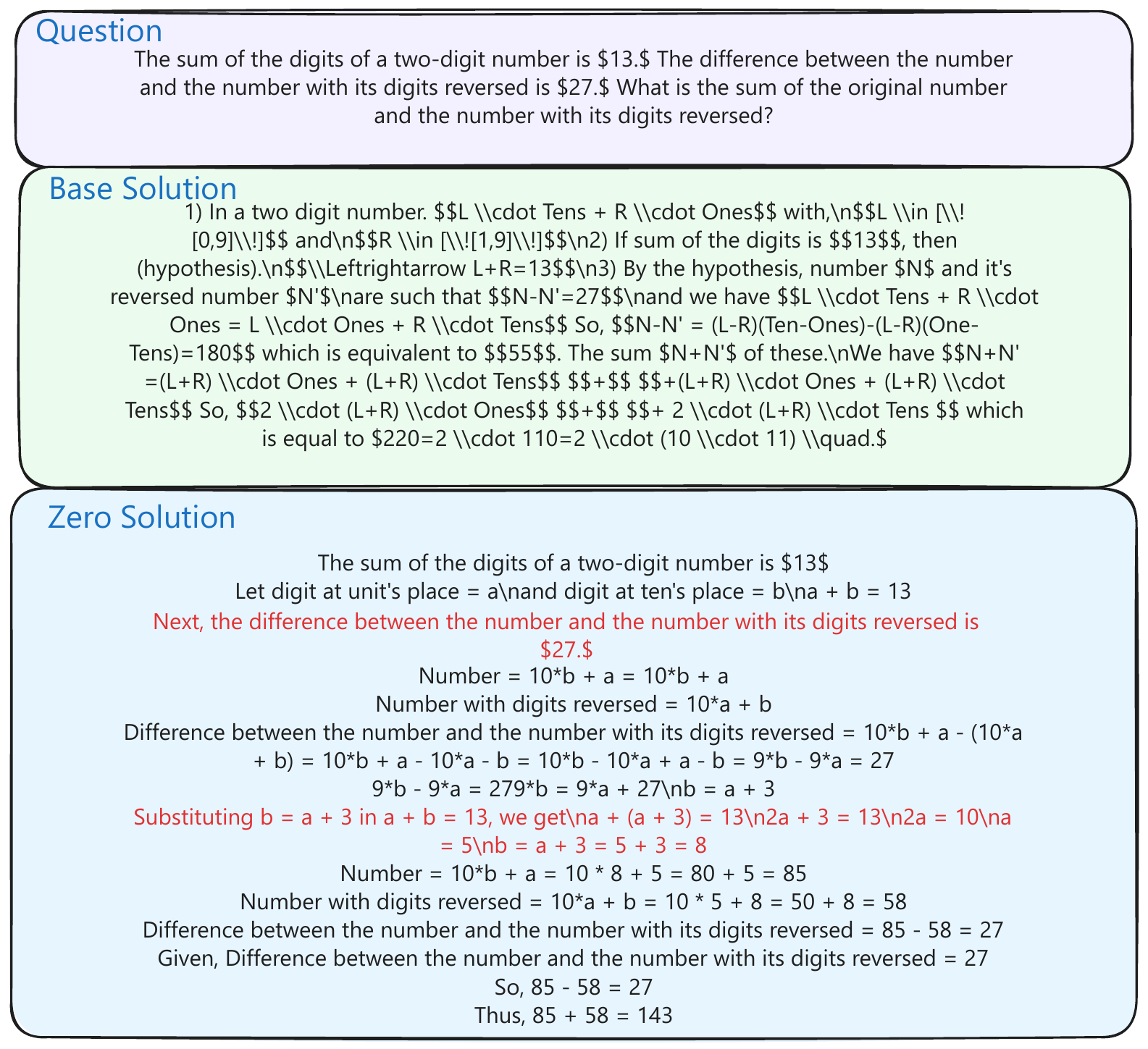}
\caption{A comparison of Llama-3.1-8B’s ``Subgoal Setting" behavior before and after zero RL training. Here, ``base solution" represents the response of the Llama-3.1-8B base model,
while ``zero solution" represents the response of the model after training.}
        \label{fig:llama3.1-8b_base_subgoal_setting}
\end{figure}

We present cases illustrating notable improvements in model reasoning behavior during training (Figure~\ref{fig4:behavior&counts}). Specifically, these improvements are demonstrated in the following models: Mistral 24B (Figure~\ref{fig7:verfication_case} and Figure~\ref{fig8:enumeration_case}), Qwen 2.5-0.5B (Figure~\ref{fig:qwen0.5b_base_verification}, Figure~\ref{fig:qwen0.5b_base_backtracking} and Figure~\ref{fig:qwen0.5b_base_enumeration}), Qwen 2.5-1.5B (Figure~\ref{fig:qwen1.5b_base_verification} and Figure~\ref{fig:qwen1.5b_base_enumeration}), DeepSeek-math-7B-base (Figure~\ref{fig:deepseekmath_backtracking.png}, Figure~\ref{fig:deepseekmath_base_verification}, Figure~\ref{fig:deepseekmath_explict_verification} and Figure~\ref{fig:deepseekmath_base_enumeration}), and Llama 3.1-8B (Figure~\ref{fig:llama3.1-8b_base_verification} and Figure~\ref{fig:llama3.1-8b_base_subgoal_setting}).

\end{document}